\documentclass[times,final]{elsarticle}
\usepackage{jcomp}
\usepackage{booktabs}
\usepackage{framed,multirow}
\usepackage{url}
\usepackage{hyperref}
\usepackage{amsmath}
\usepackage{bm}
\usepackage[font={footnotesize}]{caption} 		
\usepackage{fp}									
\usepackage[final]{pdfpages} 							
\usepackage{subcaption}
\usepackage{pgfplots}								
\usepgfplotslibrary{external} 							

\newcommand*{\shifttext}[2]{%
  \settowidth{\@tempdima}{#2}%
  \makebox[\@tempdima]{\hspace*{#1}#2}%
}

\usepackage{framed,multirow}
\usepackage{amssymb,amsmath,amsthm,bm}
\usepackage{latexsym}
\usepackage{mathtools}

\usepackage{subcaption}
\usepackage{graphicx}

\usepackage{booktabs}
\usepackage{makecell}
\usepackage{enumitem} 

\usepackage{url}
\usepackage{xcolor}

\definecolor{newcolor}{rgb}{.8,.349,.1}

\usepackage[mathlines]{lineno}

\newcommand*\patchAmsMathEnvironmentForLineno[1]{
	\expandafter\let\csname old#1\expandafter\endcsname\csname #1\endcsname
	\expandafter\let\csname oldend#1\expandafter\endcsname\csname end#1\endcsname
	\renewenvironment{#1}
	{\linenomath\csname old#1\endcsname}
	{\csname oldend#1\endcsname\endlinenomath}}
\newcommand*\patchBothAmsMathEnvironmentsForLineno[1]{
	\patchAmsMathEnvironmentForLineno{#1}
	\patchAmsMathEnvironmentForLineno{#1*}}
\AtBeginDocument{
	\patchBothAmsMathEnvironmentsForLineno{equation}
	\patchBothAmsMathEnvironmentsForLineno{align}
	\patchBothAmsMathEnvironmentsForLineno{flalign}
	\patchBothAmsMathEnvironmentsForLineno{alignat}
	\patchBothAmsMathEnvironmentsForLineno{gather}
	\patchBothAmsMathEnvironmentsForLineno{multline}
}

\newtheorem{theorem}{Theorem}[]

\newtheorem{lemma}{Lemma}[]
\theoremstyle{definition}

\theoremstyle{remark}
\newtheorem{remark1}{\bf Remark}[]

\begin{document}
	
	\verso{J.~Chen, K.~Wu}
	
	\begin{frontmatter}
		
		\title{{\bf Deep-OSG: Deep Learning of Operators in Semigroup \tnoteref{tnote1}}}%
		\tnotetext[tnote1]{This work is partially supported by Shenzhen Science and Technology Program
			(grant No.~RCJC20221008092757098) and National Natural Science Foundation of China (grant No.~12171227).}
		
		\author[1]{Junfeng \snm{Chen}}
		\ead{chenjf2@sustech.edu.cn}
		\author[1,2,3]{Kailiang \snm{Wu}\corref{cor1}}
		\cortext[cor1]{Corresponding author.}
		\ead{wukl@sustech.edu.cn}
		
		\address[1]{Department of Mathematics, Southern University of Science and Technology, Shenzhen 518055, China}
		\address[2]{SUSTech International Center for Mathematics, Southern University of Science and Technology, Shenzhen 518055, China}
		\address[3]{National Center for Applied Mathematics Shenzhen (NCAMS), Shenzhen 518055, China}
		

\begin{abstract}
This paper proposes a novel deep learning approach for learning operators in semigroup, with applications to modeling unknown autonomous dynamical systems using time series data collected at varied time lags. 
It is a sequel to the previous flow map learning (FML) works [T.~Qin, K.~Wu, and D.~Xiu, {\em J.~Comput.~Phys.}, 395:620--635,  2019], [K.~Wu and D.~Xiu, {\em J.~Comput.~Phys.}, 408:109307, 2020], and [Z.~Chen, V.~Churchill, K.~Wu, and D.~Xiu, {\em J.~Comput.~Phys.}, 449:110782, 2022], which focused on learning single evolution operator with a fixed time step.  
This paper aims to learn a family of evolution operators with variable time steps, which   constitute a semigroup for an autonomous system.  
The semigroup property is very crucial and links the system's evolutionary behaviors across varying time scales, but it was not considered in the previous works. 
We propose for the first time a framework of embedding the semigroup property into 
the data-driven learning process, 
through a novel neural network architecture and new loss functions. 
The framework is very feasible, can be combined with any suitable neural networks, and is applicable to learning general autonomous ODEs and PDEs. 
We present the rigorous error estimates and variance analysis to understand 
the prediction accuracy and robustness of our approach, showing the remarkable advantages of semigroup awareness in our model. 
Moreover, our approach allows one to arbitrarily choose the time steps for prediction and ensures that the predicted results are well self-matched and consistent. 
Extensive numerical experiments demonstrate that embedding the semigroup property  
notably reduces the data dependency of deep learning models and greatly 
improves the accuracy, robustness, and 
stability for long-time prediction. 

\end{abstract}

\end{frontmatter}



\section{Introduction}

Ordinary differential equations (ODEs) and partial differential equations (PDEs) 
have important applications in understanding the laws of
nature and  modeling complicated dynamic processes in various fields of science and engineering. 
Many fundamental laws in physics are described by time-dependent differential equations, 
for example, the Maxwell equations in electromagnetics, the Navier--Stokes equations in fluid dynamics, and the Schr\"odinger equation in quantum mechanics. 
Existing differential equations are
mostly derived from fundamental physical principles based on established human knowledge. 
However, for many practical problems, their
governing equations remain unknown due to their complex mechanisms. 
The fusion of
artificial intelligence and data science is opening a new paradigm for
modeling unknown differential equations or discovering hidden
models/equations from measurement data, which has attracted extensive attention in recent
years.

Unlike the traditional approach discovering new equations by derivation, data-driven modeling of governing equations tries to learn the underlying unknown dynamics or laws 
directly from the observed data. 
Earlier attempts in this direction can be found in \cite{bongard2007automated,schmidt2009distilling}, which were based on comparing the numerical differentiation of experimental data with the analytic derivatives of candidate functions and then applying symbolic regression to discover nonlinear dynamical systems. 
 A more recent approach endeavors to identify the terms in the  unknown governing equations from an {\em a priori}  dictionary, which includes all possible terms that could exist in the underlying equations; see e.g.~\cite{brunton2016discovering,rudy2017data}. Classic methods along this approach typically use certain  sparsity-promoting algorithms, such as LASSO \cite{tibshirani1996regression} and compressed sensing \cite{donoho2006compressed,candes2006stable}, to construct parsimonious models from a large set of dictionary encompassing all potential models; cf.~\cite{brunton2016discovering,schaeffer2017sparse,schaeffer2017learning,rudy2017data,tran2017exact}. Besides, many machine learning techniques, particularly deep learning, have been developed to discover the forms of governing equations; see, for example,  \cite{long2018pde,raissi2018deep,raissi2018multistep,long2019pde,raissi2019physics,sun2020neupde,chen2021neural}. There are also some recent efforts (e.g.~\cite{atkinson2019data,chen2022symbolic})  aiming to eliminate the requirement of specifying {\em a prior} dictionary of all potential terms in the underlying equations. 
 For more developments, 
 the readers are also referred to some related works based on Gaussian process regression \cite{raissi2017machine}, model selection \cite{mangan2017model}, Koopman theory \cite{brunton2017chaos}, classical polynomial approximation \cite{wu2019numerical,wu2020structure}, linear multistep methods \cite{keller2021discovery}, genetic algorithms \cite{xu2020dlga,xu2021deep,xu2021robust},  and the references therein.

Recently, another approach {termed {\em flow map learning} (FML)} for data-driven modeling of unknown dynamical systems was systematically proposed in a series of papers \cite{qin2019data,wu2020data,chen2022deep,churchill2023flow}. Instead of identifying the terms in the unknown governing equations, the {FML} approach focuses on approximating the {\em evolution operator} (also called {\em flow map}) of the underlying equations. For an autonomous  system, its evolution operator completely describes the system's dynamics over time. 
Consequently, the evolution operator, once successfully learned, can be recursively employed to predict the solution behavior of the unknown dynamical system in the future \cite{qin2019data,wu2020data,chen2022deep}. 
Although the {FML} approach does not aim to directly recover the exact form of the equations, it is indeed equivalent to recovering some integral form of the underlying equations \cite{qin2019data,wu2020data}.  
The {FML} approach avoids the need for numerical approximation of time derivatives 
(which is difficult to obtain and may be subject to large error especially for noisy data) and allows larger time steps during the learning and prediction process. 
Moreover, the {FML} approach based on learning evolution operators does not require a-priori knowledge about all potential terms in the unknown governing equations. 
While the approach of recovering equations should be coupled with appropriate numerical schemes to further solve the learned equations for system predictions, 
the learned evolution operator {via FML} can be directly used to perform predictions. 
 The effectiveness of the {FML} approach has been demonstrated for learning ODEs \cite{qin2019data}, modeling PDEs in both generalized Fourier spaces \cite{wu2020data} and nodal space \cite{chen2022deep}. In more recent works, the {FML} approach has also been extended to data-driven modeling of parametric differential equations \cite{qin2021deep}, partially observed systems \cite{fu2020learning}, non-autonomous systems \cite{qin2021data}, model correction \cite{chen2021generalized}, biological models \cite{su2021deep}, {chaotic systems \cite{churchill2022deep}, and stochastic dynamical systems \cite{chen2023learning}.} These advancements demonstrate the applicability and potential of the {FML} approach in various fields. 
 
More recently, deep learning techniques have also been developed for approximating general operators which are maps between two infinite-dimensional function spaces, such as integrals, the Laplace transform, and the solution operators of differential equations. The related works in this direction include but are not limited to DeepONet \cite{lu2021learning,cai2021deepm,mao2021deepm}, neural operator \cite{kovachki2021neural}, multipole graph neural operator \cite{li2020multipole}, Fourier neural operator (FNO) \cite{li2021fourier}, Markov neural operator \cite{li2021markov}, and nonlocal kernel network \cite{you2022nonlocal}, etc. 

This paper aims to develop a novel deep learning approach for approximating a family of evolution operators in semigroup, with applications to data-driven modeling of unknown autonomous ODE and PDE systems. Unlike the previous {FML} works \cite{qin2019data,wu2020data,chen2022deep} which focused on learning a single evolution operator with a constant time step, our new approach aims to learn a family of evolution operators with variable time steps. This allows us to model unknown autonomous dynamical systems from time series data collected at varied time lags. For an autonomous system, all the evolution operators with different time lags constitute a one-parameter {\em semigroup} in mathematics. 
The semigroup property is very crucial, as it links the system's evolutionary behaviors across varying time scales. However, such an important property has not been taken into account in the previous {FML} methods developed in \cite{qin2019data,wu2020data,chen2022deep}. 
The main objective of this work is to carefully embed the semigroup property into the data-driven learning process, yielding the semigroup-informed models that demonstrate improved performance in terms of robustness and reduced error accumulation over time, in comparison with the purely data-driven approach. 
The contribution, novelty, and significance of this paper are outlined as follows: 

\begin{itemize}
    \item We present a new framework for learning evolution operators in semigroup and 
    modeling unknown ODEs and PDEs using data. The framework incorporates the evolution time as an input feature in deep neural networks, enabling the learning of a family of evolution operators from observed data with varied time steps, rather than a single evolution operator defined by a fixed time step. Most notably, we carefully embed 
    the semigroup property of the evolution operators into our framework.
    The framework produces data-driven models continuous in time and allows one to perform 
    long-time predictions with arbitrary evolution time steps from any given initial states.  
    \item In order to effectively embed the semigroup property, 
    we design a new residual network architecture for our framework. 
    Besides adding the initial state to the output layer, we propose a second skip connection for the time step, by multiplying it with the network's output layer. The new architecture strictly enforces a desired semigroup constraint such that the neural network model degenerates to the identity map if the time stepsize reduces to zero. 
    \item We derive a novel loss function combining the vanilla fitting loss with a regularization term informed by the semigroup property. 
    Through comparisons by extensive numerical experiments, we construct  
    a highly effective semigroup-informed 
    regularization term, which does not depend on the observed data. 
    Instead, the regularization term 
     is carefully designed based on semigroup 
    residues calculated by randomly generated initial states and time lags, thereby effectively enforcing the semigroup property on the entire domain of interest. 
     The semigroup awareness in our model 
      significantly improves the prediction accuracy, and greatly enhances the robustness and stability of long-time prediction. 
      Moreover, our model allows one to arbitrarily choose the time steps for prediction, and also ensures that the predicted results are well self-matched and consistent for different partitions of the time interval. 
      \item We carry out the rigorous error estimates and variance analysis to understand 
      the prediction accuracy and robustness of our method. 
    \item It is worth mentioning the general flexibility of our framework. First, 
    since our network architecture and semigroup-informed loss function are non-intrusive to the basic network structures, the framework is very feasible and can be combined with any suitable neural networks including fully-connected neural networks (FNNs), convolutional neural networks, locally connected networks, and FNO, etc. In our numerical experiments, we implement the combinations of our framework with various neural networks. 
    Secondly, our framework is applicable to a wide range of autonomous ODEs and PDEs. We will first present the framework on learning unknown ODEs and then extend it to modeling unknown PDEs in either modal or nodal space. 
    \item We present extensive 
   numerical experiments on various ODEs and PDEs to demonstrate the effectiveness of our method.  
   These include a highly stiff ODE system to show that our model is able to accurately capture 
   the multiscale dynamics in long-time prediction. 
   Interestingly, we observe that our data-driven method, although resembling an explicit algorithm, allows a very large time step even for very stiff problems. 
   We also present a challenging example on modeling the Navier--Stokes equations. 
   The numerical results validate 
   the good stability and robustness of our semigroup-informed method in long-time prediction.   
\end{itemize}
Notice that {a substantial} amount of data was used in the previous works \cite{qin2019data,wu2020data,chen2022deep}. For example, tens of thousands data pairs were used in learning ODEs {in} \cite{qin2019data}, and hundreds of thousands data were employed in modeling PDEs in \cite{wu2020data}.  However, in practice, collecting measurement data from real-world systems can be very costly or difficult due to some resource constraints or limited experimental accessibility. 
In the present paper, we shall assume that the measurement data can be very limited. In this case, we find embedding the semigroup constraint is very helpful or even essential to avoid over-fitting and notably reduce the required data, as the semigroup-informed loss function plays an important role of regularization.

The remainder of this article is organized as follows. After the basic setup in section
\ref{section:pb_set}, we present the deep learning framework for learning evolution operators in semigroup for autonomous ODEs in Section \ref{section:osg_framework}. The framework is extended to modeling PDEs in Section \ref{sec:PDE}. Several numerical experiments are presented in Section \ref{section:egs}, before concluding the paper in Section \ref{section:conclude}.

\section{Setup and preliminaries}
\label{section:pb_set}


We consider a set of state variables ${\bm u}\in \mathbb R^n$, which are governed by an {\em unknown} autonomous time-dependent 
differential system. 
Assume that the state variables ${\bm u}$ are measurable or observable, namely, 
the measurement data of ${\bm u}$ are available. 
We aim to seek accurate data-driven modeling of the unknown autonomous system and construct an accurate {predictive} model for the unknown dynamics based on the state variable data.

For the sake of convenience, we first consider the ODE case that 
the underlying unknown governing equations are autonomous ODEs    
\begin{equation}
	\frac{d\bm{u}}{dt}=\bm{f}(\bm{u}),\quad \bm{u}(t_0)=\bm{u}_0 
	\label{equation:autonomous_ds}
\end{equation}
with $\bm{f}: \mathbb R^n \to \mathbb R^n$ being unknown, while the extension of our discussions to the PDE case will be studied in Section \ref{sec:PDE}. 
Assume that the measurement data of ${\bm u}$ are collected along a number of different trajectories. Let $t_0<t_1< \dots < t_K$ be a sequence of time instances. 
We use 
\begin{equation}\label{eq:ORIGdata}
	{\bm u}_k^{(i)} = {\bm u}( t_k; {\bm u}_{0}^{(i)}, t_0) + {\bm \epsilon}_k^{(i)}, \qquad k=1,\dots,K_i, \quad i=1,\dots,{I_{traj}}
\end{equation}
to denote the solution state measured at the time instance $t_k$ along the $i$-th trajectory originated from the initial state ${\bm u}_{0}^{(i)}$ at $t_0$, for a total number of $I_{traj}$ trajectories. In practice, the data may contain some measurement noises ${\bm \epsilon}_k^{(i)}$, which {is} usually modeled as random variables.

\subsection{Evolution operator}

While many of the existing {works} seek to directly learn the right-hand-side term $\bm{f}$ of the governing equations, we here adopt a different approach {that} seeks to approximate the evolution operator of the underlying equations. 
The idea of this approach was proposed in \cite{qin2019data} for modeling unknown ODEs and further developed in \cite{wu2020data,chen2022deep} for modeling unknown PDEs. 

An evolution operator ${\bf \Phi}_\Delta:\mathbb{R}^n \to \mathbb{R}^n$, which is sometimes also called {\em flow map} in the ODE community, describes the evolution of state variables from time $t_0$ to time $t_0+\Delta$. It can be defined through the integral form of  equations \eqref{equation:autonomous_ds} as follows 
\begin{equation}
	{\bf \Phi}_\Delta(\bm{u}_0) := \bm{u}(t_0+\Delta;\bm{u}_0,t_0) = \bm{u}_0 + \int_{t_0}^{t_0+\Delta} \bm{f}(\bm{u}(s)){\rm d}s = 
	\left[ {\bf I}_n + \Delta {\bm \phi} (\cdot, \Delta) \right] ( \bm{u}_0 ),
	\label{equation:flow_map}
\end{equation}
where ${\bf I}_n$ is the identity matrix of size $n \times n$, and for any ${\bm u}\in \mathbb R^n$,
\begin{equation}\label{eq:phi}
	{\bm \phi} (\cdot, \Delta ) [{\bm u}] = {\bm \phi} ( {\bm u}, \Delta ) = \frac{1}{\Delta} \int_{0}^\Delta {\bm f}(  {\bf \Phi}_s(\bm{u}) ) {\rm d}s = {\bm f}(  {\bf \Phi}_\tau(\bm{u}) )
\end{equation}
denotes the time-averaged increment, where the mean value theorem is used with some $\tau \in [0,\Delta]$.

For autonomous systems, the evolution operator completely determines the evolution of the solution from one state to another state at a future time. Learning the evolution operator from data allows us to conduct prediction of the system via recursively using the learned evolution operator, as demonstrated in the FML framework \cite{qin2019data,wu2020data,chen2022deep}.

\subsection{Semigroup property}

For an autonomous system \eqref{equation:autonomous_ds}, all the evolution operators $\{{\bf \Phi}_\Delta\}_{\Delta \ge 0}$ constitute a one-parameter {\em semigroup},  namely, they satisfy 
\begin{subequations}\label{equation:semigroup_property}
	\begin{align}
		& {\bf \Phi}_{0} = {\bf I}_n, \label{equation:semigroup_property1}\\
		\smallskip
		& {\bf \Phi}_{\Delta_1+\Delta_2} = {\bf \Phi}_{\Delta_1} \circ {\bf \Phi}_{\Delta_2} \quad \forall \Delta_1,\Delta_2\in \mathbb{R}^+. 
		\label{equation:semigroup_property2}
	\end{align}
\end{subequations}
The first constraint (\ref{equation:semigroup_property1}) requires the evolution operator to be the identity map when the time step $\Delta =0$, which is natural for time-dependent differential equations. The second constraint (\ref{equation:semigroup_property2}) distinguishes the forward dynamics of autonomous systems from non-autonomous ones, and it connects the system evolution behaviors at different time scales. 

Instead of learning a single evolution operator with a fixed time step $\Delta$ as in \cite{qin2019data,wu2020data}, 
in the present paper we aim to  establish a new numerical framework for learning a family of evolution operators $\{{\bf \Phi}_\Delta\}_{\Delta \ge 0}$ with variable time steps $\Delta \in [0,T]$ 
and especially embedding the semigroup property \eqref{equation:semigroup_property} into the learning process.  

\subsection{Data reorganization}

In order to learn the evolution operator with the semigroup property \eqref{equation:semigroup_property} embedded, we 
reorganize the data \eqref{eq:ORIGdata} into several groups, and each group has three samples at neighboring time instances separated by two time lags $\Delta_k:=t_{k+1}-t_k$ and $\Delta_{k+1}:=t_{k+2}-t_{k+1}$, as follows
\begin{equation*}
	\left\{ {\bm u}_k^{(i)}, {\bm u}_{k+1}^{(i)}, {\bm u}_{k+2}^{(i)} \right\}, \qquad k=1,\dots,K_i-2, \quad i=1, \dots, I_{traj}.
\end{equation*}
Note that for autonomous systems, the time variable $t$ can be arbitrarily shifted and only the time difference $\Delta$ is relevant.  
Hence, 
for notational convenience, we denote the entire {dataset} as 
\begin{equation}\label{eq:data}
	\left\{ \bm{u}_{0,j},~ \Delta_{1,j},~ \bm{u}_{1,j},~ \Delta_{2,j},~  \bm{u}_{3,j} \right\}_{j=1}^J
\end{equation}
with $J=\sum_{i=1}^{I_{traj}}(K_i-2)$ is the total number of data groups. If the data are noiseless, we have the following relations 
\begin{equation*}
	\bm{u}_{1,j} = {\bf \Phi}_{ \Delta_{1,j} } ( \bm{u}_{0,j} ), \quad 
	\bm{u}_{2,j} = {\bf \Phi}_{ \Delta_{2,j} } ( \bm{u}_{1,j} ), 
	\quad 
	\bm{u}_{2,j} = {\bf \Phi}_{ \Delta_{1,j} + \Delta_{2,j} } ( \bm{u}_{0,j} ).
\end{equation*}

\section{Deep-OSG approach for evolution operator learning}
\label{section:osg_framework}
In this section, we present the deep learning approach, termed as Deep-OSG, for approximating a family of evolution operators with semigroup property \eqref{equation:semigroup_property}. 

\subsection{A modified ResNet architecture: OSG-Net}

\subsubsection{ResNet}

\begin{figure}[ht!]
\def\sc{0.3}
\def\SC{0.6}
    \centering
    \begin{subfigure}[t]{\sc\textwidth}
	    \centering
        \includegraphics{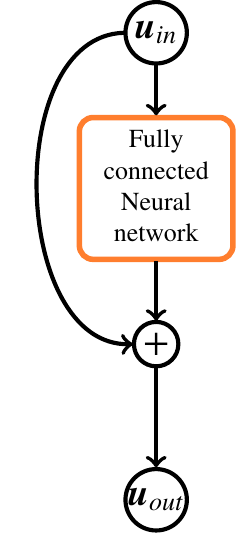} 
	    \caption{ResNet for trajectories with constant time step.}
	    \label{fig:resnet}
    \end{subfigure}\qquad
    \begin{subfigure}[t]{\sc\textwidth}
	    \centering
        \includegraphics{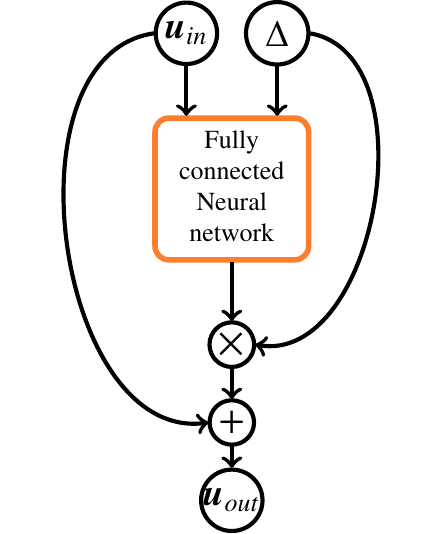} 
	    \caption{OSG-Net for trajectories with changing time step.}
	    \label{fig:resnet_t}
    \end{subfigure}
    
    \medskip
    \medskip
    
    \begin{subfigure}[t]{\SC\textwidth}
	    \centering
        \includegraphics[width=\linewidth]{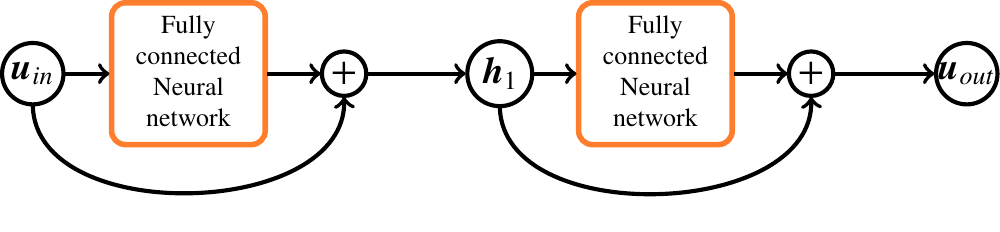} 
	    \caption{Recursive ResNets applicable to trajectories with fixed time step. Each block is a ResNet as described in Figure \ref{fig:resnet}. We set $K=2$ for illustration.}
	    \label{fig:rc_resnet}
    \end{subfigure}
    \begin{subfigure}[t]{\SC\textwidth}
	    \centering
        \includegraphics[width=\linewidth]{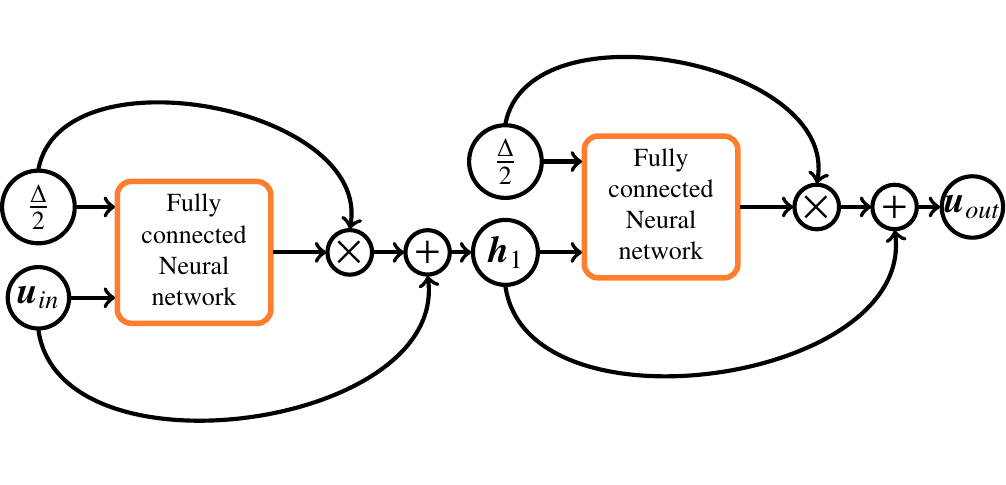} 
	    \caption{Recursive OSG-Net applicable to trajectories with changing time step. Each block is an OSG-Net described in Figure \ref{fig:resnet_t}. The time step $\Delta$ is divided by $K$ and fed to each block. We set $K=2$ for illustration.}
	    \label{fig:rc_resnet_t}
    \end{subfigure}
    \caption{\textbf{Residual neural networks (ResNets) for flow map learning}.}
\end{figure}

As shown in \cite{qin2019data,wu2020data}, the residual network (ResNet) is  particularly suitable for learning an evolution operator with a fixed time lag. 
The notion of ResNet \cite{he2016deep} is to explicitly introduce the identity mapping in the network and force the neural network to effectively learn the residue of the input-output mapping.  
The illustration of a one-step ResNet is provided in Fig.~\ref{fig:resnet}.  
Mathematically, 
a ResNet based on fully connected feedforward deep neural network may be expressed as 
\begin{equation}\label{eq:ResNet} 
	\bm{u}_{out} = {\bf N}_{\bm \theta} (\bm{u}_{in}) = \bm{u}_{in} + {\mathcal N}_{\bm{\theta}}(\bm{u}_{in}) = \left(  {\bf I}_n + {\mathcal N}_{\bm{\theta}} \right) ( \bm{u}_{in} ),
\end{equation}
where $\bm{u}_{in}$ denotes the network input, $\bm{u}_{out}$ is the network output,  $\bm{\theta}$ denotes all the trainable parameters in the neural network, and 
\begin{equation*}
	{\mathcal N}_{\bm{\theta}}(\bm{u}_{in}) = {\bf W}_{L+1} \circ ( \sigma_L \circ  {\bf W}_{L} ) \circ \cdots \circ ( \sigma_1 \circ  {\bf W}_1 ) (\bm{u}_{in})
\end{equation*}
denotes a fully connected feedforward neural network (FNN) of $L$ hidden layers, with ${\bf W}_j$ being the weight matrix between the $j$th layer and the $(j+1)$th layer, 
$\sigma_j$ denoting the activation function, and $\circ$ being the composition operator. 
Multiple ResNet blocks can be stacked recursively, providing a deeper recursive ResNet architecture (see Figure \ref{fig:rc_resnet}). 
Mathematically, a multi-step recursive ResNet structure can be formulated as  
\begin{equation*}
	{\bf N}_{\bm \theta} =  \left(  {\bf I}_n + {\mathcal N}_{{\bm \theta}_K} \right) \circ \cdots 
	\circ \left(  {\bf I}_n + {\mathcal N}_{{\bm \theta}_1} \right),
\end{equation*}
where $K$ is the number of ResNet blocks, and $\bm{\theta}_i$ is the network parameters in the $i$th block.  
The recursive ResNet enables the network to learn dynamics at smaller time scales than the time stepsize of the collected samples and is thus more advantageous than a single step ResNet for evolution operator learning \cite{qin2019data}. 

When the time lag $\Delta$ in the data is a constant, the ResNet structure \eqref{eq:ResNet} is particularly suitable for approximating the evolution operator ${\bf \Phi}_\Delta$. By comparing \eqref{equation:flow_map} with \eqref{eq:ResNet}, one can see that the FNN operator ${\mathcal N}_{\bm{\theta}}$ becomes an approximation to the effective increment 
$\int_{0}^\Delta {\bm f}(  {\bf \Phi}_s(\bm{u}) ) {\rm d} s$. 
However, the network architecture \eqref{eq:ResNet} does not include the time step as its input and thus is not applicable for data collected with varied time lags. 
Moreover, the effective increment
$ \int_{0}^\Delta {\bm f}(  {\bf \Phi}_s(\bm{u}) ) {\rm d} s = {\mathcal O}(\Delta)$ approaches zero as $\Delta \to 0$, which is, however, not accommodated by the ResNet structure \eqref{eq:ResNet}.

\subsubsection{OSG-Net}\label{sec:OSG-Net}

To address the above-mentioned limitations of ResNet, we propose a modified ResNet architecture 
\begin{equation}\label{eq:OSG-Net}
	\bm{u}_{out} = {\bf N}_{\bm \theta} (\bm{u}_{in}, \Delta) = \bm{u}_{in} + \Delta {\mathcal N}_{\bm{\theta}}(\bm{u}_{in}, \Delta),
\end{equation}
which is called operator semigroup network (OSG-Net) as it naturally incorporates 
the first semigroup property \eqref{equation:semigroup_property} of the evolution operator. 
The OSG-Net architecture \eqref{eq:OSG-Net} is illustrated in Figure \ref{fig:resnet_t}, where the initial state variables $\bm{u}_{in}$ and the evolution time $\Delta$ are concatenated before being fed into the network ${\mathcal N}_{\bm{\theta}}$. Compared to Figure \ref{fig:resnet}, we introduce a second skip connection which multiplies $\Delta$ with the output of ${\mathcal N}_{\bm{\theta}}$. Such a new design makes the whole architecture continuously reduce to an identity map as $\Delta \to 0$, thereby strictly enforcing the property (\ref{equation:semigroup_property1}). 
By comparing \eqref{equation:flow_map} with \eqref{eq:OSG-Net}, we see 
that the OSG-Net architecture matches the formula ${\bf \Phi}_\Delta(\bm{u}_0)  = 
\left[ {\bf I}_n + \Delta {\bm \phi} (\cdot, \Delta) \right] ( \bm{u}_0 )$, and the FNN operator ${\mathcal N}_{\bm{\theta}}$ becomes an approximation to the time-averaged effective increment ${\bm \phi} (\cdot, \Delta)$, namely, 
\begin{equation}\label{eq:appeff}
	{\mathcal N}_{\bm{\theta}}(\bm{u}, \Delta) \approx {\bm \phi} (\bm{u}, \Delta) =  \frac{1}{\Delta}\int_{0}^{\Delta} \bm{f}( {\bf \Phi}_s (\bm{u}) ){\rm d}s.
\end{equation}

Similar to the recursive ResNet in Figure \ref{fig:rc_resnet}, we can also propose deep recursive OSG-Net architectures. 
To do this, we devide the time lag $\Delta$ by $K$, with $K$ being the total number of blocks (see Figure \ref{fig:rc_resnet_t}). The time-step input of each OSG-Net block is $\Delta/K$. 
Mathematically, a multi-step recursive OSG-Net structure can be formulated as  
\begin{equation*}
	{\bf N}_{\theta} =  \left[  {\bf I}_n + \frac{\Delta}K {\mathcal N}_{{\bm \theta}_K} \bigg(\cdot, \frac{\Delta}K \bigg) \right] \circ \cdots 
	\circ \left[  {\bf I}_n + \frac{\Delta}K {\mathcal N}_{{\bm \theta}_1} \bigg( \cdot, \frac{\Delta}K \bigg) \right],
\end{equation*}
where $K$ is the number of OSG-Net blocks, and $\bm{\theta}_i$ is the network parameters in the $i$th block. 
In such a stacked architecture, each OSG-Net block is expected to learn dynamics at smaller time scales. A study on the number of blocks will be conducted in Section \ref{section:advection_eq}.

\begin{remark1}\label{rem1}
	The proposed OSG-Net architecture is very feasible: it can be combined with other suitable basic networks, such as convolution neural networks and locally connected networks, in addition to FNN. 
	In fact, such an architecture can also be combined with 
	other suitable approximators, e.g.~polynomials. 
\end{remark1}

\begin{remark1}[Multiscale stepsizes]\label{rem2}
	In the multiscale problems (such as the chemical reaction problem of Robertson, which will be simulated in Section \ref{ex:robertson}), the time stepsize $\Delta$ may vary from a small scale such as $10^{-5}$ to a large scale such as $1$ or $10$. 
	In such cases, we propose a slightly modified multiscale version of the OSG-Net architecture
	\begin{equation}\label{eq:multiscaleOSG-Net}
		\bm{u}_{out} = {\bf N}_{\bm \theta} (\bm{u}_{in}, \Delta) = \bm{u}_{in} + \Delta {\mathcal N}_{\bm{\theta}}(\bm{u}_{in}, -\log_{10}\Delta),
	\end{equation}
	which we find is quite effective to model the multiscale evolution structures of the underlying system over time; see Section \ref{ex:robertson}. 
\end{remark1}

Following the recurrent ResNet in \cite{qin2019data}, one may also design  
the multi-step OSG-Net with parameters shared among the blocks, termed {\em recurrent OSG-Net}. 
It seems more reasonable to enforce uniformly divided time steps $\Delta/K$ for the recurrent OSG-Net and to employ a non-uniformly divided time steps $\{\delta_1+\delta_2+\cdots+\delta_K=\Delta \}$ for the proposed recursive OSG-Net. 
However, it is challenging to determine a suitable non-uniform division of the time steps, as the dynamics in each OSG block is learned via training. Thus, we simply use the uniformly divided time steps in the recursive OSG-Net, but one can certainly use other non-uniform divisions. 
In \ref{section:share_parameters}, we present a comparison between the recursive and recurrent OSG-Nets, which indicates the advantage of the 
recursive OSG-Net. 

\subsection{Semigroup-informed learning}
\label{section:sg_loss}

The proposed OSG-Net is trained by minimizing {a} suitable loss function. 
In the purely data-driven case, the conventional loss function is 
\begin{equation}\label{eq:data-loss}
	L({\bm \theta})=\frac{1}{J}\sum_{j=1}^J L_{data,j} ( {\bm \theta} ) \quad \mbox{with} \quad L_{data,j} ( {\bm \theta} ) = \frac12 \left[ \ell \left({\bm{u}}_{1,j}, \widehat{{\bm{u}}}_{01,j}({\bm \theta}) \right) + \ell \left({\bm{u}}_{2,j}, \widehat{{\bm{u}}}_{12,j}({\bm \theta}) \right) \right],
\end{equation}
where 
\begin{equation*}
	\widehat{{\bm{u}}}_{01,j}({\bm \theta}):= {\bf N}_{\bm{\theta}}({\bm{u}}_{0,j},\Delta_{1,j}) \quad \mbox{and} \quad   \widehat{{\bm{u}}}_{12,j}({\bm \theta}):= {\bf N}_{\bm{\theta}}({\bm{u}}_{1,j},\Delta_{2,j})	
\end{equation*}
are two single-step predictions given by the OSG-Net parametrized by $\bm{\theta}$. 
In \eqref{eq:data-loss}, 
$\ell(\cdot,\cdot)$ is a ``loss metric'' which measures the {difference} between the network outputs and the data. A standard metric is based on the vector $l^2$ norm square, namely,   $\ell( {\bm u}, \hat{{\bm{u}}}) = \| {\bm u} - \hat{{\bm{u}}} \|_2^2$. Another option is the relative $l_2$ norm: $\ell ({\bm{u}}, \hat{{\bm{u}}})= \frac{\|{\bm{u}}-\hat{{\bm{u}}}\|_2}{\|{\bm{u}}\|_2}$ as suggested in \cite{li2021fourier}.

However, neural network model based on the purely data-driven loss function \eqref{eq:data-loss} does not embed the second semigroup property \eqref{equation:semigroup_property2}, which can be expressed as 
\begin{equation}\label{eq:SG}
	{\bf \Phi}_{\Delta_1 + \Delta_2} ({\bm u}_0) = {\bf \Phi}_{\Delta_2} \left( {\bf \Phi}_{\Delta_1} ({\bm u}_0) \right) = {\bf \Phi}_{\Delta_1} \left( {\bf \Phi}_{\Delta_2} ({\bm u}_0) \right) \qquad \forall \Delta_1,\Delta_2>0,\quad \forall {\bm u}_0 \in D,
\end{equation} 
where $D \subseteq \mathbb R^n$ is the domain of interest in the phase space. 

\begin{figure}[ht!]
    \centering
    \includegraphics[width=0.8\linewidth]{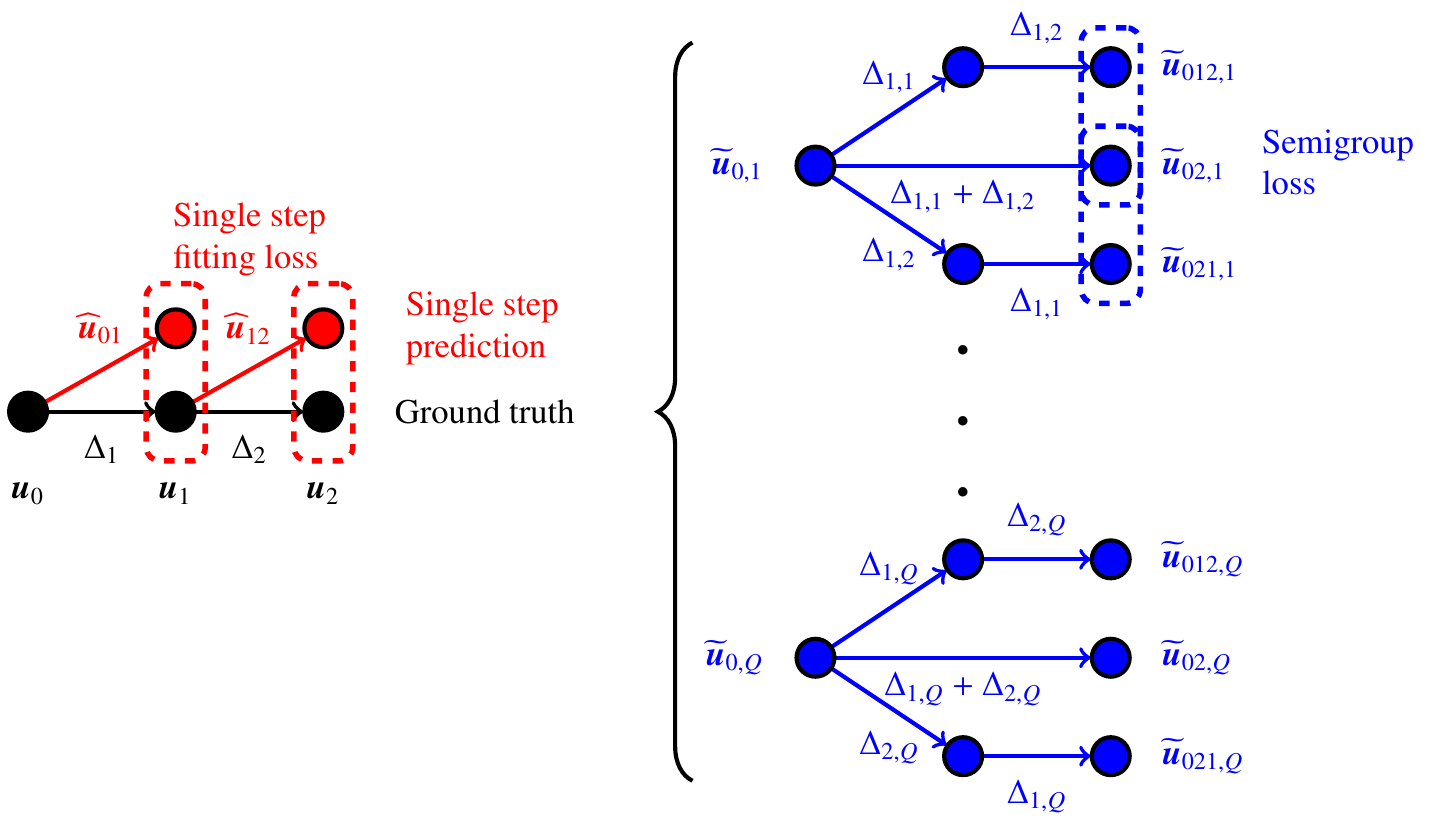} 
    \caption{\textbf{Semigroup-informed loss function}. As shown in equation (\ref{equation:methodC}), each observed trajectory (on the left) is grouped with $Q$ randomly generated tuples (on the right). The semigroup residues are calculated on the tuples, so no supplementary labeled data is required.}
    \label{fig:loss_randsg}
\end{figure}

Note that the identity \eqref{eq:SG} can be treated as a constraint not relying on labeled data. 
We propose a novel regularization method, which efficiently incorporates the semigroup property \eqref{eq:SG} into our OSG-Net model without requiring any extra measurement data. 
Specifically, 
we introduce the following new loss function 
\begin{equation}\label{equation:methodC}
	L({\bm \theta})=\frac{1}{J}\sum_{j=1}^J \left( \frac{ 1}{1+\lambda} 
	\left( L_{data,j} ( {\bm \theta} ) + \lambda 
	L_{SG,j} ( {\bm \theta} )  \right) \right)
\end{equation}
where {$\lambda>0$} is a regularization factor. The semigroup-informed loss function is defined as 
\begin{equation}\label{eq:SG0}
	L_{SG,j} ( {\bm \theta} ) :=	\frac{1}{2Q}\sum_{i=1}^Q \left[ \ell \left(\widetilde{{\bm{u}}}_{02,ij} ({\bm \theta}), \widetilde{{\bm{u}}}_{012,ij} ({\bm \theta}) \right) + \ell \left(\widetilde{{\bm{u}}}_{02,ij} ({\bm \theta}), \widetilde{{\bm{u}}}_{021,ij} ({\bm \theta}) \right) \right], 
\end{equation}
where  
\begin{align*}
	\widetilde{{\bm{u}}}_{02,ij} & = {\bf N}_{\bm{\theta}} \left(\widetilde{\bm{u}}_{0,ij}, \Delta_{1,ij}+\Delta_{2,ij} \right),\\
	\widetilde{{\bm{u}}}_{012,ij} &= {\bf N}_{\bm{\theta}} \left({\bf N}_{\bm{\theta}} \left(\widetilde{\bm{u}}_{0,ij}, \Delta_{1,ij}\right), \Delta_{2,ij}\right),\\
	\widetilde{{\bm{u}}}_{021,ij} &= {\bf N}_{\bm{\theta}}\left({\bf N}_{\bm{\theta}} \left(\widetilde{\bm{u}}_{0,ij}, \Delta_{2,ij} \right), \Delta_{1,ij} \right)
\end{align*}
represent the prediction of the state at time $t=t_0+\Delta_{1,ij}+\Delta_{2,ij}$ via three different forward passages{.  The} ``initial'' state $\widetilde{\bm{u}}_{0,ij}$ is randomly sampled in $D$, and $\{\Delta_{1,ij}, \Delta_{2,ij}\}$ are two randomly selected time stepsizes. See Figure~\ref{fig:loss_randsg} for an illustration of the semigroup loss function \eqref{eq:SG0}. 
By minimizing (\ref{equation:methodC}), we expect that the trained OSG-Net model successfully embeds the semigroup property \eqref{equation:semigroup_property2}. 

\begin{remark1}
	Since the stepsizes $\Delta_{1,ij}$ and $\Delta_{2,ij}$ are random, it 
	seems unnecessary to simultaneously include both $\ell \left(\widetilde{{\bm{u}}}_{02,ij} ({\bm \theta}), \widetilde{{\bm{u}}}_{012,ij} ({\bm \theta}) \right)$ and 
	$\ell \left(\widetilde{{\bm{u}}}_{02,ij} ({\bm \theta}), \widetilde{{\bm{u}}}_{021,ij} ({\bm \theta}) \right)$ in the loss function \eqref{eq:SG0}. However, our numerical results demonstrate  
	that including both of them is beneficial for training the network to enforce the semigroup property  \eqref{equation:semigroup_property2}. 
\end{remark1}

Besides the semigroup-informed loss function \eqref{eq:SG0}, we also introduce another ``indirect'' semigroup-embedding approach, which implicitly enhances the semigroup property  \eqref{equation:semigroup_property2}, based on a purely data-driven two-step loss function
\begin{equation}\label{equation:methodB}
	L({\bm \theta})=\frac{1}{J}\sum_{j=1}^J L_{ISG,j} ( {\bm \theta} ) 
\end{equation}
by adding three terms to $L_{data,j}$ as follows:
\begin{equation*}
	L_{ISG,j} ( {\bm \theta} ) = \frac15 \left[ \ell \left({\bm{u}}_{1,j}, \widehat{{\bm{u}}}_{01,j}({\bm \theta}) \right) + \ell \left({\bm{u}}_{2,j}, \widehat{{\bm{u}}}_{12,j}({\bm \theta}) \right) +  \ell \left({\bm{u}}_{2,j},\widehat{{\bm{u}}}_{02,j} ({\bm \theta})  \right) + \ell \left({\bm{u}}_{2,j},\widehat{{\bm{u}}}_{012,j} ({\bm \theta})  \right) + \ell \left({\bm{u}}_{2,j},\widehat{{\bm{u}}}_{021,j} ({\bm \theta})  \right) \right],  
\end{equation*}
where 
\begin{align*}
	\widehat{{\bm{u}}}_{02,j} ({\bm \theta})  & = {\bf N}_{\bm{\theta}}({\bm{u}}_{0,j}, \Delta_{1,j}+\Delta_{2,j}),\\
	\widehat{{\bm{u}}}_{012,j} ({\bm \theta})  & = {\bf N}_{\bm{\theta}}\big({\bf N}_{\bm{\theta}}({\bm{u}}_{0,j}, \Delta_{1,j}), \Delta_{2,j}\big),\\
	\widehat{{\bm{u}}}_{021,j} ({\bm \theta})  & = {\bf N}_{\bm{\theta}}\big({\bf N}_{\bm{\theta}}({\bm{u}}_{0,j}, \Delta_{2,j}), \Delta_{1,j}\big)
\end{align*}
are three different approximations of the state ${\bm{u}}_{2,j}$. By minimizing \eqref{equation:methodB}, we expect that the trained network model satisfies 
$L_{ISG,j} ( {\bm \theta} ) \le \epsilon$ for some small $\epsilon>0$ so that 
\begin{align*}
\begin{aligned}
		\left\| {\bf N}_{\bm{\theta}}\big({\bf N}_{\bm{\theta}}({\bm{u}}_{0,j}, \Delta_{1,j}), \Delta_{2,j}\big) - {\bf N}_{\bm{\theta}}({\bm{u}}_{0,j}, \Delta_{1,j}+\Delta_{2,j}) \right\|_2 &=  \left\| \widehat{{\bm{u}}}_{012,j} ({\bm \theta}) - \widehat{{\bm{u}}}_{02,j} ({\bm \theta}) \right\|_2 
	\\
	& \le   \left\| {\bm{u}}_{2,j} - \widehat{{\bm{u}}}_{012,j} ({\bm \theta}) \right\|_2 + 
	\left\| {\bm{u}}_{2,j} - \widehat{{\bm{u}}}_{02,j} ({\bm \theta}) \right\|_2 
	\le 2 \sqrt{5 \epsilon},
\end{aligned}
	\\
\begin{aligned}
		\left\| {\bf N}_{\bm{\theta}}\big({\bf N}_{\bm{\theta}}({\bm{u}}_{0,j}, \Delta_{2,j}), \Delta_{1,j}\big) - {\bf N}_{\bm{\theta}}({\bm{u}}_{0,j}, \Delta_{1,j}+\Delta_{2,j}) \right\|_2 &=  \left\| \widehat{{\bm{u}}}_{021,j} ({\bm \theta}) - \widehat{{\bm{u}}}_{02,j} ({\bm \theta}) \right\|_2 
	\\
	& \le   \left\| {\bm{u}}_{2,j} - \widehat{{\bm{u}}}_{021,j} ({\bm \theta}) \right\|_2 + 
	\left\| {\bm{u}}_{2,j} - \widehat{{\bm{u}}}_{02,j} ({\bm \theta}) \right\|_2 
	\le 2 \sqrt{5 \epsilon},
\end{aligned}
\end{align*} 
if the mean squared loss metric is used. Therefore, the two-step loss function (\ref{equation:methodB}) implicitly enforces the semigroup property \eqref{equation:semigroup_property2} along the {sampled} trajectories. 

\begin{remark1} For clarity, 
	we refer to the conventional  
	data-driven approach \eqref{eq:data-loss} as the {\em baseline method}, 
	we refer to our main approach based on \eqref{equation:methodC} as the {\em global direct  semigroup-informed (GDSG) method}, and we refer to the approach based on \eqref{equation:methodB} as the {\em local indirect semigroup-informed (LISG) method}. This is because the GDSG method directly enforces the semigroup property \eqref{equation:semigroup_property2} globally over the random samples widely  distributed in the entire domain $D$, while the LISG method indirectly enhances the semigroup property  \eqref{equation:semigroup_property2} locally only along the {sampled} trajectories.      
\end{remark1}

\begin{remark1}[Limited data] 
	When adequate amount of data is available, the neural network model can be effectively trained to learn the evolution operator \cite{qin2019data,wu2020data}.  However, in practice, collecting measurement data from real-world systems can be very costly or 
	difficult 
	due to resource 
	constraints or limited experimental accessibility. 
	When the measurement data \eqref{eq:data} is very limited, embedding the semigroup constraint \eqref{equation:semigroup_property2} becomes very helpful or even essential to avoid over-fitting, as the semigroup-informed loss function plays an important role of regularization. It not only significantly reduces the required data and improves the deep learning accuracy but also greatly enhances the robustness and stability of long-time prediction; see Section \ref{section:egs}. 
\end{remark1}

\subsection{Model prediction}
Once the OSG-Net model \eqref{eq:OSG-Net} is successfully trained, we can recursively employ the model to conduct prediction for any arbitrarily given initial state ${\bm u}(t_0)$ as follows 
\begin{equation}\label{eq:upre}
	\begin{cases}
		{\bm u}^{pre}(t_0)={\bm u}(t_0)
		\\
	{\bm u}^{pre}( t_{j+1} ) =
	{\bm u}^{pre}( t_{j} ) + \Delta_j {\mathcal N}_{\bm{\theta}}( {\bm u}^{pre}( t_{j} ), \Delta_j), \qquad j=0,1,\dots 
	\end{cases} 
\end{equation}
where $\Delta_j=t_{j+1}-t_{j}$. 

	Even though the form of the prediction model \eqref{eq:upre} resembles the first-order forward Euler method for time stepping, the model \eqref{eq:upre} is indeed an approximation of the exact (rather than first-order) time integrator, because there is no explicit discretization error in time. The only source of error in the prediction model \eqref{eq:upre} is the approximation error of the time-averaged effective increment \eqref{eq:appeff}. 

In order to estimate a bound for the prediction error, we first introduce the following lemma. 

\begin{lemma}\label{lem1}
	Suppose $\bm f$ is Lipschitz continuous with Lipschitz constant $L_f$ on a set $D \subseteq \mathbb{R}^n$. For any $T >0$, define 
	\begin{equation*}
		D_T =\big\{ {\bm u} \in D:~{\bf \Phi}_\Delta(\bm{u}) \in D~\forall \Delta \in [0,T]  \big\}.
	\end{equation*} 
	Then we have
	\begin{equation}\label{eq:phiLp}
		\big\| {\bm \phi} ({\bm u}_1,\Delta) - {\bm \phi} ({\bm u}_2,\Delta) \big \|_2 
		\le \frac{ {\rm e}^{ L_f \Delta } - 1 }{ \Delta }  \big\| {\bm u}_1 - {\bm u}_2 \big\|_2 \qquad \forall {\bm u}_1, {\bm u}_2 \in D_T\quad \forall \Delta \in [0,T]. 
	\end{equation}
\end{lemma}

\begin{proof}
	Note that ${\bf \Phi}_t (u_1) = {\bm u}_1 + t {\bm \phi} ({\bm u}_1,t)$ is the solution to the ODE $\frac{d {\bm u}}{dt}={\bm f}({\bm u})$ with initial data ${\bm u}_1$. Thus, we have 
	\begin{equation}\label{eq:3122}
		 \frac{ d }{dt} \Big( t {\bm \phi}({\bm u}_1,t) \Big) 
		= {\bm f} \big( {\bf \Phi}_t (u_1) \big).
	\end{equation}
	Similarly, we obtain 
		\begin{equation}\label{eq:3123}
		\frac{ d }{dt} \Big( t {\bm \phi}({\bm u}_2,t) \Big) 
		= {\bm f} \big( {\bf \Phi}_t (u_2) \big).
	\end{equation}
	Combining \eqref{eq:3122} with \eqref{eq:3123} gives 
		\begin{equation}\label{eq:3124}
		\frac{ d }{dt} \Big( t ( {\bm \phi}({\bm u}_1,t) - {\bm \phi}({\bm u}_2,t) ) \Big) 
		= {\bm f} \big( {\bf \Phi}_t (u_1) \big) - {\bm f} \big( {\bf \Phi}_t (u_2) \big).
	\end{equation}
It follows that 
		\begin{align*}
 \frac12	\frac{ d }{dt} \Big( t^2 \big\| {\bm \phi}({\bm u}_1,t) - {\bm \phi}({\bm u}_2,t) \big\|^2_2 \Big) 
 &= \Big( t ( {\bm \phi}({\bm u}_1,t) - {\bm \phi}({\bm u}_2,t) ) \Big) \cdot \frac{ d }{dt} \Big( t ( {\bm \phi}({\bm u}_1,t) - {\bm \phi}({\bm u}_2,t) ) \Big) 
 \\
&	=  t ( {\bm \phi}({\bm u}_1,t) - {\bm \phi}({\bm u}_2,t) ) \cdot \Big(  {\bm f} \big( {\bf \Phi}_t (u_1) \big) - {\bm f} \big( {\bf \Phi}_t (u_2) \big) \Big)
\\
& \le t  \big\| {\bm \phi}({\bm u}_1,t) - {\bm \phi}({\bm u}_2,t) \big\|_2 
\big\| {\bm f} \big( {\bf \Phi}_t (u_1) \big) - {\bm f} \big( {\bf \Phi}_t (u_2) \big) \big\|_2
\\
& \le t  \big\| {\bm \phi}({\bm u}_1,t) - {\bm \phi}({\bm u}_2,t) \big\|_2 
L_f \big\|  {\bf \Phi}_t (u_1)  -  {\bf \Phi}_t (u_2)  \big\|_2.
\end{align*}
Therefore, 
\begin{align*}
	\frac{ d }{dt} \Big( t \big\| {\bm \phi}({\bm u}_1,t) - {\bm \phi}({\bm u}_2,t) \big\|_2 \Big) &\le L_f \big\|  {\bf \Phi}_t (u_1)  -  {\bf \Phi}_t (u_2)  \big\|_2 
	\\
	&= L_f \big\| {\bm u}_1 + t {\bm \phi} ({\bm u}_1,t) - {\bm u}_2 - t {\bm \phi} ({\bm u}_2,t) \big\|_2 
	\\
	&\le L_f \| {\bm u}_1 - {\bm u}_2 \| + L_f t \big\| {\bm \phi}({\bm u}_1,t) - {\bm \phi}({\bm u}_2,t) \big\|_2 \qquad  \forall t \in [0,T].
\end{align*}
Applying the Gronwall inequality yields 
	\begin{equation}\label{eq:phiLpt}
	t \big\| {\bm \phi} ({\bm u}_1,t) - {\bm \phi} ({\bm u}_2,t) \big \|_2 
	\le \left( {\rm e}^{ L_f t} - 1 \right) \big\| {\bm u}_1 - {\bm u}_2 \big\|_2,
\end{equation}
which gives \eqref{eq:phiLp} by taking $t=\Delta$ and then dividing it by $\Delta$. The proof is completed. 
\end{proof}

We are now in a position to derive an error bound for the prediction model \eqref{eq:upre}. Since it is well known that neural networks are universal approximator for a general class of functions, we will assume that the error of the approximation \eqref{eq:appeff} is bounded and small. 

\begin{theorem}
	Suppose $\bm f$ is Lipschitz continuous with Lipschitz constant $L_f$ on a set $D \subseteq \mathbb{R}^n$. Assume that 
	${\bm u}^{pre}( t_{j} )\in D_{\Delta_j}$ and ${\bm u}( t_{j} ) \in D_{\Delta_j}$ for all $0 \le j < m$, and suppose that 
	the generalization error of the approximation \eqref{eq:appeff} for the trained neural network model is bounded, namely, 
	\begin{equation*}
		\big\| {\mathcal N}_{\bm{\theta}} - {\bm \phi}  \big\|_{L^\infty( D \times[\Delta_{\min},\Delta_{\max}] )} < +\infty, 
	\end{equation*}
	where $\Delta_{\min} = \min_j \Delta_j$ and $\Delta_{\max} = \max_j \Delta_j$. Then we have 
	\begin{equation*}
		\big\| {\bm u}^{pre}( t_{j} ) - {\bm u}( t_{j} ) \big\|_2 \le \big\| {\mathcal N}_{\bm{\theta}} - {\bm \phi}  \big\|_{L^\infty( D \times[\Delta_{\min},\Delta_{\max}] )} \sum_{s=1}^{j} \Delta_s  {\rm e}^{L_f (t_{j}- t_{s}) } \qquad j=1,\dots,m.
	\end{equation*}
In particular, if $\Delta_j \equiv \Delta$ for all $0 \le j < m$, then we have 
\begin{equation*}
	\big\| {\bm u}^{pre}( t_{j} ) - {\bm u}( t_{j} ) \big\|_2 \le \big\| {\mathcal N}_{\bm{\theta}} - {\bm \phi}  \big\|_{L^\infty( D \times[\Delta_{\min},\Delta_{\max}] )} \frac{ \Delta ( 1 - {\rm e}^{ L_f j \Delta} ) }{ 1- {\rm e}^{L_f \Delta} }.
\end{equation*}
\end{theorem}

\begin{proof}
Let ${\mathcal E}_j = \| {\bm u}^{pre}( t_{j} ) - {\bm u}( t_{j} ) \|_2$, $j=0,1,\dots,m$, denote the prediction errors.  Note that 
\begin{equation*}
	{\bm u}( t_{j+1} ) = {\bm u}( t_{j} )+ \Delta_j {\bm \phi}( {\bm u}( t_{j} ), \Delta_j  ).
\end{equation*}
This together with the second equation in \eqref{eq:upre} yields 
\begin{align*}
	{\mathcal E}_{j+1} &= \big\| {\bm u}^{pre}( t_{j+1} ) - {\bm u}( t_{j+1} ) \big\|_2 
	\\
	&=  \big\| {\bm u}^{pre}( t_{j} ) + \Delta_j {\mathcal N}_{\bm{\theta}}( {\bm u}^{pre}( t_{j} ), \Delta_j) - {\bm u}( t_{j} ) - \Delta_j {\bm \phi}( {\bm u}( t_{j} ), \Delta_j  )   \big\|_2 
	\\
	& \le   \big\| {\bm u}^{pre}( t_{j} )  - {\bm u}( t_{j} )   \big\|_2  
	+ \Delta_j \big\| {\mathcal N}_{\bm{\theta}}( {\bm u}^{pre}( t_{j} ), \Delta_j) -  {\bm \phi}( {\bm u}( t_{j} ), \Delta_j  )   \big\|_2 
	\\
	& \le {\mathcal E}_j +  \Delta_j \left(   
	 \big\| {\mathcal N}_{\bm{\theta}}( {\bm u}^{pre}( t_{j} ), \Delta_j) -  {\bm \phi} ( {\bm u}^{pre}( t_{j} ), \Delta_j)   \big\|_2
	 + \big\| {\bm \phi} ( {\bm u}^{pre}( t_{j} ), \Delta_j)  -  {\bm \phi}( {\bm u}( t_{j} ), \Delta_j  )   \big\|_2
	\right),
\end{align*}
where we have used the triangle inequalities. Using Lemma \ref{lem1} gives 
\begin{align}\label{key}
	{\mathcal E}_{j+1} &\le {\mathcal E}_j +  \Delta_j \left(   
	\big\| {\mathcal N}_{\bm{\theta}}( {\bm u}^{pre}( t_{j} ), \Delta_j) -  {\bm \phi} ( {\bm u}^{pre}( t_{j} ), \Delta_j)   \big\|_2
	+ \frac{ {\rm e}^{ L_f \Delta_j } - 1 }{ \Delta_j } \big\|  {\bm u}^{pre}( t_{j} ) - {\bm u}( t_{j} )   \big\|_2
	\right)
	\\
	& = {\rm e}^{ L_f \Delta_j } {\mathcal E}_j  + \Delta_j  \big\| {\mathcal N}_{\bm{\theta}}( {\bm u}^{pre}( t_{j} ), \Delta_j) -  {\bm \phi} ( {\bm u}^{pre}( t_{j} ), \Delta_j)   \big\|_2
	\\
	& \le {\rm e}^{ L_f \Delta_j } {\mathcal E}_j  + \Delta_j   \big\| {\mathcal N}_{\bm{\theta}} - {\bm \phi}  \big\|_{L^\infty( D \times[\Delta_{\min},\Delta_{\max}] )}.
\end{align}
It follows that 
\begin{align*}
	{\rm e}^{-L_f t_{j+1}}{\mathcal E}_{j+1} &\le {\rm e}^{-L_f t_{j}}{\mathcal E}_{j}
+ \Delta_j  {\rm e}^{-L_f t_{j+1}}     \big\| {\mathcal N}_{\bm{\theta}} - {\bm \phi}  \big\|_{L^\infty( D \times[\Delta_{\min},\Delta_{\max}] )}
\\
& \le {\rm e}^{-L_f t_{j-1}}{\mathcal E}_{j-1}
+ \left(  \Delta_j  {\rm e}^{-L_f t_{j+1}}  + \Delta_{j-1}  {\rm e}^{-L_f t_{j}}   \right) \big\| {\mathcal N}_{\bm{\theta}} - {\bm \phi}  \big\|_{L^\infty( D \times[\Delta_{\min},\Delta_{\max}] )}
\\
& \le \cdots
\\
& \le {\rm e}^{-L_f t_{0}}{\mathcal E}_{0}
+ \left(  \Delta_j  {\rm e}^{-L_f t_{j+1}}  + \Delta_{j-1}  {\rm e}^{-L_f t_{j}} + \cdots 
+\Delta_{0}  {\rm e}^{-L_f t_{1}} 
  \right) \big\| {\mathcal N}_{\bm{\theta}} - {\bm \phi}  \big\|_{L^\infty( D \times[\Delta_{\min},\Delta_{\max}] )}
  \\
  & = \big\| {\mathcal N}_{\bm{\theta}} - {\bm \phi}  \big\|_{L^\infty( D \times[\Delta_{\min},\Delta_{\max}] )} \sum_{s=0}^{j} \Delta_s  {\rm e}^{-L_f t_{s+1}},  
\end{align*}
where ${\mathcal E}_{0}=0$ has been used. 
Therefore, we have 
\begin{equation*}
	{\mathcal E}_{j+1} \le \big\| {\mathcal N}_{\bm{\theta}} - {\bm \phi}  \big\|_{L^\infty( D \times[\Delta_{\min},\Delta_{\max}] )} \sum_{s=0}^{j} \Delta_s  {\rm e}^{L_f (t_{j+1} -t_{s+1}) }
	= \big\| {\mathcal N}_{\bm{\theta}} - {\bm \phi}  \big\|_{L^\infty( D \times[\Delta_{\min},\Delta_{\max}] )} \sum_{s=1}^{j+1} \Delta_s  {\rm e}^{L_f (t_{j+1} -t_{s}) }.
\end{equation*}
If $\Delta_j \equiv \Delta$ for all $j$, then 
\begin{equation*}
	\sum_{s=1}^{j} \Delta_s  {\rm e}^{L_f (t_{j} -t_{s}) }  = \Delta\sum_{s=1}^{j}   {\rm e}^{L_f (j-s)\Delta } = \frac{ \Delta ( 1 - {\rm e}^{ L_f j   \Delta} ) }{ 1- {\rm e}^{L_f \Delta} }.
\end{equation*}
The proof is completed. 
\end{proof}

\subsection{Robustness and variance of prediction}

Let $\Delta_{1}+ \Delta_{2} = T$ and $\Delta_{1}'+ \Delta_{2}' = T$ represent two partitions of the interval $[0,T]$. 
By incorporating the semigroup property \eqref{eq:SG} into the learning process, the trained OSG-Net model satisfies  
\begin{equation*}
	{\bf N}_{\bm{\theta}} \left({\bf N}_{\bm{\theta}} \left({\bm{u}}_{0}, \Delta_{1}\right), \Delta_{2}\right) 
	 \approx {\bf N}_{\bm{\theta}} \left({\bm{u}}_{0}, T \right)
	 \approx {\bf N}_{\bm{\theta}}\left({\bf N}_{\bm{\theta}} \left({\bm{u}}_{0}, \Delta_{1}' \right), \Delta_{2}' \right),
\end{equation*}
which implies that 
\begin{equation*}
	\big\| {\bf N}_{\bm{\theta}} \left({\bf N}_{\bm{\theta}} \left({\bm{u}}_{0}, \Delta_{1}\right), \Delta_{2}\right) - {\bf N}_{\bm{\theta}}\left({\bf N}_{\bm{\theta}} \left({\bm{u}}_{0}, \Delta_{1}' \right), \Delta_{2}' \right)  \big\|_2 
\end{equation*}
is small. 
In general, let 
\begin{equation*}
	\left\{\Delta_j^{(k)} >0: \Delta_0^{(k)}+\Delta_1^{(k)}+...+\Delta_n^{(k)}=T \right\}, \qquad k=1,\dots,K 
\end{equation*}
denote $K$ different partitions of $[0,T]$. It is expected that 
\begin{equation}\label{eq:partions}
	\left\|
	{\bf N}_{\bm{\theta}} \left( \cdots 
	 {\bf N}_{\bm{\theta}} \left({\bf N}_{\bm{\theta}} \left({\bm{u}}_{0}, \Delta_0^{(k)} \right), \Delta_1^{(k)} \right) 
	\cdots \Delta_n^{(k)} \right)
	 - {\bf N}_{\bm{\theta}} \left( \cdots 
	 {\bf N}_{\bm{\theta}} \left({\bf N}_{\bm{\theta}} \left({\bm{u}}_{0}, \Delta_0^{(s)} \right), \Delta_1^{(s)} \right) 
	 \cdots \Delta_n^{(s)} \right) \right\|_2 \le \epsilon \quad \forall k,s \in \{1,\dots,K\} 
\end{equation}
for some small number $\epsilon >0$, 
which means the prediction results of the OSG-Net model are not sensitive to the time partition steps.   Define 
\begin{equation}\label{eq:upren}
	\begin{cases}
		{\bm u}^{pre}_k(0)={\bm u}_0
		\\
		{\bm u}^{pre}_k( t_{j+1}^{(k)} ) = {\bf N}_{\bm{\theta}}( {\bm u}^{pre}_k( t_j^{(k)} ), 
		\Delta_j^{(k)}), \qquad j=0,1,\dots,n 
	\end{cases} 
\end{equation}
as the predicted states through different partitions,  
where $t_{j+1}^{(k)} = t_j^{(k)}  + \Delta_j^{(k)} $ with $t_0^{(k)}=0$ and $t_{n+1}^{(k)}=T$ for all $1\le k \le K$.  

\begin{theorem}\label{thm:variance}
	Let ${\mathcal E}^{(k)}_T := \left\| {\bm u}^{pre}_k( T ) - u(T) \right \|_2$ be the prediction error by the trained OSG-Net model via the $k$th temporal partition, where $u(T)$ denotes the true solution at time $t=T$. Under the assumption \eqref{eq:partions}, we have the following estimate for the standard deviation of the prediction errors: 
	\begin{equation}\label{eq:variance}
		\sqrt{\frac{1}{K} \sum_{k=1}^K \left( {\mathcal E}^{(k)}_T - \frac{1}{K} \sum_{s=1}^K {\mathcal E}^{(s)}_T \right)^2} \le \left( \frac{K-1}{K} \right) \epsilon.
	\end{equation} 
\end{theorem}

\begin{proof}
	Based on \eqref{eq:partions}, we have 
\begin{equation*}
		\left\| {\bm u}^{pre}_k( T ) - {\bm u}^{pre}_s( T )  \right\| \le \epsilon  \qquad \forall k,s \in \{1,\dots,K\}.
\end{equation*}
	The triangle inequality implies that 
\begin{equation*}
		{\mathcal E}^{(k)}_T := \left\| {\bm u}^{pre}_k( T ) - u(T) \right \|_2 \le   \left\| {\bm u}^{pre}_s( T ) - u(T) \right \|_2 + \left\| {\bm u}^{pre}_k( T ) - {\bm u}^{pre}_s( T ) \right \|_2
	= {\mathcal E}^{(s)}_T + \epsilon \qquad \forall k,s \in \{1,\dots,K\}.
\end{equation*}
	It follows that 
\begin{equation*}
		(K-1) {\mathcal E}^{(k)}_T \le   \sum_{s \neq k} \left(  {\mathcal E}^{(s)}_T + \epsilon \right) 
	= (K-1) \epsilon + \sum_{s=1}^K   {\mathcal E}^{(s)}_T 
	- {\mathcal E}^{(k)}_T.
\end{equation*}
	Thus we have 
\begin{equation}\label{eq:2122}
	{\mathcal E}^{(k)}_T \le \frac{K-1}K \epsilon + \frac{1}{K} \sum_{s=1}^K   {\mathcal E}^{(s)}_T, \qquad 1\le k \le K.
\end{equation}
	Similar, one can derive that 
\begin{equation*}
		{\mathcal E}^{(k)}_T := \left\| {\bm u}^{pre}_k( T ) - u(T) \right \|_2 \ge   \left\| {\bm u}^{pre}_s( T ) - u(T) \right \|_2 -\left\| {\bm u}^{pre}_k( T ) - {\bm u}^{pre}_s( T ) \right \|_2
	= {\mathcal E}^{(s)}_T - \epsilon \qquad \forall k,s \in \{1,\dots,K\}.
\end{equation*}
	which yields 		
\begin{equation}\label{eq:2123}
		{\mathcal E}^{(k)}_T \ge -\frac{K-1}K \epsilon + \frac{1}{K} \sum_{s=1}^K   {\mathcal E}^{(s)}_T,  \qquad 1\le k \le K.
\end{equation}
	Combining \eqref{eq:2122} with \eqref{eq:2123} gives 
\begin{equation*}
		\left| {\mathcal E}^{(k)}_T - \frac{1}{K} \sum_{s=1}^K {\mathcal E}^{(s)}_T \right| \le \frac{K-1}K \epsilon \qquad \forall k\in \{1,\dots,K\}.
\end{equation*}
	Therefore, we obtain 
	\begin{equation*}
		\frac{1}{K} \sum_{k=1}^K \left( {\mathcal E}^{(k)}_T - \frac{1}{K} \sum_{s=1}^K {\mathcal E}^{(s)}_T \right)^2 \le \frac{1}{K} \sum_{k=1}^K  \left( \frac{K-1}{K} \right)^2 \epsilon^2 = \left( \frac{K-1}{K} \right)^2 \epsilon^2,
	\end{equation*}
which leads to \eqref{eq:variance}. The proof is completed. 
\end{proof}

\begin{remark1}[Robustness and long-time stability]
	Theorem \ref{thm:variance} shows the variance of the prediction errors via different partitions is small, indicating that the predicted results of the OSG-Net model are well self-matched and consistent for different time stepsizes. This property is highly desirable for robust prediction. A conventional model without embedding the semigroup structure  may deviate from this property due to the error accumulation at each forward step and thus suffers from poor performance in long-time prediction, as the error often grows in an exponential manner when the model is recursively composed. Our numerical experiments in Section \ref{section:egs} will further demonstrate our OSG-Net model (especially the main GDSG approach) is very stable in long-time prediction and is much more robust than the baseline approach; see, for example,  Figures \ref{fig:err_linear} and \ref{fig:err_ns}. 
\end{remark1}

\begin{remark1}
	The predictive OSG-Net model \eqref{eq:upre} resembles an explicit algorithm for evolution prediction. However, 
	the time stepsize of our model can be set very large and would not suffer from 
	a standard time step restriction, which is typically required by traditional explicit numerical schemes. This is true even for some highly stiff problems (see the examples in Section \ref{section:egs}). 
\end{remark1}

\section{Extension of Deep-OSG approach to modeling unknown PDEs}\label{sec:PDE}

Motivated by the previous {FML} works  \cite{wu2020data,chen2022deep}, 
our Deep-OSG framework can be easily extended to data-driven modeling of unknown PDEs. 
 Consider an unknown time-dependent PDE:
\begin{equation}
	\begin{cases}
		\partial_t {u}  = \mathcal{L}(u), & (x,t)\in \Omega\times \mathbb{R}^+,\\
		\mathcal{B}(u) = 0, & (x,t)\in \partial\Omega\times \mathbb{R}^+,\\
		u(x,0)   = u_0(x), & x\in \bar{\Omega},
	\end{cases}
	\label{equation:eg_vbg}
\end{equation}
\noindent where $\Omega\subseteq \mathbb{R}^d$ is the physical domain, and $\mathcal{L}$ and $\mathcal{B}$  respectively represent the operators in the equations and boundary conditions. We assume that the operator $\mathcal{L}$ is unknown, and we focus on learning PDE in the interior of the domain with given boundary conditions.

As our attention is restricted to autonomous PDEs, the unknown governing equations admit an evolution operator 
\begin{equation*}
	{\bf \Phi}_\Delta: \mathbb V \to \mathbb V, \qquad {\bf \Phi}_\Delta u(\cdot,t) = u(\cdot, t+\Delta),
\end{equation*}
where $\mathbb V$ is assumed to be an infinite-dimensional Hilbert space. Note that only the time difference $\Delta$ is relevant for the evolution operator of autonomous systems, as the time variable $t$ can be shifted arbitrarily. 
Assume that the solution $u(x,t)$ is measurable or observable, namely, the snapshots of $u$ are available at certain time instances: 
\begin{equation}\label{eq:PDEdata}
	 u(x,t_k^{(i)}), \qquad k=1,\dots,K_i,  \qquad i=1,\cdots, I_{traj},
\end{equation}
where $i$ denotes the $i$-th ``trajectory'' along which $K_i$ snapshot data are collected. Our goal is to learn the evolution operator of the underlying PDE based on the snapshot data.

\begin{figure}[h!]
	\centering
	\includegraphics[width=0.88\linewidth]{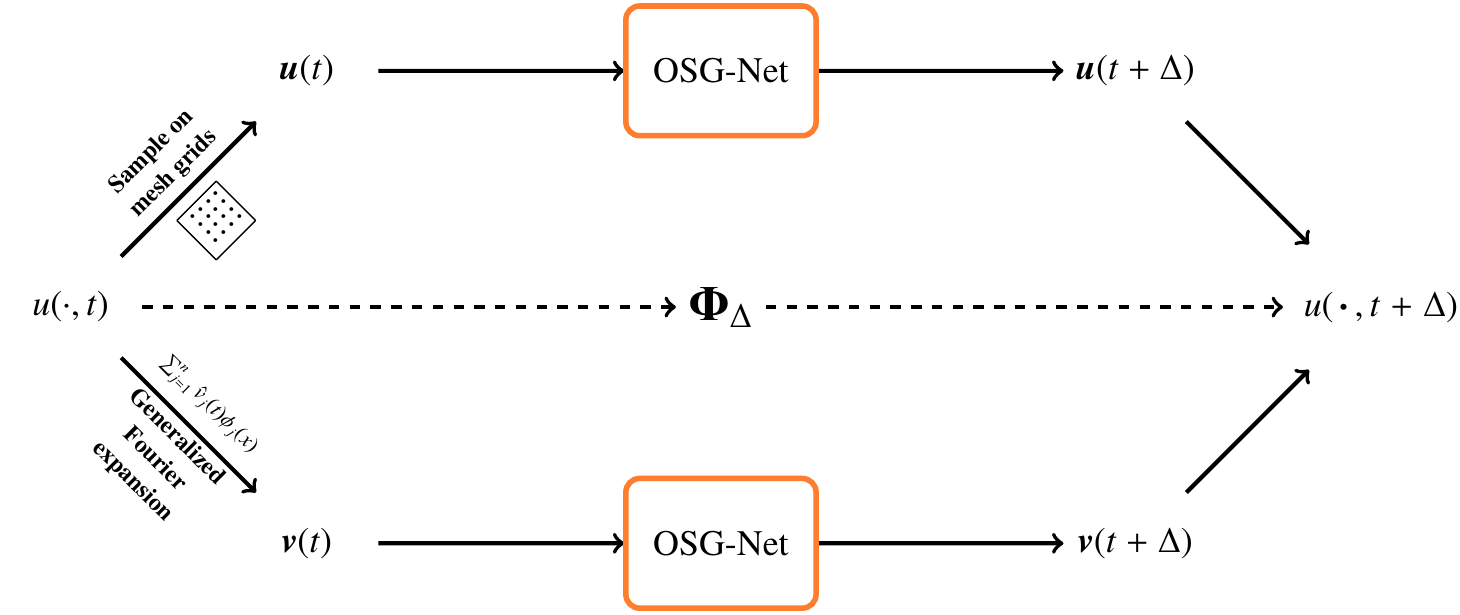} 
	\caption{\textbf{Learning PDE evolution operator in nodal and modal spaces}.}
	\label{fig:pde_learning}
\end{figure}

Unlike ODEs, for PDEs the exact evolution operator is defined on an infinite-dimensional space. To make the PDE learning problem tractable, we 
should first reduce the problem into finite dimensions in either nodal or modal spaces  
 \cite{wu2020data,chen2022deep}; see Figure \ref{fig:pde_learning} for illustration.
 
\subsection{Learning PDEs via Deep-OSG approach in nodal space}

Assume that the {the snapshots of $u$} \eqref{eq:PDEdata} are sampled on a set of spatial nodal points or grids 
\begin{equation*}
	{\mathbb X}_N = \{ x_1, \dots, x_N  \} \subset \Omega.
\end{equation*}

Define 
\begin{equation}\label{eq:PDEdatanodal}
 {\bf U}_k^i = \left(	u(x_1,t_k^{(i)}), \dots, u(x_N,t_k^{(i)}) \right)^\top,\qquad k=1,\dots,K_i,  \qquad i=1,\cdots, I_{traj}. 
\end{equation}

Define 
\begin{equation*}
	{\bf U}(t) = \big(	u(x_1,t), \dots, u(x_N,t) \big)^\top. 
\end{equation*}
In order to approximate the (unknown) infinite-dimensional evolution operator ${\bf \Phi}_\Delta$ in the nodal space, we consider a finite-dimensional evolution operator $\widehat {\bf \Phi}_\Delta$, 
which evolves ${\bf U}\in \mathbb R^N$, namely, 
\begin{equation}\label{eq:hatPhi}
	\widehat {\bf \Phi}_\Delta: \mathbb R^N \rightarrow \mathbb R^N, \qquad \widehat {\bf \Phi}_\Delta {\bf U}(t) = {\bf U}(t+\Delta). 
\end{equation}
This operator closely resembles the evolution operators of spectral collocation methods and finite difference methods for solving a known PDE. Note that here we 
assume that an autonomous PDE can be effectively approximated by an autonomous discrete system via proper spatial approximation as in a semi-discrete numerical scheme for a known PDE. 

Now our goal becomes to approximate $\widehat {\bf \Phi}_\Delta$ using the data \eqref{eq:PDEdatanodal}. 
{Like} the ODE case, we regroup the data \eqref{eq:PDEdatanodal} in the following form 
\begin{equation}\label{eq:dataPDEnodal}
	\left\{ \bm{U}_{0,j},~ \Delta_{1,j},~ \bm{U}_{1,j},~ \Delta_{2,j},~  \bm{U}_{{2},j} \right\}_{j=1}^J
\end{equation}
with $J=\sum_{i=1}^{I_{traj}}(K_i-2)$. Then we use this dataset to train an OSG-Net ${\bf N}_{\bm \theta}$ for learning the finite-dimensional evolution operator $\widehat {\bf \Phi}_\Delta$, namely, ${\bf N}_{\bm \theta} \approx \widehat {\bf \Phi}_\Delta$. 

Once the OSG-Net ${\bf N}_{\bm \theta}$ is successfully trained, we can use it to conduct prediction of the unknown PDE system on the nodal grids. For an arbitrary initial condition $u_0(x) \in \mathbb V$, we recursively apply the trained OSG-Net ${\bf N}_{\bm \theta}$ to obtain  
\begin{equation}\label{eq:uprePDE1}
	\begin{cases}
		{\bf U}^{pre}(t_0)=\left(	u_0(x_1), \dots, u_0(x_N) \right)^\top,
		\\
		{\bm U}^{pre}( t_{j+1} ) = {\bf N}_{\bm{\theta}}( {\bm U}^{pre}( t_{j} ), \Delta_j), \qquad j=0,1,\dots 
	\end{cases} 
\end{equation}
with $\Delta_j = t_{j+1}-t_j$. The function approximation to $u(x,t_j)$ can be obtained by using some 
interpolation, fitting, or reconstruction techniques. 

	Since the proposed OSG-Net architecture and the semigroup-informed loss function are non-intrusive to the basic network structures, 
	the Deep-OSG framework is very feasible. 
	Besides FNN, the OSG-Net architecture can also be combined with other suitable basic networks, such as the special neural network designed in \cite{chen2022deep} for learning PDEs in nodal space, 
	the convolution neural networks, the Fourier neural operators (FNO) \cite{li2021fourier}, and the locally connected networks,  etc.

\subsection{Learning PDEs via Deep-OSG approach in modal space}

The proposed Deep-OSG approach can also be generalized to learn evolution operator in modal space, i.e.~the generalized Fourier space. 
Let $\mathbb V_n$ denote a finite-dimensional subspace of $\mathbb V$.   
We first select a set of basis functions of $\mathbb V_n$: 
\begin{equation*}
	\left\{\phi_1(x),...,\phi_n(x)\right\}.
\end{equation*}
{defined} on the physical domain $\Omega$. The solution of the underlying PDE can be approximated in $\mathbb V_n$ by a finite-term series as 
\begin{equation*}
	u(x,t) \approx \sum_{j=1}^n {v}_j(t)\phi_j(x) =: u_n(x,t),
\end{equation*}
where $\{{v}_j\}_{j=1}^n$ are the modal expansion coefficients. Such an approximation may be given by a projection of $u(x,t)$ onto the finite-dimensional subspace $\mathbb V_n$. Let ${\mathcal P}_n: \mathbb V \rightarrow \mathbb V_n$ denote the projection operator. 
 Define the vectors 
 $$
 {\bf V}(t):=\big( {v}_1(t),\dots,{v}_n(t) \big)^\top, \quad {\bf \Phi}(x) = \big( \phi_1(x),...,\phi_n(x) \big)^\top.
 $$
We introduce the following linear mapping 
$$
\Pi: \mathbb R^n \to \mathbb V_n, \qquad  \Pi {\bf V} = \langle {\bf V},  {\bf \Phi}(x) \rangle,
$$
which defines a unique correspondence between a solution in $\mathbb V_n$ and its modal expansion coefficients vector in $\mathbb R^n$, because $\Pi$ is a bijective mapping.  

In order to approximate the (unknown) infinite-dimensional evolution operator ${\bf \Phi}_\Delta$ in the modal space, we consider a finite-dimensional evolution operator $\widetilde {\bf \Phi}_\Delta$, 
which evolves the modal expansion coefficients vector ${\bf V}(t)\in \mathbb R^n$, namely, 
\begin{equation}\label{eq:tildePhi}
	\widetilde {\bf \Phi}_\Delta: \mathbb R^n \rightarrow \mathbb R^n, \qquad \widetilde {\bf \Phi}_\Delta {\bf V}(t) = {\bf V}(t+\Delta). 
\end{equation}
Mathematically, this evolution operator can be expressed as 
$$\widetilde {\bf \Phi}_\Delta = \Pi^{-1} {\mathcal P}_n {\bf \Phi}_\Delta \Pi.$$ 
This operator closely resembles the evolution operators of spectral Galerkin methods for solving a known PDE. Again, we 
assume that an autonomous PDE can be effectively approximated by an autonomous discrete system via proper spatial approximation as in a semi-discrete numerical scheme for a known PDE. 

Now our goal is transferred to learn the finite-dimensional operator $\widetilde {\bf \Phi}_\Delta$ for modal expansion coefficients. 
Given the snapshot data \eqref{eq:PDEdata}, we first project them onto the finite-dimensional subspace $\mathbb V_n$ and compute the modal expansion coefficients:
\begin{equation}\label{eq:PDEdatamodal}
 {\bf V}_k^i = \Pi^{-1} {\mathcal P}_n	u(x,t_k^{(i)}), \qquad k=1,\dots,K_i,  \qquad i=1,\cdots, I_{traj}.
\end{equation}
We then follow the ODE case and regroup the data \eqref{eq:PDEdatamodal} in the following form 
\begin{equation}\label{eq:dataPDEmodal2}
	\left\{ \bm{V}_{0,j},~ \Delta_{1,j},~ \bm{V}_{1,j},~ \Delta_{2,j},~  \bm{V}_{{2},j} \right\}_{j=1}^J
\end{equation}
with $J=\sum_{i=1}^{I_{traj}}(K_i-2)$. We employ this dataset \eqref{eq:dataPDEmodal2} to train an OSG-Net ${\bf N}_{\bm \theta}$ for learning the finite-dimensional evolution operator $\widetilde {\bf \Phi}_\Delta$, namely, 
${\bf N}_{\bm \theta} \approx \widetilde {\bf \Phi}_\Delta.$

Once the OSG-Net ${\bf N}_{\bm \theta}$ is successfully trained, we can use it to conduct prediction of the unknown PDE system. For an arbitrary initial condition $u_0(x) \in \mathbb V$, we recursively apply the trained OSG-Net ${\bf N}_{\bm \theta}$ to obtain  
\begin{equation}\label{eq:uprePDE2}
	\begin{cases}
		{\bf V}^{pre}(t_0)= \Pi^{-1} {\mathcal P}_n u_0(x),
		\\
		{\bm V}^{pre}( t_{j+1} ) = {\bf N}_{\bm{\theta}}( {\bm V}^{pre}( t_{j} ), \Delta_j), \qquad j=0,1,\dots 
	\end{cases} 
\end{equation}
with $\Delta_j = t_{j+1}-t_j$. The predicted solution $u^{pre}(x,t_j)$ is then obtained by 
$$
u^{pre}(x,t_j) = \Pi {\bm V}^{pre}( t_{j} ) = \langle {\bm V}^{pre}( t_{j} ), {\bf \Phi}(x)\rangle = \sum_{i=1}^n v_i^{pre} (t_j) \phi_i(x)
$$
with $(v_1^{pre},\dots,v_n^{pre})^\top = {\bm V}^{pre}$.

\section{Numerical experiments}
\label{section:egs}
In this section, we conduct several numerical experiments to demonstrate the effectiveness and robustness of the proposed Deep-OSG approach. The studied cases cover a wide range of ODEs and PDEs. 
For benchmarking purpose, the true governing equations are given in all the examples, but we only use the true equations to generate training, validation, and test data. 
To mimic practical measurements with varied time steps, we randomly sample the evolution time stepsizes from a pre-defined interval $[\Delta_\text{min}, \Delta_\text{max}]$. 
Unless otherwise specified, all the data are generated by using either analytical solutions or high-order accurate numerical methods (with a sufficiently small time stepsize $\tau$) of the true equations. 
For the numerical simulation data, we compute the high-order accurate numerical solutions with $\tau \ll \Delta_\text{min}$. The dataset is organized into a number of bursts with two forward steps as in \eqref{eq:data}. 
Instead of generating a sufficiently large dataset as in  \cite{qin2019data,wu2020data,chen2022deep}, we will only use a relatively small dataset for training, so as to mimic the practical situations that the measurement data are very limited due to resource constraints or experimental accessibility. 
Before training we normalize the data to $[-1,1]$ for a better performance (see \ref{section:dn} for a comparison between normalized and non-normalized datasets) \footnote{The relative error on the test set is computed on the non-normalized data.}
When training the neural networks, $10\%$ of data are used for validation, and the updated model is saved after each mini-batch only when the validation loss decreases. Throughout this section, we employ the Gaussian Error Linear Unit (GELU) activation for the hidden neurons in OSG-Nets.
The Adam algorithm with a cyclic learning rate scheduler \cite{smith2017cyclical} is employed for training. 
All our neural network models are trained by using the open-source Tensorflow library  \cite{tensorflow2015-whitepaper}.

\begin{remark1}[Dynamic validation]\label{rem:dyval}
	When the dataset is very small (for example, there are only $10$ bursts of trajectory data for learning linear ODEs in Section \ref{section:linear}), the validation data ($10\%$ of whole data) would be too small to 
	 represent the model generalization error. 
	 To address this issue, we propose a dynamic dataset splitting technique for training and validation, {\em i.e.}, the entire dataset is shuffled after every epoch and then we dynamically split it into a training set and a validation set. As a result, the dynamic validation dataset is a more reliable estimator to the generalization error, and the whole dataset is dynamically used for back-propagation. We observe a notable advantage of this technique when the entire dataset is small; see \ref{section:dv} for a comparison.
\end{remark1}

Once a fully trained OSG-Net model is obtained, we will use it to conduct predictions of the solution and compare them against the reference solution produced by the true governing equations. Specifically, we evaluate the performance for long-time prediction based on the average relative $l_2$ error after the model marches forward many steps. This error is computed on a separate test dataset containing $I$ long trajectories $\left\{\bm{u}_{i}(t_0), \bm{u}_{i}(t_1),..., \bm{u}_i(t_M)\right\}_{i=1}^I$ with $I=100$ as follows:
\begin{equation}
 \overline{\mathcal E} = \frac{1}{M} \sum_{m=1}^M {\mathcal E}(t_m) \qquad \mbox{with} \quad 
 	{\mathcal E}(t_m)=\frac{1}{I}\sum_{i=1}^I\frac{\|\bm{u}_i(t_m) - \bm{u}^{pre}_i(t_m)\|_2}{\|\bm{u}_i(t_m)\|_2},
	\label{equation:rel_err}
\end{equation}
where each trajectory is generated by marching forward $M$ times from the initial state $\bm{u}_{k,0}$ with the time stepsize taken as  $\Delta = (\Delta_\text{min}+\Delta_\text{max})/2$, and ${\bm{u}}_{i}(t_m)$ and 
 $\bm{u}^{pre}_{i}(t_m)$ respectively denote the reference and predicted solutions at time $t_m=m\Delta$. 
To examine the prediction robustness of the neural network models, we will also investigate the variance of prediction errors obtained through different partitions of the time interval $[0,T]$, where $T$ is the final prediction time. We consider $K$ random partitions with $K=100$, 
and estimate the standard deviation of the prediction errors based on the $I$ test  trajectory data with $I=100$ as follows:
\begin{equation}\label{eq:std}
  \sigma =	\frac{1}{I}\sum_{i=1}^{I} \sqrt{\frac{1}{K}\sum_{k=1}^K \left(  {\mathcal E}_T^{k,i}-\frac{1}{K}\sum_{s=1}^K {\mathcal E}_T^{s,i} \right)^2}
\qquad \mbox{with} \quad 
	{\mathcal E}_T^{k,i} := \frac{\|\bm{u}_{i}(T) - \bm{u}^{pre}_{k,i}(T)\|_2}{\|\bm{u}_{i}(T)\|_2},\; 1 \leq k \leq K,\; 1 \leq i \leq I,
\end{equation}
where $\bm{u}^{pre}_{k,i}(T)$ denotes the $i$th trajectory solution at time $T$ predicted by the trained network model via the $k$th partition. 


\subsection{Numerical experiments on ODEs}
\subsubsection{ODE example 1: Linear ODEs}
\label{section:linear}
We first consider the following linear ODE system: 
\begin{equation}
    \begin{dcases}
        \frac{d u_1}{dt}=u_1-4u_2 + 3,\\
        \frac{d u_2}{dt}=4u_1-7u_2 + 3.
    \end{dcases}
\label{equation:eg_linear}
\end{equation}
The training data consists of only $10$ bursts of short trajectories with two forward time steps $\{\Delta_{1,j},\Delta_{2,j}\}_{j=1}^{10}$ randomly sampled from the interval $[0.05, 0.15]$ and the initial states of the trajectories randomly sampled from the domain $D=[0,2]\times [0,2]$. 
It is worth noting that the size of this dataset is much smaller than that used in \cite{qin2019data}, where more than ten thousands data pairs were employed for training. 
We adopt an OSG-Net with $3$ fully-connected hidden layers, each of which has $30$ neurons. This is an over-parameterization learning, as the number of the network parameters is much larger than the size of training dataset.  The parameters in the loss of the GDSG method are set as  $\lambda=1$ and $Q=5$. The network is trained for up to $100,000$ epochs with a batch size equal to $5$.

\begin{figure}[h!]
	\def\sc{0.45}
	\def\shift{-8mm}
	\centering
	\begin{subfigure}[t]{\sc\textwidth}
		\centering
		\shifttext{\shift}{\includegraphics[width=\linewidth]{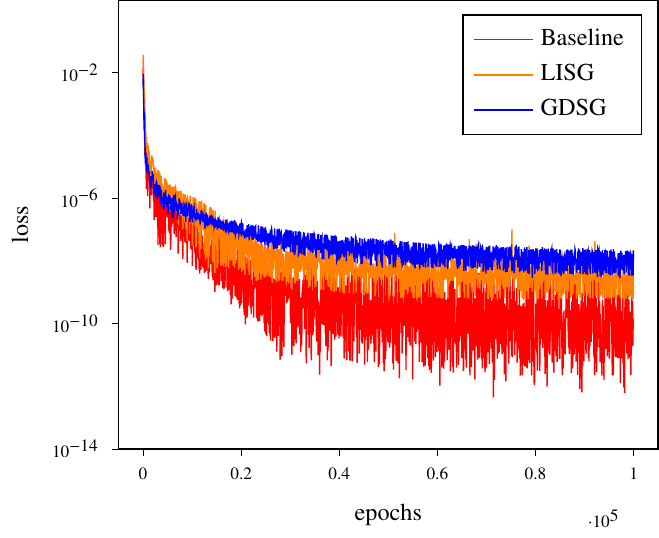}}
	\end{subfigure}\qquad
	\begin{subfigure}[t]{\sc\textwidth}
		\centering
		\shifttext{\shift}{\includegraphics[width=\linewidth]{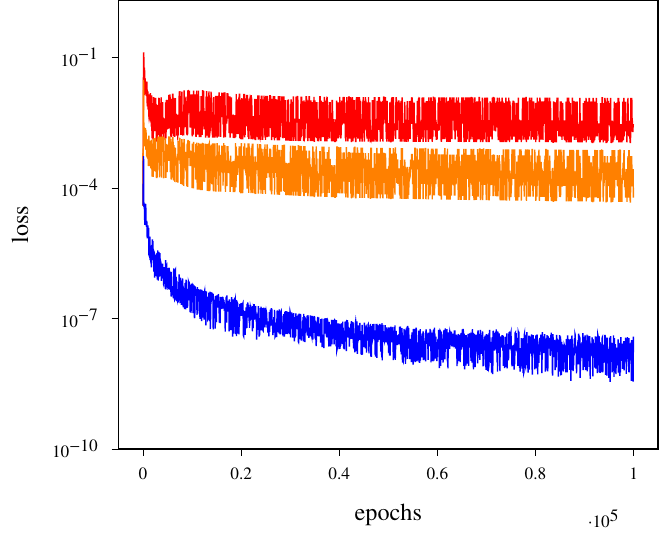}}
	\end{subfigure}
	\caption{Linear ODEs: Validation loss (left) and semi-group loss (right) during the training stage. The history is recorded every $50$ epochs.}
	\label{fig:history_linear}
\end{figure}

Figure \ref{fig:history_linear} shows the loss histories for the baseline, LISG, and GDSG methods. We see that the validation losses of  
the LISG and GDSG methods are larger than that of the baseline method. This is because 
the loss functions of the LISG and GDSG methods contain more terms. 
However, the GDSG method reduces the semigroup loss much faster, reaching a level of  $4$--$6$ orders smaller than the baseline and LISG methods at the end of training. The semigroup loss of the baseline method almost does not decrease after the first $10,000$ epochs, while its validation loss is continuously reduced.

\begin{figure}[h!]
	\centering
	\includegraphics[width=0.45\linewidth]{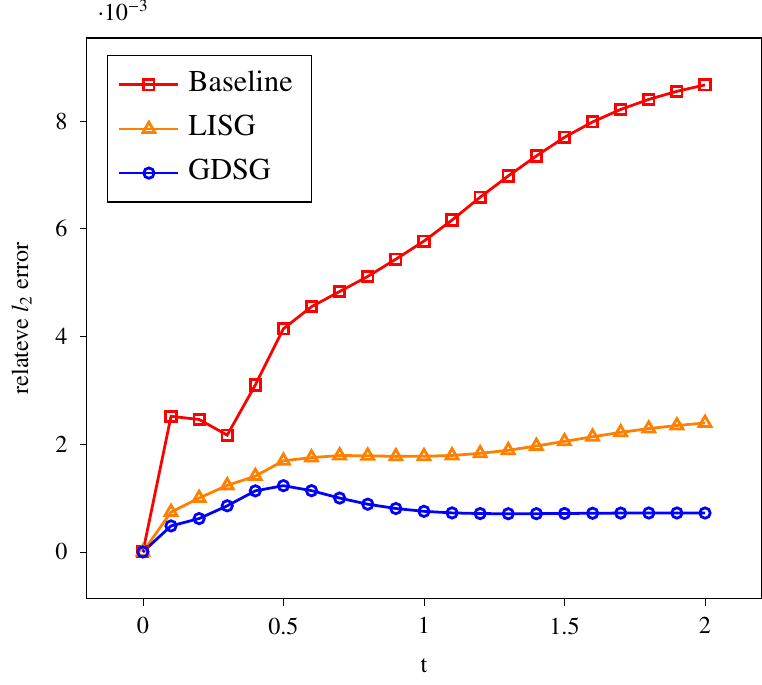}
	\caption{Linear ODEs: Evolution of the prediction error over time.}
	\label{fig:err_linear}
\end{figure}

To validate the trained models, we employ 100 different initial conditions and march the trained models forward in time up to $t=2$. The average relative $l_2$ errors $\{ {\mathcal E}(t_m) \}_{m=1}^M$ defined by \eqref{equation:rel_err} with $M=20$ are computed and shown in Figure \ref{fig:err_linear}. 
It is seen that the prediction errors of the GDSG method stay at a small level, 
the prediction errors of the LISG method increase slowly, while the prediction errors of the baseline method grow very fast. 
We choose a trajectory and show the predicted solutions as well as the portrait on the $(u_1,u_2)$ phase plane. One can observe that the GDSG method produces the most accurate prediction. We also compute the standard deviation $\sigma$ of the prediction errors defined in \eqref{eq:std} for 100 different partitions of the time interval $[0,2]$. The comparison is given in Table \ref{tab:resume_Linear}. 
The results show that the GDSG method is very robust and outperforms the baseline and LISG methods. 

\begin{figure}[h!]
	\def\sc{0.3}
	\def\shift{-7mm}
	\centering
	\begin{subfigure}[t]{\sc\textwidth}
		\centering
		\shifttext{\shift}{\includegraphics[width=\linewidth]{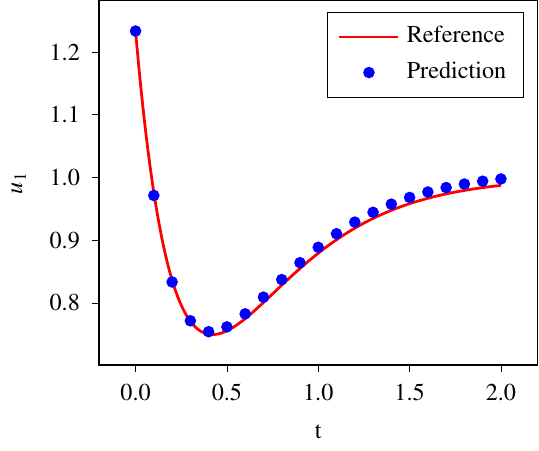}}
		\caption{$u_1$, baseline}
		\label{fig:pred_Linear_A_x1}
	\end{subfigure}\quad
	\begin{subfigure}[t]{\sc\textwidth}
		\centering
		\shifttext{\shift}{\includegraphics[width=\linewidth]{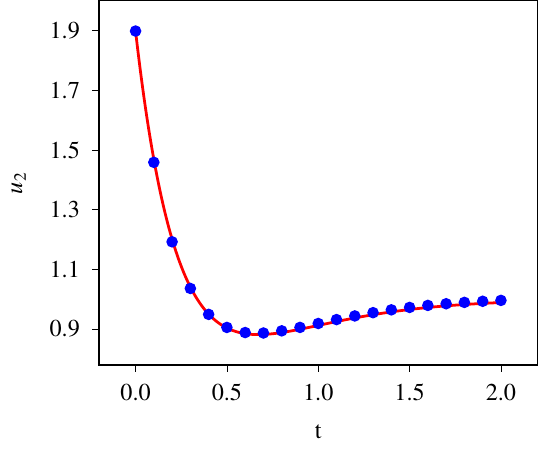}}
		\caption{$u_2$, baseline}
		\label{fig:pred_Linear_A_x2}
	\end{subfigure}\quad
	\begin{subfigure}[t]{\sc\textwidth}
		\centering
		\shifttext{\shift}{\includegraphics[width=\linewidth]{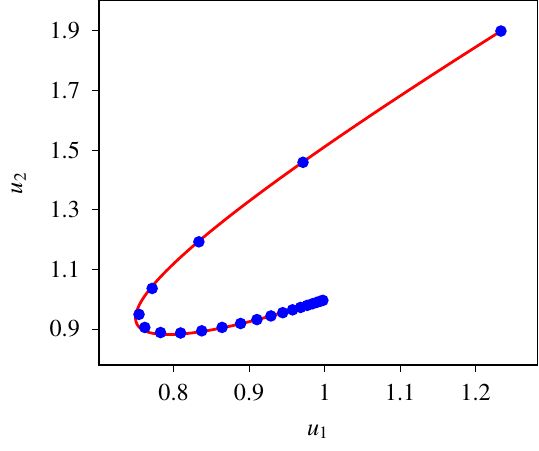}}
		\caption{Phase plot, baseline}
		\label{fig:pred_Linear_A_phase}
	\end{subfigure}%
	
	\smallskip
	
	\begin{subfigure}[t]{\sc\textwidth}
		\centering
		\shifttext{\shift}{\includegraphics[width=\linewidth]{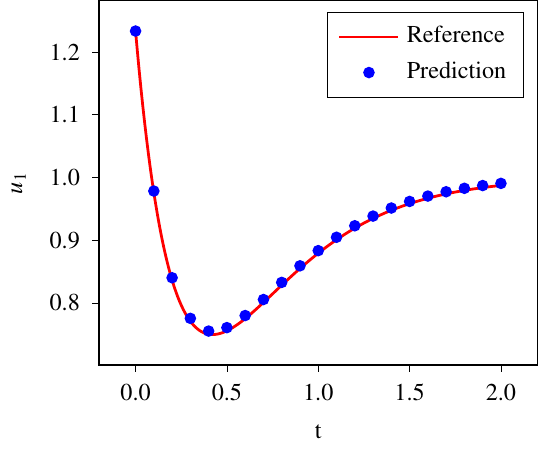}}
		\caption{$u_1$, LISG}
		\label{fig:pred_Linear_B_x1}
	\end{subfigure}\quad
	\begin{subfigure}[t]{\sc\textwidth}
		\centering
		\shifttext{\shift}{\includegraphics[width=\linewidth]{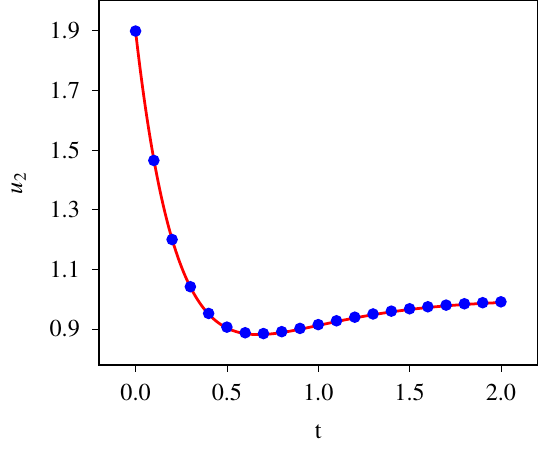}}
		\caption{$u_2$, LISG}
		\label{fig:pred_Linear_B_x2}
	\end{subfigure}\quad
	\begin{subfigure}[t]{\sc\textwidth}
		\centering
		\shifttext{\shift}{\includegraphics[width=\linewidth]{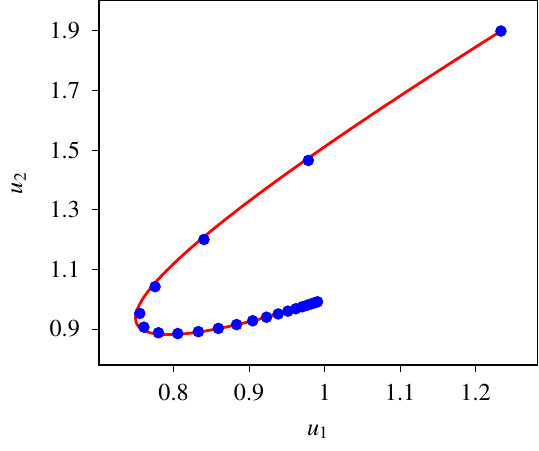}}
		\caption{Phase plot, LISG}
		\label{fig:pred_Linear_B_phase}
	\end{subfigure}%
	
	\smallskip
	
	\begin{subfigure}[t]{\sc\textwidth}
		\centering
		\shifttext{\shift}{\includegraphics[width=\linewidth]{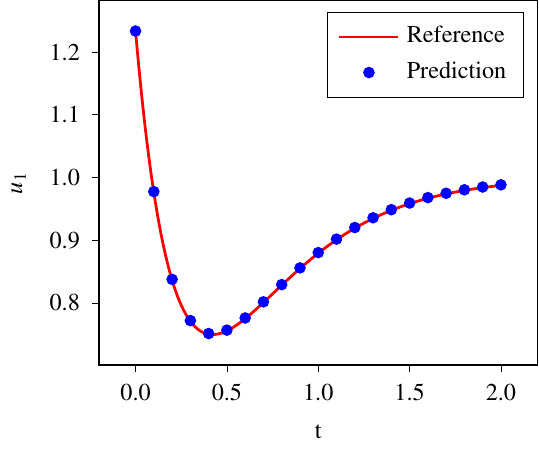}}
		\caption{$u_1$, GDSG}
		\label{fig:pred_Linear_C_x1}
	\end{subfigure}\quad
	\begin{subfigure}[t]{\sc\textwidth}
		\centering
		\shifttext{\shift}{\includegraphics[width=\linewidth]{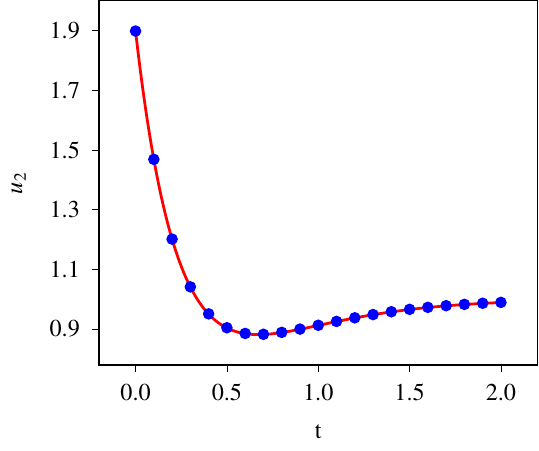}}
		\caption{$u_2$, GDSG}
		\label{fig:pred_Linear_C_x2}
	\end{subfigure}\quad
	\begin{subfigure}[t]{\sc\textwidth}
		\centering
		\shifttext{\shift}{\includegraphics[width=\linewidth]{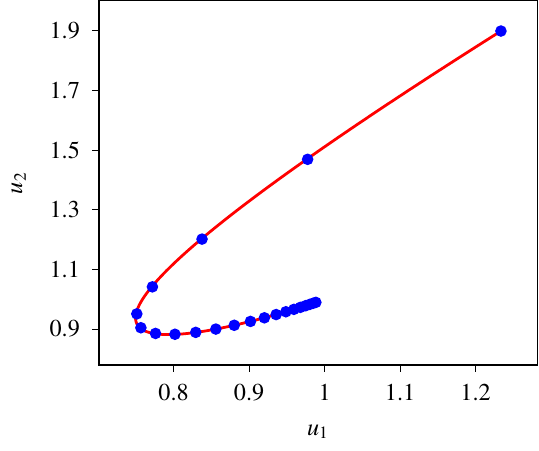}}
		\caption{Phase plot, GDSG}
		\label{fig:pred_Linear_C_phase}
	\end{subfigure}
	\caption{Linear ODEs: Trajectory and phase plots with the initial state $(1.234,1.898)$ for $t\in [0,2]$. Top: baseline method. Middle: LISG method. Bottom: GDSG method with $\lambda=1$ and $Q=5$.}
	\label{fig:pred_Linear}
\end{figure}

\begin{table}[!ht]
	\caption{Linear ODEs: Comparisons of the average prediction error $\overline{\mathcal{E}}$ on the test set, the standard deviation $\sigma$ of prediction errors, and the training time.}
	\label{tab:resume_Linear}
	\begin{center}
		\begin{tabular}{llll}
			\toprule
			\multicolumn{1}{c}{}                            &Baseline             &LISG          &GDSG\\
			\midrule
			Prediction error $\overline{\mathcal{E}}$                  &$5.558\times 10^{-3}$    &$1.708\times 10^{-3}$   &$\mathbf{7.652\times 10^{-4}}$      \\
			Standard deviation $\sigma$                   &$1.951\times 10^{-3}$    &$3.440\times 10^{-4}$   &$\mathbf{4.932\times 10^{-5}}$       \\
			\shortstack{Training time per epoch}   &$\mathbf{5.613\times 10^{-3}}$    &$8.124\times 10^{-3}$   &$8.909\times 10^{-3}$   		\\
			\bottomrule
		\end{tabular}
	\end{center}
\end{table}

\begin{figure}[h!]
	\def\sc{0.45}
	\def\shift{-14mm}
	\centering
	\begin{subfigure}[t]{\sc\textwidth}
		\centering
		\shifttext{\shift}{\includegraphics[width=\linewidth]{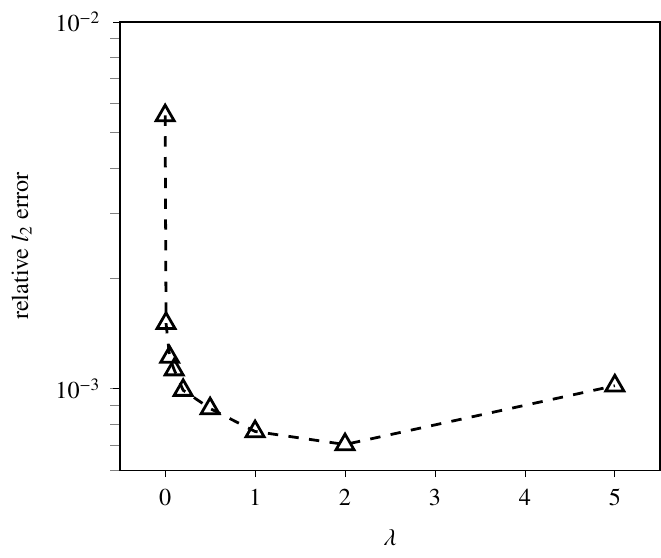}}
		\caption{}
		\label{fig:impact_weight}
	\end{subfigure}\qquad
	\begin{subfigure}[t]{\sc\textwidth}
		\centering
		\shifttext{\shift}{\includegraphics[width=\linewidth]{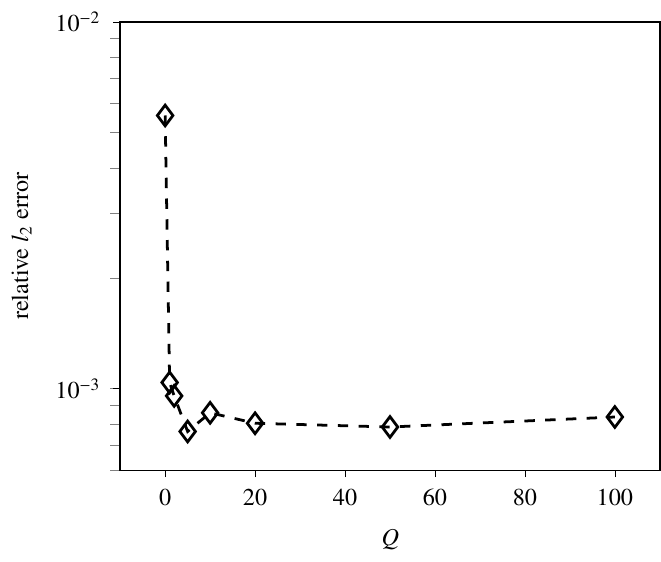}}
		\caption{}
		\label{fig:impact_M}
	\end{subfigure}
	\caption{Linear ODEs: Average prediction errors for the GDSG method with varied $\lambda$ (left) and $Q$ (right) in the loss function \eqref{equation:methodC}.}
	\label{fig:weight_M}
\end{figure}

We now study the effects of the two parameters $(\lambda,Q)$ in the loss function \eqref{equation:methodC}--\eqref{eq:SG0} of the GDSG method. 
The regularization factor $\lambda$  weighs the importance of semigroup-informed loss over data-driven loss. 
As shown in Figure \ref{fig:impact_weight}, changing the value of $\lambda$ has a notable  impact on the prediction accuracy, and there exists an optimal value around $2$ in the present case. For the second parameter $Q$, we observe positive effect when increasing $Q$ until $5$, after what the benefit becomes negligible (see Figure \ref{fig:impact_M}). It is worth noting that the loss function (\ref{equation:methodC}) with larger $Q$ requires more storage and computational costs. Therefore, a moderate $Q$, e.g.~$Q=5$, is preferred. Table \ref{tab:resume_Linear} gives a comparison of the CPU time (in seconds) of each training epoch\footnote{The training time is measured on the Intel\textregistered\; Core\textsuperscript{\texttrademark} i9$-$12900K platform.} for the baseline, LISG, and GDSG methods.

\begin{figure}[h!]
	\def\sc{0.3}
	\def\shift{-7mm}
	\centering
	\begin{subfigure}[t]{\sc\textwidth}
		\centering
		\shifttext{\shift}{\includegraphics[width=\linewidth]{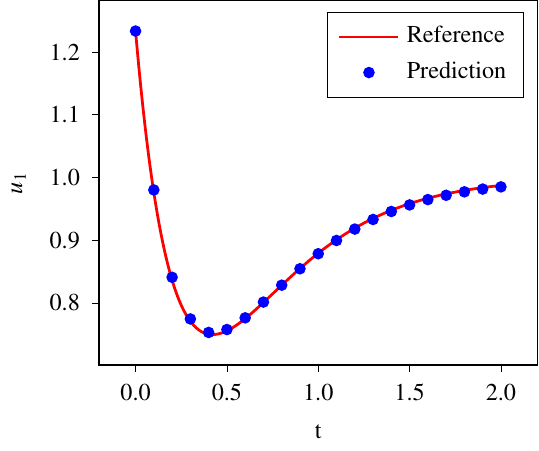}}
		\caption{$u_1$, noise level $\eta=0.02$}
	\end{subfigure}\quad
	\begin{subfigure}[t]{\sc\textwidth}
		\centering
		\shifttext{\shift}{\includegraphics[width=\linewidth]{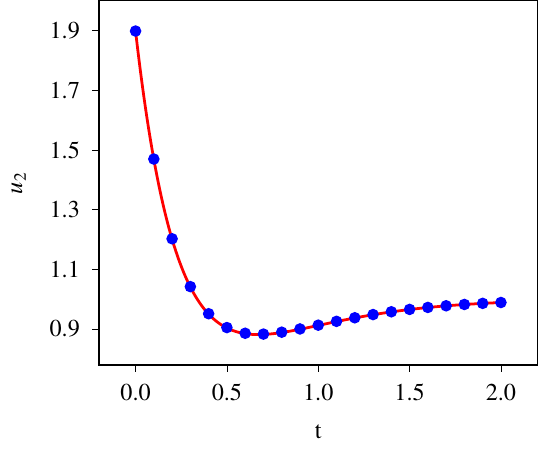}}
		\caption{$u_2$, noise level $\eta=0.02$}
	\end{subfigure}\quad
	\begin{subfigure}[t]{\sc\textwidth}
		\centering
		\shifttext{\shift}{\includegraphics[width=\linewidth]{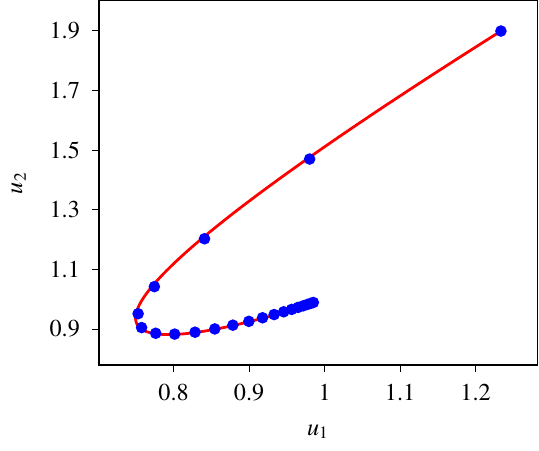}}
		\caption{Phase plot, noise level $\eta=0.02$}
	\end{subfigure}%
	
	\smallskip
	
	\begin{subfigure}[t]{\sc\textwidth}
		\centering
		\shifttext{\shift}{\includegraphics[width=\linewidth]{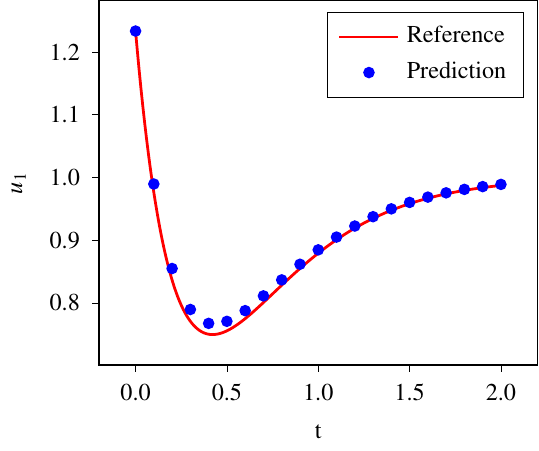}}
		\caption{$u_1$, noise level $\eta=0.05$}
	\end{subfigure}\quad
	\begin{subfigure}[t]{\sc\textwidth}
		\centering
		\shifttext{\shift}{\includegraphics[width=\linewidth]{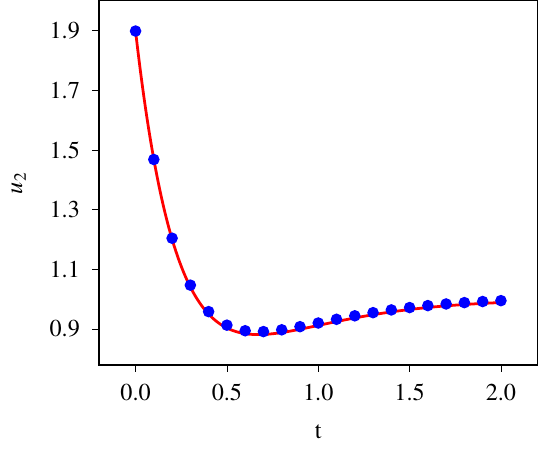}}
		\caption{$u_2$, noise level $\eta=0.05$}
	\end{subfigure}\quad
	\begin{subfigure}[t]{\sc\textwidth}
		\centering
		\shifttext{\shift}{\includegraphics[width=\linewidth]{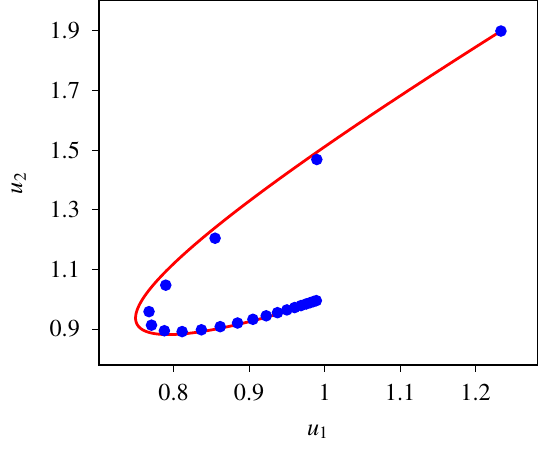}}
		\caption{Phase plot, noise level $\eta=0.05$}
	\end{subfigure}%
	\caption{Linear ODEs: Trajectory and phase plots with the initial state $(1.234,1.898)$ for $t\in [0,2]$. The prediction is obtained by the GDSG method with the training data containing $2\%$ (top) and $5\%$ noise, respectively.}
	\label{fig:pred_Linear_noise}
\end{figure}

We then consider the case of noisy data. The training data are set as $$\left\{\bm{u}_{0,j}(1+\bm{\epsilon}_{0,j}), \Delta_{1,j}, \bm{u}_{1,j}(1+\bm{\epsilon}_{1,j}), \Delta_{2,j}, \bm{u}_{2,j}(1+\bm{\epsilon}_{2,j})\right\}_{j=1}^{400},$$ 
where the relative noises $\bm{\epsilon}_{0,j}$, $\bm{\epsilon}_{1,j}$, and $\bm{\epsilon}_{2,j}$ are drawn from the uniform distribution over $[-\eta,\eta]\times [-\eta,\eta]$, with $\eta$ standing for the noise level. We conduct two experiments with $\eta$ set as $0.02$ and $0.05$, respectively. 
In Figure \ref{fig:pred_Linear_noise}, we show the phase plots generated by the GDSG method. 
It can be seen that the main structure of the solution is still well captured by our method. Thanks to the use of integral form of the underlying equations in learning the evolution operator, our method tolerates the training noise quite well, as expected. 

\subsubsection{ODE example 2: Periodic attractor}
\label{section:periodic_attractor}
This example considers a nonlinear ODE system with two variables:
\begin{equation}
	\begin{dcases}
		\frac{d u_1}{dt}=u_2-u_1(u_1^2+u_2^2-1),\\
		\frac{d u_2}{dt}=-u_1-u_2(u_1^2+u_2^2-1)
	\end{dcases}
	\label{equation:eg_circle}
\end{equation}
with the unit circle $\{(u_1,u_2): u_1^2+u_2^2=1\}$ being its periodic attractor. In this problem, the training data consists of $50$ bursts of short trajectories with two forward time steps $\{\Delta_{1,j},\Delta_{2,j}\}_{j=1}^{50}$ randomly sampled from the interval $[0.05, 0.15]$, and the initial states of the trajectories randomly sampled from 
the domain $D=[-2,2]\times [-2,2]$.

\begin{figure}[h!]
	\def\sc{0.45}
	\def\shift{-8mm}
	\centering
	\begin{subfigure}[t]{\sc\textwidth}
		\centering
		\shifttext{\shift}{\includegraphics[width=\linewidth]{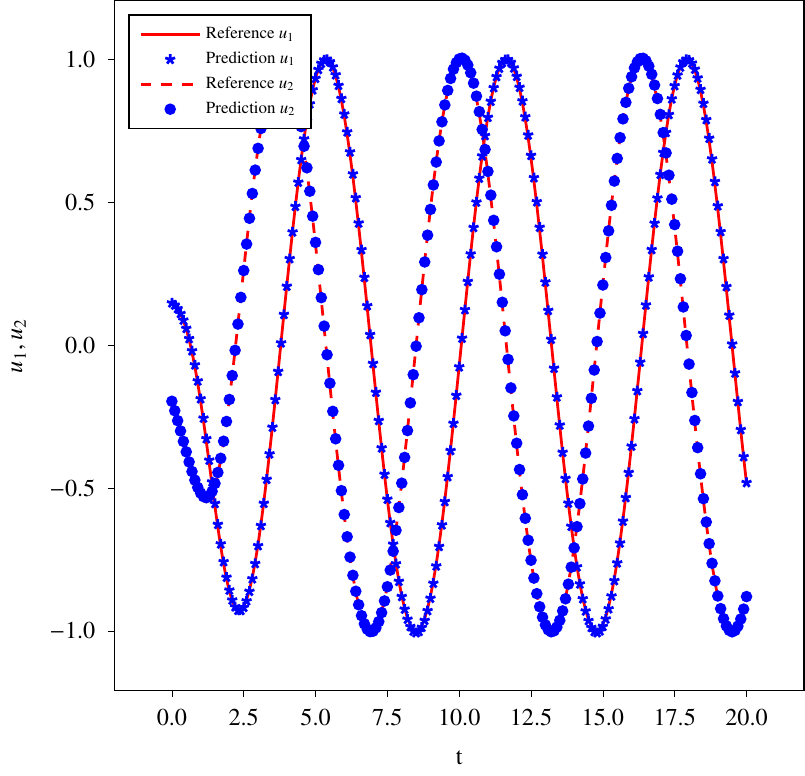}}
		\caption{$u_1$ and $u_2$, GDSG}
		\label{fig:pred_Circle_C_x1}
	\end{subfigure}\qquad
	\begin{subfigure}[t]{\sc\textwidth}
		\centering
		\shifttext{\shift}{\includegraphics[width=\linewidth]{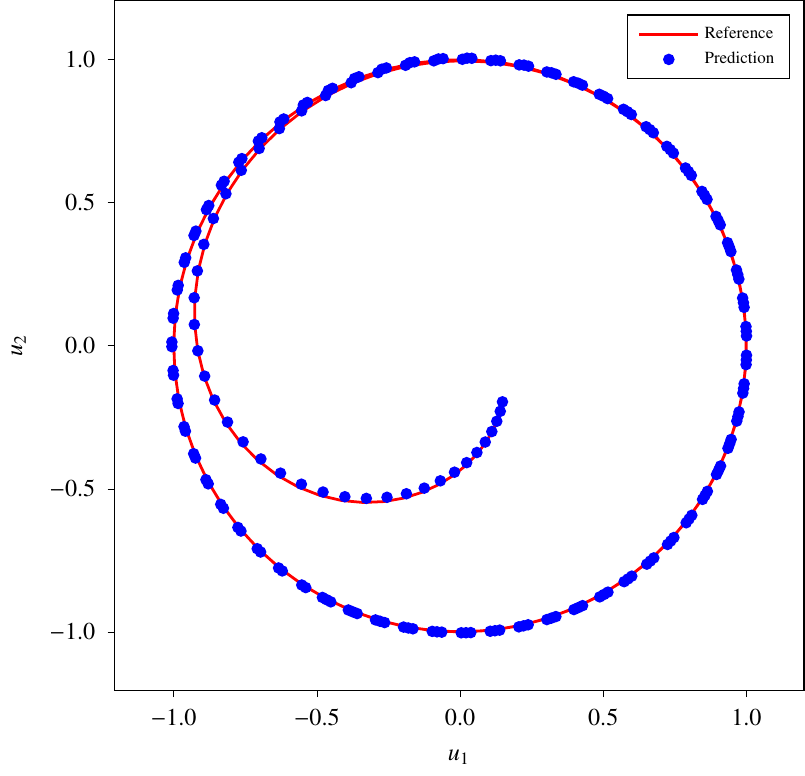}}
		\caption{Phase plot, GDSG}
		\label{fig:pred_Circle_C_phase}
	\end{subfigure}%
	\caption{Periodic attractor: Trajectory and phase plots 
		with ${\bf u}_0=(0.148,-0.196)$ for $t\in [0,20]$. The prediction is obtained by the GDSG method. }
	\label{fig:pred_Circle}
\end{figure}

The two parameters in the loss function \eqref{equation:methodC}--\eqref{eq:SG0} of the GDSG method are set as $\lambda=2$ and $Q=5$. For neural network modeling, we adopt an OSG-Net with $3$ fully-connected hidden layers, each of which has $60$ neurons. The network is trained for up to $100,000$ epochs with a batch size of $5$. Figure \ref{fig:pred_Circle} presents the long-time prediction results of the GDSG method over time up to $t=20$, \textit{i.e.} forward steps $M=200$. We see that the predicted trajectories agree well with the reference solution. The evolution of prediction error versus time is shown for the baseline, LISG, and GDSG methods in Figure \ref{fig:rel_err_PenduCircle}, from which we observe slower error growth for the GDSG and LISG methods than the baseline method. This demonstrates the importance of embedding the semigroup property for the stability of long-time prediction.  

\subsubsection{ODE example 3: Damped pendulum}
Now we consider the damped pendulum system \cite{qin2019data}, where the angle $u_1$ and velocity $u_2$ of the pendulum satisfy the following equations:
\begin{equation}
	\begin{dcases}
		\frac{d u_1}{dt}=u_2,\\
		\frac{d u_2}{dt}=-\alpha u_2- \beta \text{sin}(u_1)
	\end{dcases}
	\label{equation:eg_pendulum}
\end{equation}
with $\alpha=0.2$ and $\beta=9.8$. The training data consists of $100$ bursts of short trajectories with two forward time steps $\{\Delta_{1,j},\Delta_{2,j}\}_{j=1}^{50}$ randomly sampled from the interval $[0.05, 0.15]$, and the initial states of the trajectories randomly sampled from the domain $D=\left\{(u_1,u_2) : |u_1|\leq \pi, |u_2|\leq \text{min}\Big(2\pi,\sqrt{2\beta L\big(1+\text{cos}(u_1)\big)}\Big)\right\}$. 
Here $L=1$ is the pendulum length, and the initial gravitational potential and kinetic energy $-\beta L\text{cos}(u_1)+\frac{u_2^2}{2}$ is small enough to confine the pendulum in the domain $\{(u_1,u_2) : |u_1|\leq \pi\}$.

\begin{figure}[th!]
	\def\sc{0.45}
	\def\shift{-8mm}
	\centering
	\begin{subfigure}[t]{\sc\textwidth}
		\centering
		\shifttext{\shift}{\includegraphics[width=\linewidth]{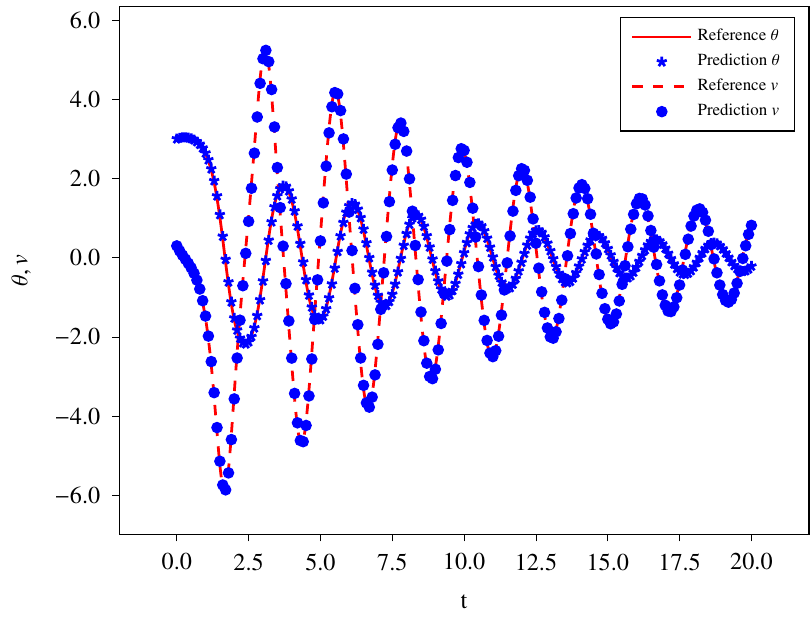}}
		\caption{$u_1$ and $u_2$, GDSG}
		\label{fig:pred_pendu_C_x1}
	\end{subfigure}\qquad
	\begin{subfigure}[t]{\sc\textwidth}
		\centering
		\shifttext{\shift}{\includegraphics[width=\linewidth]{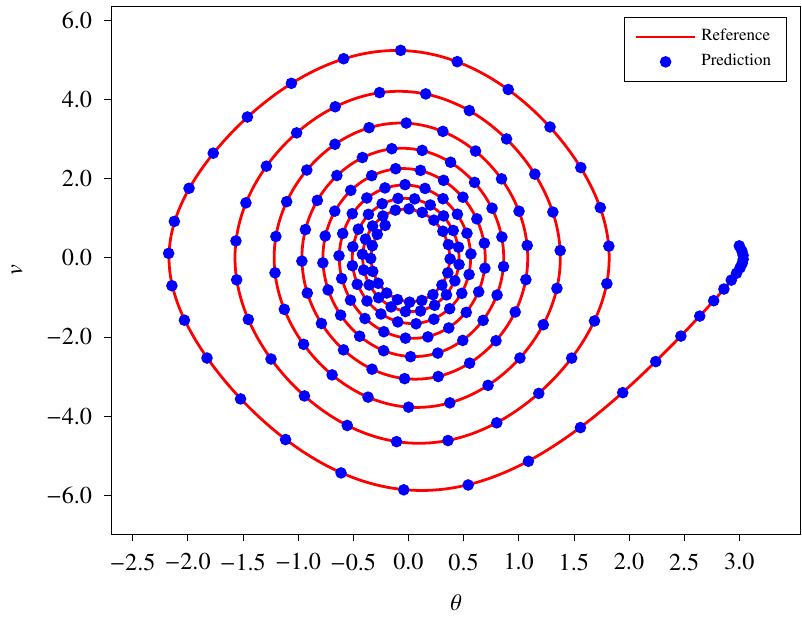}}
		\caption{Phase plot, GDSG}
		\label{fig:pred_pendu_C_phase}
	\end{subfigure}%
	\caption{Damped pendulum: 
		Trajectory and phase plots 
		with ${\bf u}_0=(3.0,0.3)$ for $t\in [0,20]$. The prediction is obtained by the GDSG method.}
	\label{fig:pred_Pendu}
\end{figure}

\begin{figure}[h!]
	\def\sc{0.45}
	\def\shift{-5mm}
	\centering
	\begin{subfigure}[t]{\sc\textwidth}
		\shifttext{\shift}{\includegraphics[width=\linewidth, height=0.8\linewidth]{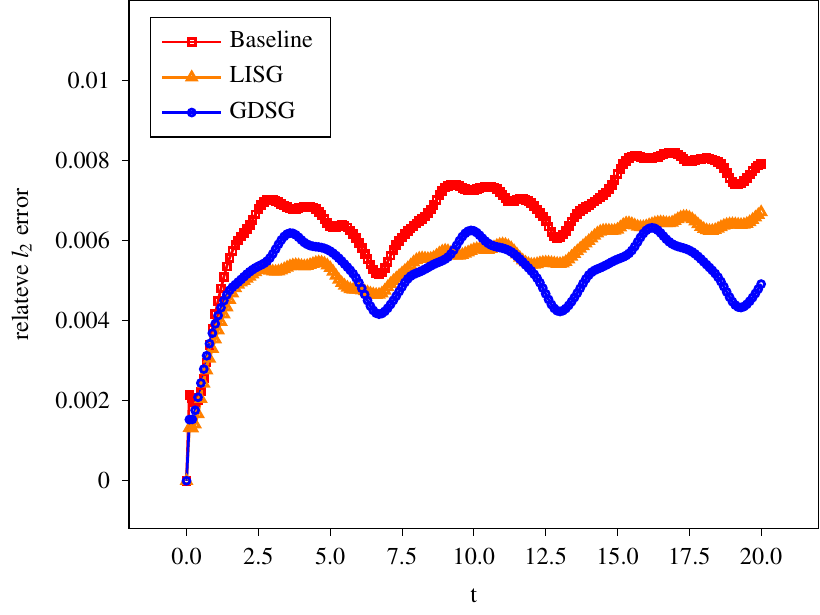}}
	\end{subfigure}\qquad
	\begin{subfigure}[t]{\sc\textwidth}
		\shifttext{\shift}{\includegraphics[width=\linewidth, height=0.8\linewidth]{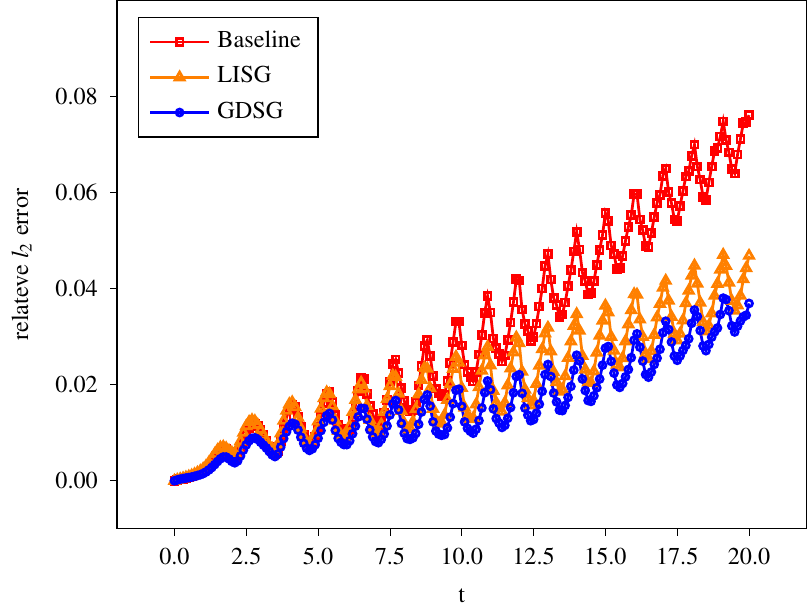}}
	\end{subfigure}
	\caption{Evolution of the average prediction error over time. Left: periodic attractor; Right: damped pendulum.}
	\label{fig:rel_err_PenduCircle}
\end{figure}

The two parameters of the GDSG method are set as $\lambda=1$ and $Q=5$. For neural network modeling, we adopt an OSG-Net with $3$ fully-connected hidden layers, each of which has $60$ neurons. The network is trained for up to $100,000$ epochs with a batch size of $5$. Figure \ref{fig:pred_Pendu} presents the long-time prediction results of the GDSG method over time with $M=200$ forward steps. The evolution of prediction error versus time is shown for the baseline, LISG, and GDSG methods in Figure \ref{fig:rel_err_PenduCircle}. Similar to the periodic attractor problem, these results show again the importance of the embedded semigroup property for prediction stability and robustness.

\subsubsection{ODE example 4: Multiscale autocatalytic chemical reaction system}\label{ex:robertson}
\label{section:robertson} 
\begin{figure}[ht!]
	\def\sc{0.75}
	\def\shift{-12mm}
	\centering
	\begin{subfigure}[t]{\sc\textwidth}
		\centering
		\shifttext{\shift}{\includegraphics[width=\linewidth]{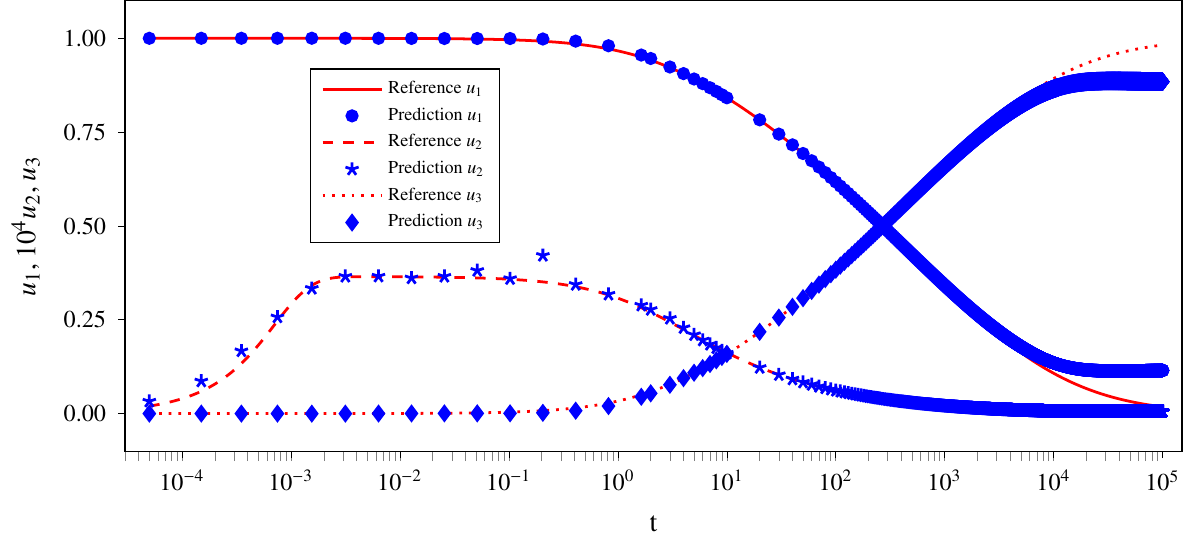}}
		\caption{Baseline method}
	\end{subfigure}
	
	\smallskip
	
	\begin{subfigure}[t]{\sc\textwidth}
		\centering
		\shifttext{\shift}{\includegraphics[width=\linewidth]{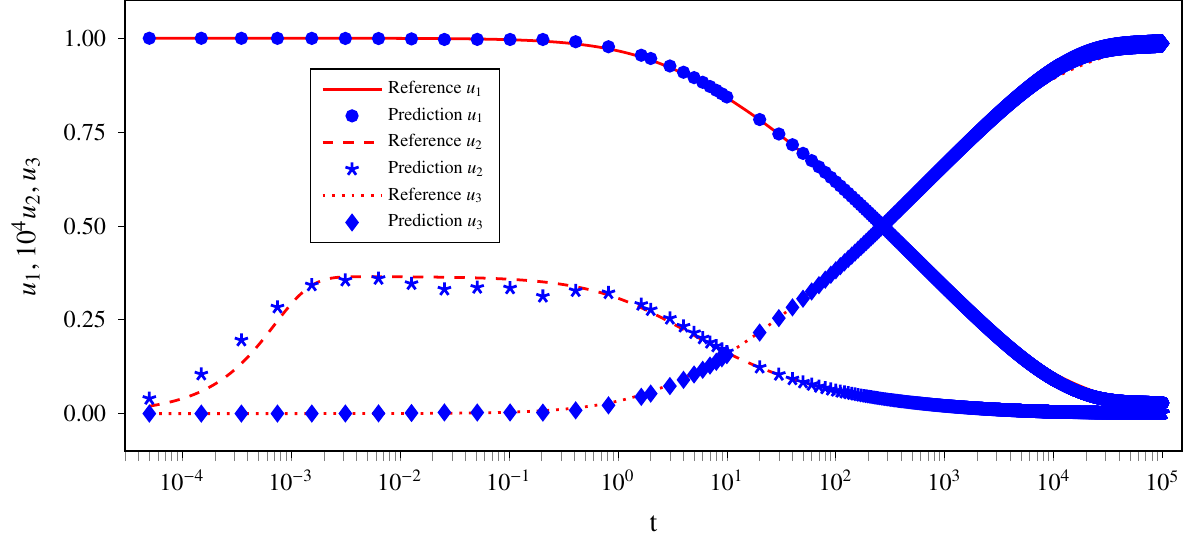}}
		\caption{LISG method}
	\end{subfigure}
	
	\smallskip
	
	\begin{subfigure}[t]{\sc\textwidth}
		\centering
		\shifttext{\shift}{\includegraphics[width=\linewidth]{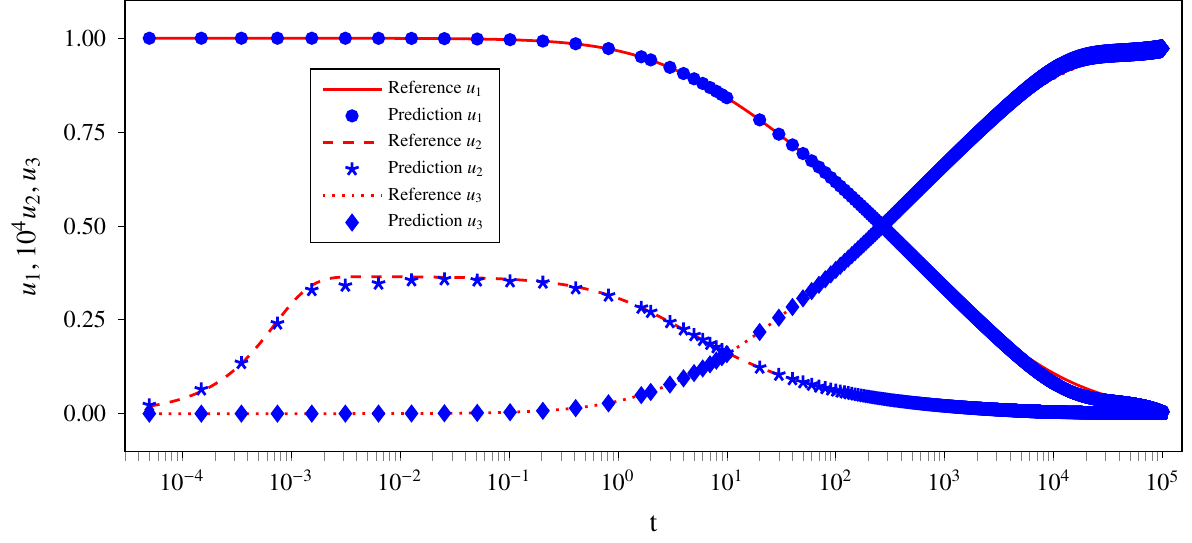}}
		\caption{GDSG method}
	\end{subfigure}
	\caption{Autocatalytic chemical reaction: Predicted and reference solutions with the initial state $(1,0,0)$ for $t\in [0,500000]$. The value of $u_2$ is multiplied by $10^4$ for clear visualization.}
	\label{fig:pred_robertson}%
\end{figure}

This is a challenging example concerning the kinetics of three chemical species, denoted by $A$, $B$, and $C$, respectively. The relevant dynamical system was proposed by Robertson \cite{robertson1966solution} and is described by the following nonlinear stiff ODEs:
\begin{equation}
	\begin{dcases}
		\frac{d u_1}{dt}=-k_1u_1 + k_2u_2u_3,\\
		\frac{d u_2}{dt}=k_1u_1 - k_2u_2u_3 - k_3u_2^2,\\
		\frac{d u_3}{dt}=k_3u_2^2,
	\end{dcases}
	\label{equation:eg_robertson}
\end{equation}
where $(u_1,u_2,u_3)$ represent the concentrations of $(A,B,C)$, with $u_1+u_2+u_3=1$. The reaction rates are $k_1=0.04$, $k_2=10^4$, and $k_3=3\times 10^7$, making the system \eqref{equation:eg_robertson} very  stiff. As shown in Figure \ref{fig:pred_robertson}, for the initial state $(1,0,0)$, the solution of this system goes through a rapid transition with the generation of catalyst $B$ when $t \in [0,10^{-3}]$, after what the reaction becomes smooth as $A$ and $B$ are transformed into $C$. To capture the  dynamics at both small and large time scales, variable step model is highly desirable. 

In order to demonstrate the capability of our methods in capturing such multiscale 
dynamics, we train different models based on two datasets. 
\begin{itemize}
		\item The first dataset consists of $500$ bursts of trajectory data, which are used to train multiscale time-step models for learning the evolution operators from small to large time scales. For this purpose, we sample the time stepsizes from the interval $10^{U[-4.9,0.1]}$, with $U[-4.9,0.1]$ being the uniform distribution on $[-4.9,0.1]$. To effectively capture the multiscale dynamics, the time-step input of the OSG-Net is modified to $-\text{log}_{10}(\Delta)$, as described in Remark \ref{rem2}. 
	We observe that such a modification is beneficial for efficiently reducing the training loss.
	\item The second dataset consists of $200$ bursts of short trajectories with two forward time steps $\{\Delta_{1,j},\Delta_{2,j}\}_{j=1}^{200}$ randomly sampled from a large interval $[5,15]$ and the initial states of the trajectories randomly sampled from the domain $D=[0,1]\times [0,10^{-4}]\times[0,1]$. 
	The dataset is used to train large-scale models for predictions with large time steps.  
\end{itemize}

Since the true equations \eqref{equation:eg_robertson} are nonlinear, we generate the datasets by using a numerical solver. In particular, we 
	adopt the variable-step, variable-order \texttt{ode15s} solver in \textsc{Matlab}, because the true  equations \eqref{equation:eg_robertson} are highly stiff. For all the models, we use an OSG-Net with $3$ fully-connected hidden layers, each of which has $60$ neurons. The parameters in the loss function of the GDSG method are set as $\lambda=1$ and $Q=5$. The network is trained for up to $100,000$ epochs with a batch size of $5$. 

In Figure \ref{fig:pred_robertson}, we present the long-term prediction results 
by the trained models, 
starting from the initial condition $(1,0,0)$ and for time up to $t=10^5$. 
In the first stage, the multi-scale model is used to predict the evolution from $t=0$ to $t=10$, with the time stepsize starting from $\Delta=5\times 10^{-5}$ and doubling after each forward step until it reaches $\Delta=1$. After that, 
the large-scale model is used for 
the second-stage prediction from $t=10$ to $t=10^5$, with the time stepsize fixed as  $\Delta=10$. The numerical solutions of the GDSG method agree well with the reference solutions in the entire long-term prediction, being more accurate than those predicted by the baseline and LISG methods. 
It is worth mentioning that the trained models, although resembling explicit algorithms, do not suffer from the standard time step restriction for such highly stiff equations. 

\subsubsection{ODE example 5: Glycolytic oscillator system}
\label{section:gly}
The Glycolytic oscillator system describes the complicated nonlinear dynamics for the concentrations of seven biochemical species \cite{raissi2018multistep,daniels2015efficient}. Its true governing equations are given by 
\begin{equation}
	\begin{dcases}
		\frac{d u_1}{dt}= J_0 - \frac{k_1u_1u_6}{1+(u_6/K_1)^q},\\
		\frac{d u_2}{dt}= 2\frac{k_1u_1u_6}{1+(u_6/K_1)^q} - k_2u_2(N-u_5) - k_6u_2u_5,\\
		\frac{d u_3}{dt}= k_2u_2(N-u_5) - k_3u_3(A-u_6),\\
		\frac{d u_4}{dt}= k_3u_3(A-u_6) - k_4u_4u_5 - \kappa (u_4-u_7),\\
		\frac{d u_5}{dt}= k_2u_2(N-u_5) - k_4u_4u_5 - k_6u_2u_5,\\
		\frac{d u_6}{dt}= -2\frac{k_1u_1u_6}{1+(u_6/K_1)^q} + 2k_3u_3(A-u_6) - k_5u_6,\\
		\frac{d u_7}{dt}= \psi\kappa(u_4-u_7) - ku_7,
	\end{dcases}
	\label{equation:Glycolytic}
\end{equation}
where the parameters are taken from \cite{daniels2015efficient} and listed in Table  \ref{tab:model_Gly}. The training data consists of $4,000$ bursts of short trajectories with two forward time steps $\{\Delta_{1,j},\Delta_{2,j}\}_{j=1}^{4000}$ randomly sampled from the interval $[0.05,0.15]$ and the initial states of the trajectories randomly sampled from the domain $D=[0.15,1.6]\times[0.19,2.16]\times[0.04,0.2]\times[0.1,0.35]\times[0.08,0.3]\times[0.14,2.67]\times[0.05,0.1]$.

\begin{table}[ht!]
	\setlength{\aboverulesep}{0pt}
	\setlength{\belowrulesep}{0pt}
	\caption{The values of parameters in the ODE system \eqref{equation:Glycolytic}.}
	\label{tab:model_Gly}
	\begin{center}
		\begin{tabular}{c|c|c|c|c|c|c|c|c|c|c|c|c|c}
			\toprule
			$J_0$  &$k_1$    &$k_2$  &$k_3$  &$k_4$  &$k_5$  &$k_6$  &$k$  &$\kappa$  &$q$  &$K_1$  &$\psi$  &$N$  &$A$\\
			\midrule
			$2.5$  &$100$  &$6$  &$16$  &$100$  &$1.28$  &$12$  &$1.8$  &$13$  &$4$  &$0.52$  &$0.1$  &$1$  &$4$\\
			\bottomrule
		\end{tabular}
	\end{center}
\end{table}

\begin{figure}[ht!]
	\def\sc{0.5}
	\def\SC{0.95}
	\def\shift{-8mm}
	\centering
	\begin{subfigure}[t]{\sc\textwidth}
		\centering
		\shifttext{\shift}{\includegraphics[width=\SC\linewidth]{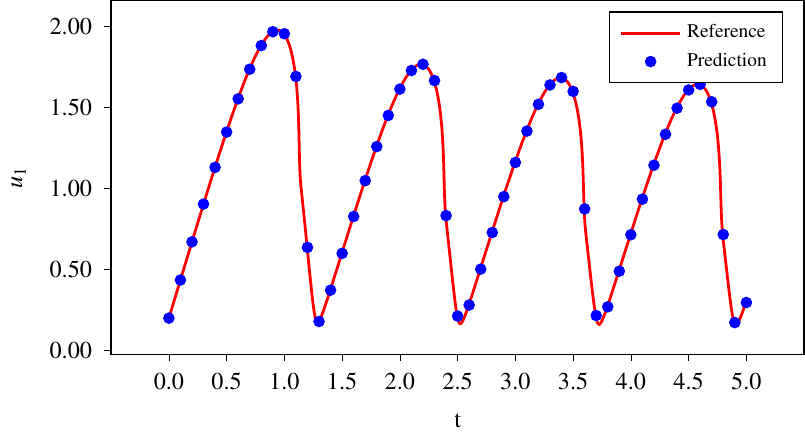}}
	\end{subfigure}%
	\begin{subfigure}[t]{\sc\textwidth}
		\centering
		\shifttext{\shift}{\includegraphics[width=\SC\linewidth]{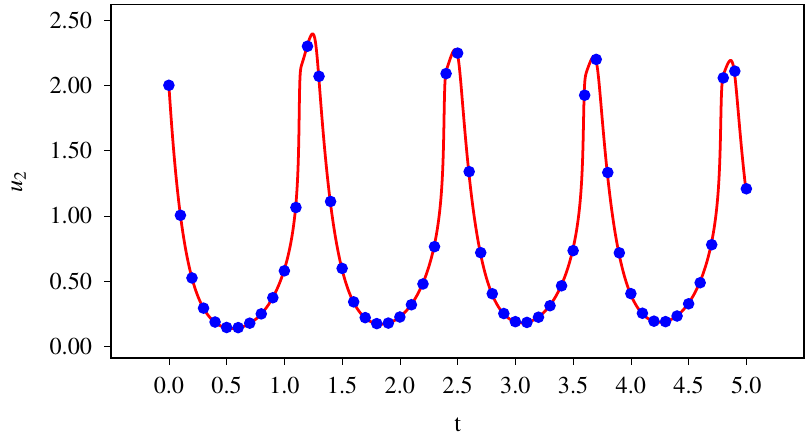}}
	\end{subfigure}%
	
	\smallskip
	
	\begin{subfigure}[t]{\sc\textwidth}
		\centering
		\shifttext{\shift}{\includegraphics[width=\SC\linewidth]{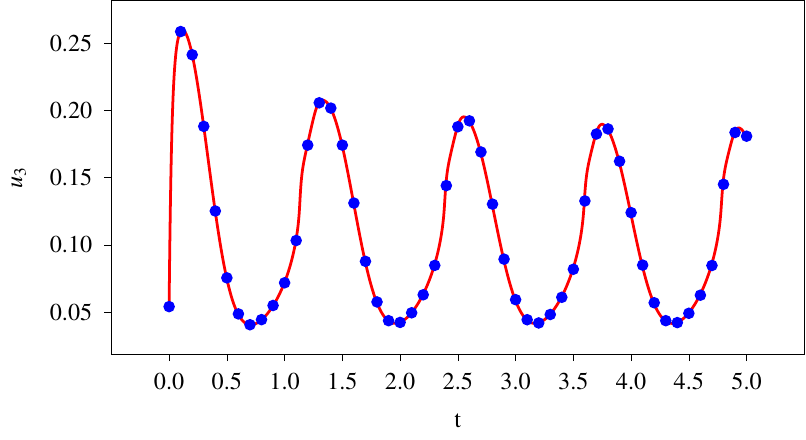}}
	\end{subfigure}%
	\begin{subfigure}[t]{\sc\textwidth}
		\centering
		\shifttext{\shift}{\includegraphics[width=\SC\linewidth]{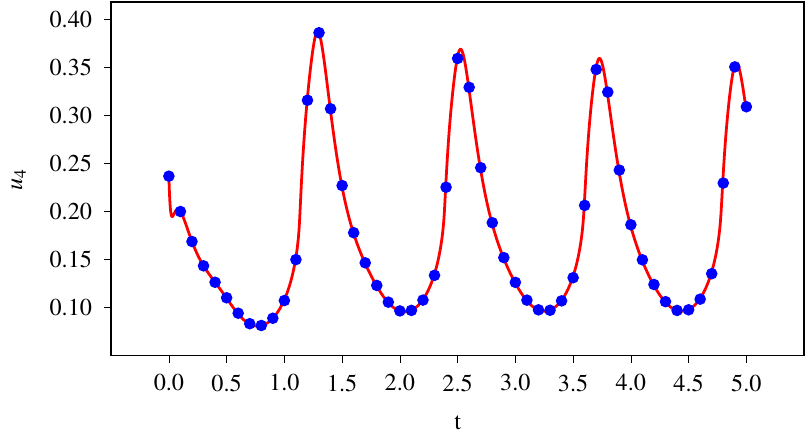}}
	\end{subfigure}%
	
	\smallskip
	
	\begin{subfigure}[t]{\sc\textwidth}
		\centering
		\shifttext{\shift}{\includegraphics[width=\SC\linewidth]{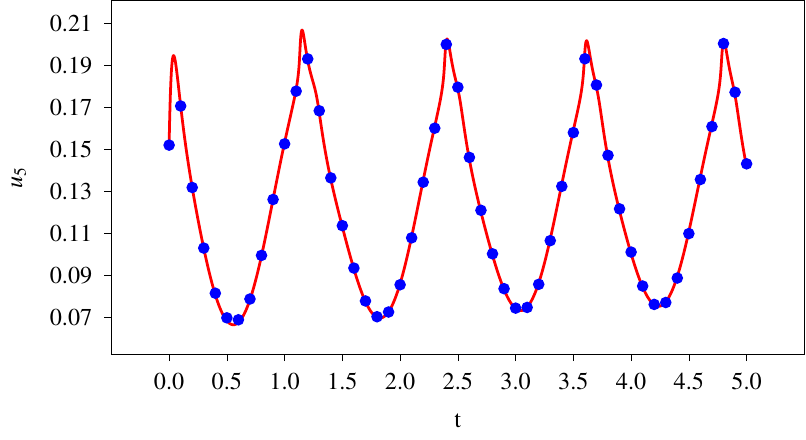}}
	\end{subfigure}%
	\begin{subfigure}[t]{\sc\textwidth}
		\centering
		\shifttext{\shift}{\includegraphics[width=\SC\linewidth]{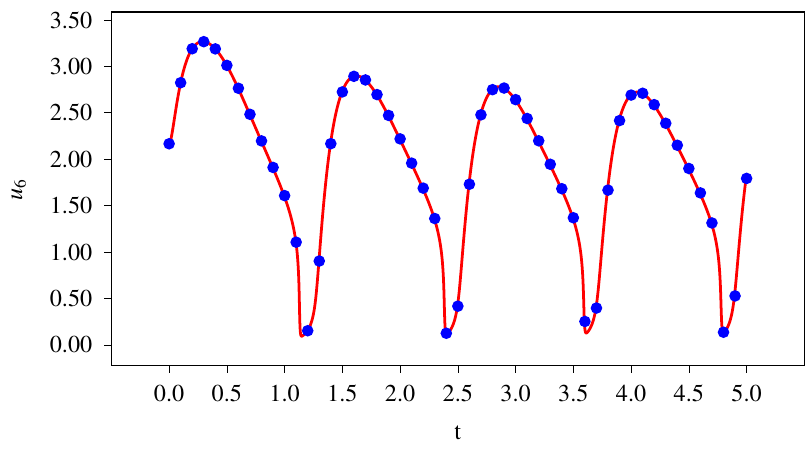}}
	\end{subfigure}%
	
	\smallskip
	
	\begin{subfigure}[t]{\sc\textwidth}
		\centering
		\shifttext{\shift}{\includegraphics[width=\SC\linewidth]{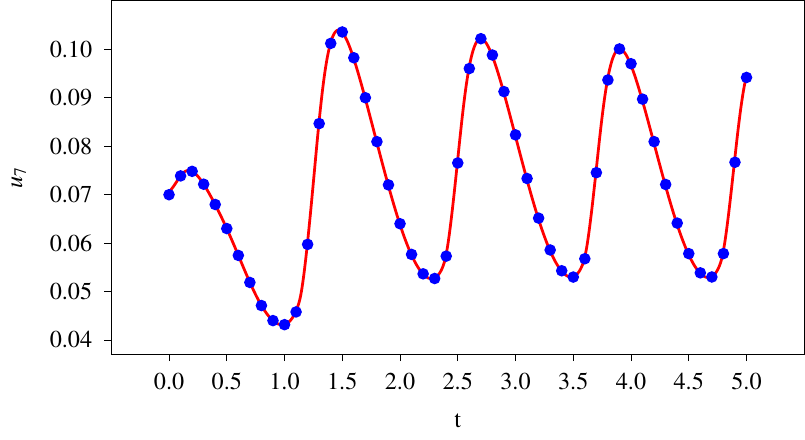}}
	\end{subfigure}
	\caption{Glycolytic oscillator: Reference solution and predicted solution by the GDSG method. The initial state is $(0.2,2.0,0.054,0.237,0.152,2.167,0.07)$.}
	\label{fig:pred_Gly}%
\end{figure}

In order to capture the complicated high-dimensional dynamics, 
we adopt a deep multi-step recursive OSG-Net with $4$ blocks, and each block has $3$ hidden layers with the equal width of $40$ neurons.  The network is trained for up to $100,000$ epochs with a batch size of $90$. The parameters in the loss function of the GDSG method are set as $\lambda=1$ and $Q=5$. 
Once the models are trained satisfactorily, we conduct predictions and evaluate the prediction errors on 100 test trajectory data with forward steps $M=50$. 
The average prediction errors and the standard deviation are shown in Table \ref{tab:resume_Gly}, from which we see the advantages of the proposed GDSG method in both accuracy and robustness. 
Figure \ref{fig:pred_Gly} presents the prediction results of the GDSG method for the trajectory starting from the initial state $(0.2,2.0,0.054,0.237,0.152,2.167,0.07)$. 
One can observe the excellent agreement between the predicted and reference solutions. 

\begin{table}[ht!]
	\caption{Glycolytic oscillator: Average prediction error $\overline{\mathcal{E}}$ on the test set and the standard deviation $\sigma$ of prediction errors.}
	\label{tab:resume_Gly}
	\begin{center}
		\begin{tabular}{llll}
			\toprule
			\multicolumn{1}{c}{}                            &Baseline             &LISG          &GDSG  \\ 
			\midrule
			Prediction error $\overline{\mathcal{E}}$                  &$0.04439$    &$0.0216$   &$\mathbf{8.138\times 10^{-3}}$         \\
			Standard deviation $\sigma$                  &$0.0259$    &$0.0132$   &$\mathbf{4.94\times 10^{-3}}$           \\
			\bottomrule
		\end{tabular}
	\end{center}
\end{table}

\subsection{Numerical experiments on PDEs}
In this section, we present several numerical experiments on a wide variety of PDEs. 
\subsubsection{PDE example 1: Advection equation}
\label{section:advection_eq}
This example considers a 1D advection equation with periodic boundary condition:
\begin{equation}
\begin{cases}
            \partial_t u +  \partial_x u= 0, \quad & (x,t)\in (0,2\pi)\times \mathbb{R}^+\\
            u(0,t) = u(2\pi,t), & t \ge 0. 
\end{cases}
\label{equation:eg_advection}
\end{equation}
We learn the evolution operators via the Deep-OSG approach in Fourier modal space. 
As in \cite{wu2020data}, the finite-dimensional approximation space is taken as ${\mathbb V}_7 =
\text{span}\{1,\text{cos}(x),\text{sin}(x),\text{cos}(2x),\text{sin}(2x),\text{cos}(3x),\text{sin}(3x)\}$. The training data consists of $J=1,000$ bursts of short ``trajectories'' with two forward time stepsizes $\{\Delta_{1,j},\Delta_{2,j}\}_{j=1}^J$ randomly sampled in the interval $[0.05,0.15]$. 
The initial conditions of the trajectories are generated in ${\mathbb V}_7$ with the Fourier coefficients randomly sampled from the domain $D=[-0.8,0.8]^3\times[-0.2,0.2]^2\times[-0.03,0.03]^2$ in modal space by using uniform distribution. It is worth noting that the size of our dataset is much smaller than that used in \cite{wu2020data}, where $80,000$ data pairs were employed for training. Our numerical experiments indicate that the proposed Deep-OSG approach is quite suitable for learning tasks with limited data. 

\begin{figure}[ht!]
	\centering
	\shifttext{-8mm}{\includegraphics[width=0.45\linewidth]{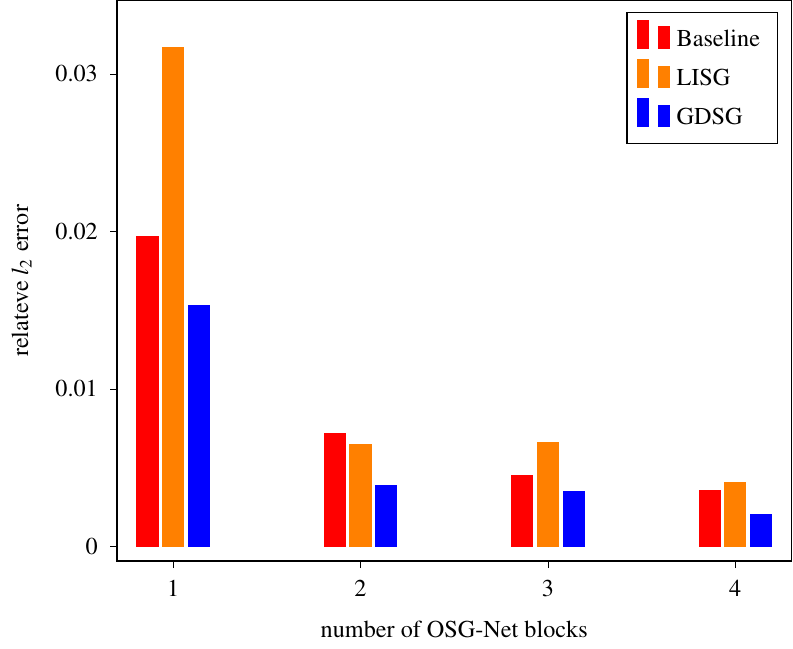}}
	\caption{Advection equation: Average prediction errors $\overline{\mathcal{E}}$ for the multi-step recursive OSG-Net with different numbers of blocks.} 
	\label{fig:err_advection}
\end{figure}

For neural network modeling, we use the multi-step recursive OSG-Net architecture. 
In order to study the effect of the number of OSG-Net blocks, we adopt four recursive OSG-Nets with $1$, $2$, $3$, and $4$ blocks, respectively. 
Each OSG-Net block contains $3$ hidden fully-connected layers of equal width of $20$ neurons. The networks are trained for up to $10,000$ epochs with a batch size of $30$. 
We set $\lambda=1$ and $Q=5$ in the loss function of the GDSG method. 
After training the networks satisfactorily, we conduct predictions and evaluate the prediction errors on 100 test trajectory data with forward steps $M=200$. 
The average relative prediction errors for different network architectures  
are compared in Figure  \ref{fig:err_advection} for the baseline, LISG and GDSG methods. 
As we can see, the errors decrease {when} more OSG-Net blocks are used.   

\begin{figure}[ht!]
	\def\sc{0.31}
	\def\shift{-8mm}
	\centering
	\begin{subfigure}[t]{\sc\textwidth}
		\centering
		\shifttext{\shift}{\includegraphics[width=\linewidth]{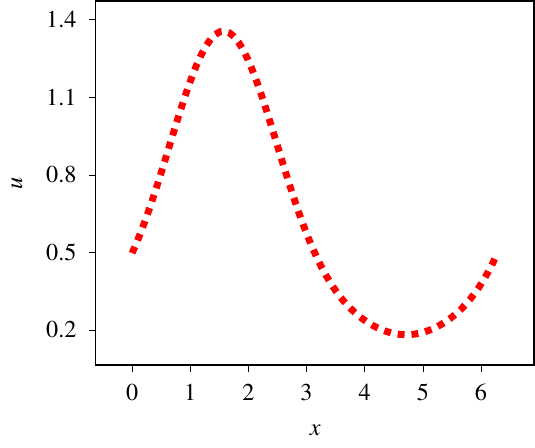}}
		\caption{$u(x,t=0)=\frac{1}{2}\text{exp}(\text{sin}(x))$}
	\end{subfigure}\quad
	\begin{subfigure}[t]{\sc\textwidth}
		\centering
		\shifttext{\shift}{\includegraphics[width=\linewidth]{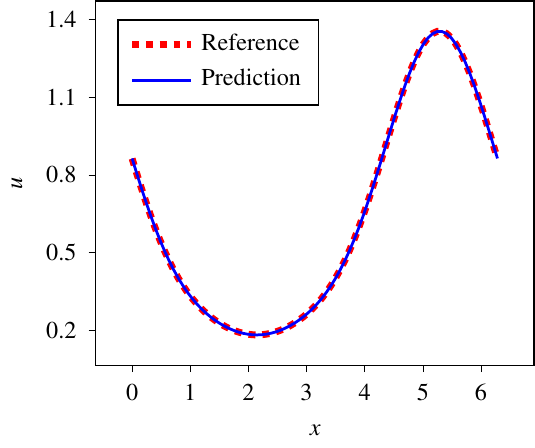}}
		\caption{$u(x,t=10)$}
	\end{subfigure}\quad
	\begin{subfigure}[t]{\sc\textwidth}
		\centering
		\shifttext{\shift}{\includegraphics[width=\linewidth]{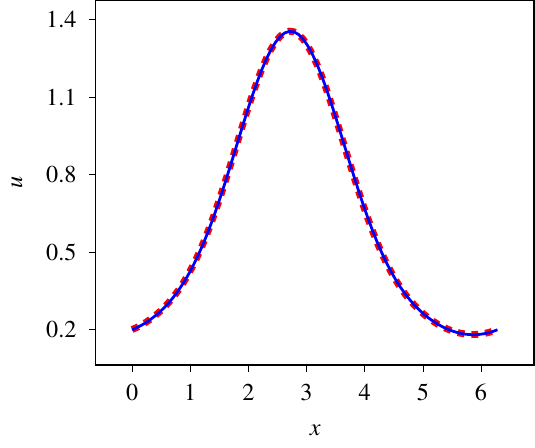}}
		\caption{$u(x,t=20)$}
	\end{subfigure}%
	\caption{Advection equation: Comparison of the true solution and the learned model solution at $t=10$ and $t=20$ given by the GDSG method with $4$ OSG-Net blocks. }
	\label{fig:sol_adv}
\end{figure}

\begin{figure}[ht!]
	\def\sc{0.5}
	\def\SC{0.95}
	\def\h{0.5}
	\def\shift{-10mm}
	\centering
	\begin{subfigure}[t]{\sc\textwidth}
		\centering
		\shifttext{\shift}{\includegraphics[width=\SC\linewidth, height=\h\linewidth]{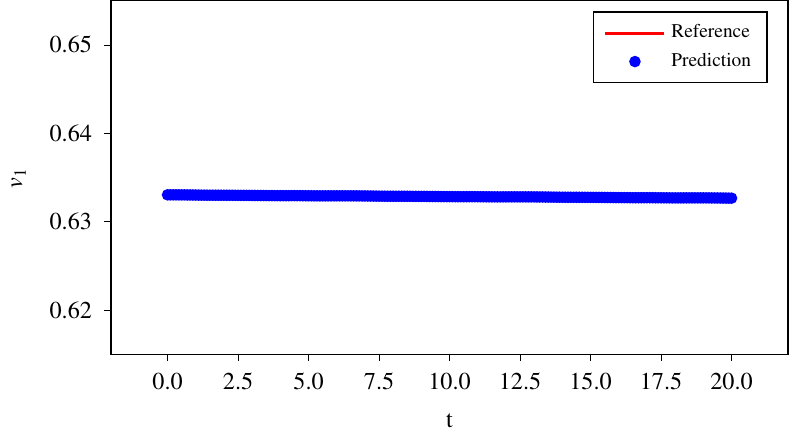}}
	\end{subfigure}%
	\begin{subfigure}[t]{\sc\textwidth}
		\centering
		\shifttext{\shift}{\includegraphics[width=\SC\linewidth, height=\h\linewidth]{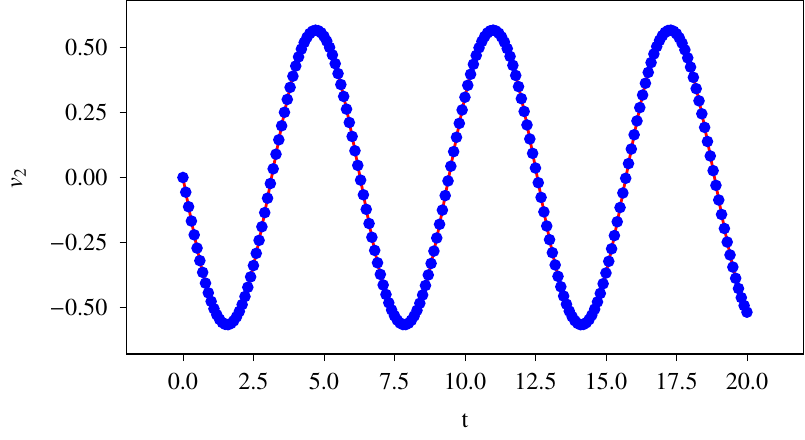}}
	\end{subfigure}%
	
	\smallskip
	
	\begin{subfigure}[t]{\sc\textwidth}
		\centering
		\shifttext{\shift}{\includegraphics[width=\SC\linewidth, height=\h\linewidth]{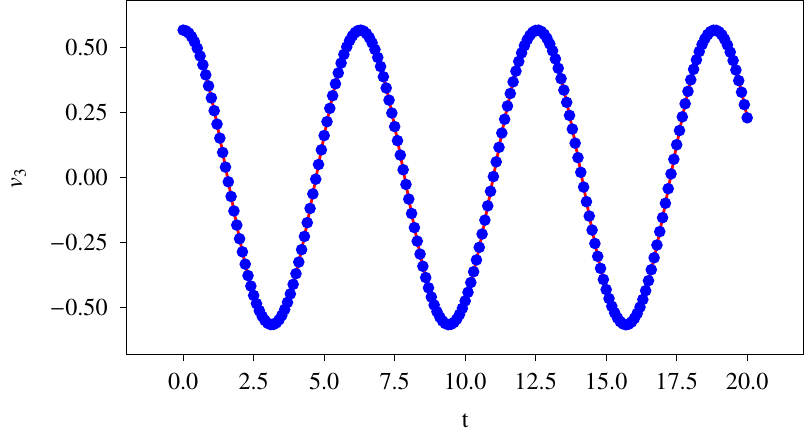}}
	\end{subfigure}%
	\begin{subfigure}[t]{\sc\textwidth}
		\centering
		\shifttext{\shift}{\includegraphics[width=\SC\linewidth, height=0.505\linewidth]{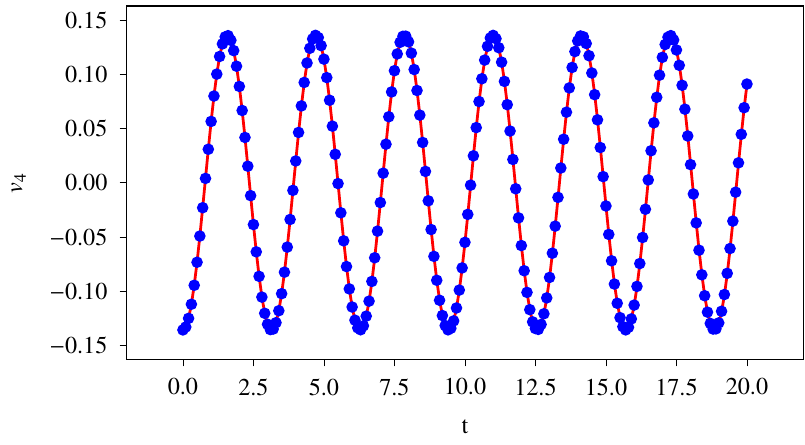}}
	\end{subfigure}%
	
	\smallskip
	
	\begin{subfigure}[t]{\sc\textwidth}
		\centering
		\shifttext{\shift}{\includegraphics[width=\SC\linewidth, height=\h\linewidth]{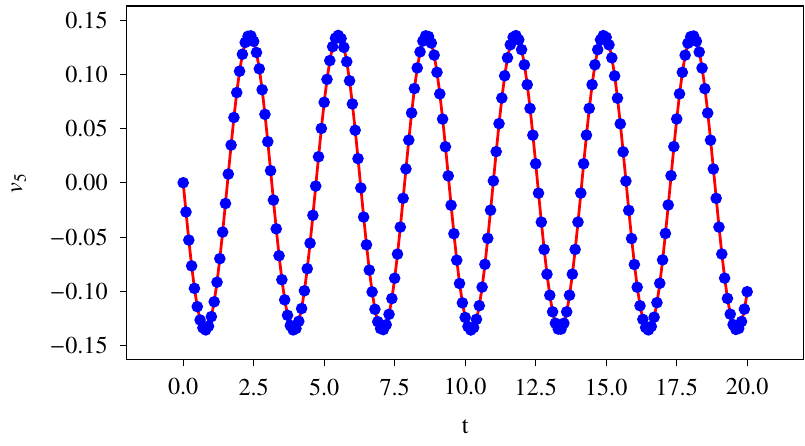}}
	\end{subfigure}%
	\begin{subfigure}[t]{\sc\textwidth}
		\centering
		\shifttext{\shift}{\includegraphics[width=\SC\linewidth, height=0.495\linewidth]{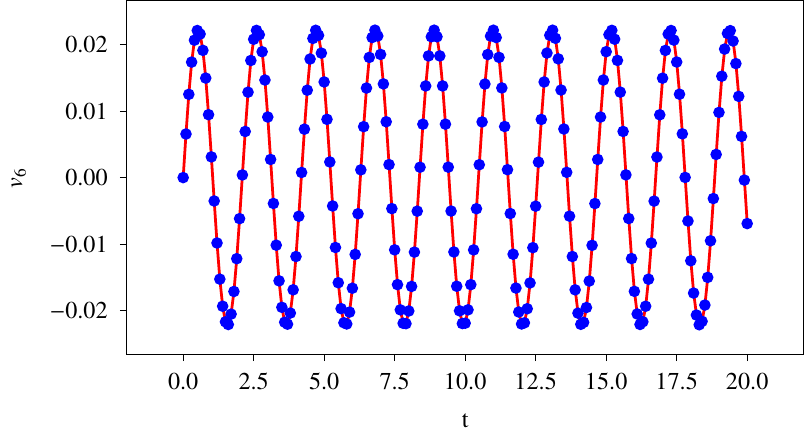}}
	\end{subfigure}%
	
	\smallskip
	
	\begin{subfigure}[t]{\sc\textwidth}
		\centering
		\shifttext{\shift}{\includegraphics[width=\SC\linewidth, height=\h\linewidth]{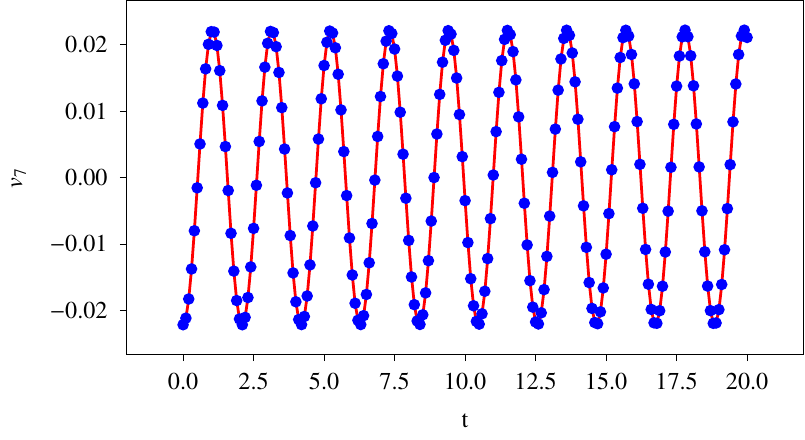}}
	\end{subfigure}
	\caption{Advection equation: Evolution of the expansion Fourier coefficients for the projection of the true solution and the learned model solution given by the GDSG method with $4$ OSG-Net blocks.}
	\label{fig:proj_adv}
\end{figure}

To validate the model, we take an initial condition $u(x,0)=\frac{1}{2}\text{exp}(\text{sin}(x))$ outside the approximation space $\mathbb{V}_7$, and conduct simulations in the form \eqref{eq:uprePDE2} using the trained model given by the GDSG method with $4$ blocks.  
In Figure \ref{fig:sol_adv}, the solution prediction of the trained model at $t=10$ and $t=20$ is plotted, along with the true solution for reference. 
One can observe that the network model produces accurate prediction results. 
To further validate the prediction accuracy, we also present in  Figure \ref{fig:proj_adv} 
the evolution of the learned expansion coefficients $v_j$, $1\le j \le 7$. For comparison, the optimal Fourier coefficients obtained by the orthogonal projection of the true solution onto $\mathbb V_7$ are also plotted. We see excellent agreement between the predicted and true solutions. 

\subsubsection{PDE example 2: Viscous Burgers' equation}
We consider the viscous Burgers' equation with Dirichlet boundary condition:
\begin{equation}
	\begin{dcases}
			\partial_t u +  \partial_x \left(\frac{u^2}{2}\right)= \frac{1}{10} \partial_{xx}u, \quad & (x,t)\in (-\pi,\pi)\times \mathbb{R}^+,\\
			u(-\pi,t) = u(\pi,t)  = 0, & t \ge 0.
	\end{dcases}
\end{equation}
In this example, the evolution operators are also learned in Fourier modal space. 
Following \cite{wu2020data}, the finite-dimensional approximation space is taken as $\mathbb{V}_9 = \text{span}\{\text{sin}(ix)\}_{i=1}^9$. The training data consists of $J=2,000$ bursts of short ``trajectories'' with two forward time stepsizes $\{\Delta_{1,j},\Delta_{2,j}\}_{j=1}^J$ randomly sampled in the interval $[0.025,0.075]$. 
The initial conditions of the trajectories are generated in ${\mathbb V}_9$ with the Fourier coefficients randomly sampled from the domain $D=[-1.5,1.5]\times[-0.5,0.5]\times[-0.2,0.2]^2\times[-0.1,0.1]^2\times[-0.05,0.05]^2\times[-0.02,0.02]$ in modal space by using uniform distribution. It is worth noting that the size of our dataset is much smaller than that employed in \cite{wu2020data}, where $500,000$ data pairs were employed for training. 

\begin{figure}[ht!]
	\def\sc{0.31}
	\def\shift{-8mm}
	\centering
	\begin{subfigure}[t]{\sc\textwidth}
		\centering
		\shifttext{\shift}{\includegraphics[width=\linewidth]{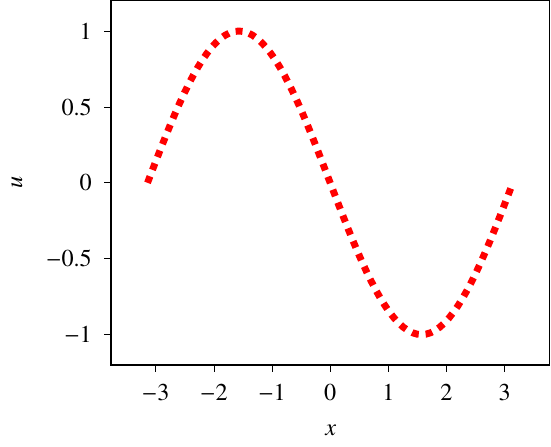}}
		\caption{$u(x,0)=-\text{sin}(x)$}
		\label{fig:pred_Burger_init}
	\end{subfigure}\quad
	\begin{subfigure}[t]{\sc\textwidth}
		\centering
		\shifttext{\shift}{\includegraphics[width=\linewidth]{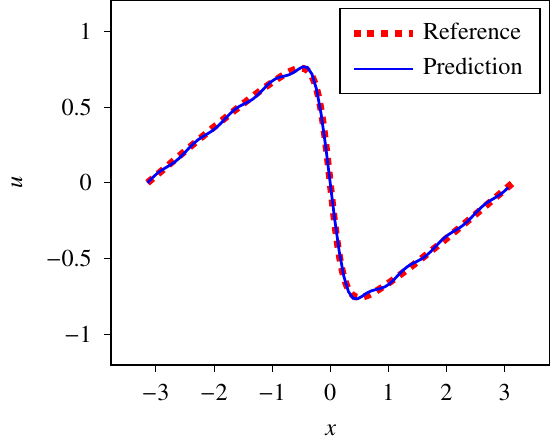}}
		\caption{$u(x,t=2)$}
		\label{fig:pred_Burger2}
	\end{subfigure}\quad
	\begin{subfigure}[t]{\sc\textwidth}
		\centering
		\shifttext{\shift}{\includegraphics[width=\linewidth]{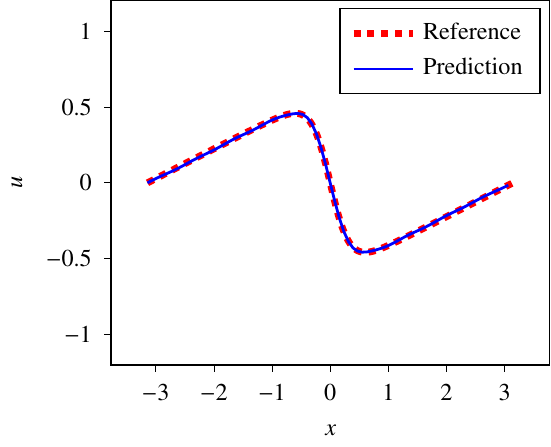}}
		\caption{$u(x,t=4)$}
		\label{fig:pred_Burger4}
	\end{subfigure}%
	\caption{Viscous Burgers' equation: Comparison of the true solution and the learned model solution at $t=2$ and $t=4$ given by the GDSG method.} 
	\label{fig:pred_Burger}
\end{figure}

We adopt a deep multi-step recursive OSG-Net with $4$ blocks, each of which contains $3$ hidden layers of equal width of $30$ neurons. The parameters in the GDSG method are set as $\lambda=2$ and $Q=5$. We train the network for up to $20,000$ epochs with a batch size of $90$. We validate the trained model on $100$ test trajectories with forward steps $M=80$. Table \ref{tab:resume_Burger} displays the average relative 
errors and the standard deviation. We observe the superior performance of the GDSG method over the LISG and baseline methods. 
We also employ $u(x,0)=-\text{sin}(x)$ as the initial condition and predict the solution at $t=2$ and $t=4$ by using the trained model of the GDSG method; see Figure \ref{fig:pred_Burger}. 
It is seen that the predicted solution preserves 
the symmetry about $x=0$ and agrees well with the reference solution. 
 
\begin{table}[ht!]
	\caption{Viscous Burgers' equation - Average prediction error $\overline{\mathcal{E}}$ on the test set and the standard deviation $\sigma$ of prediction error.}
	\label{tab:resume_Burger}
	\begin{center}
		\begin{tabular}{llll}
			\toprule
			\multicolumn{1}{c}{}                            &Baseline             &LISG          &GDSG  \\ 
			\midrule
			Prediction error $\overline{\mathcal{E}}$                 &$0.5235$    &$5.653\times 10^{-2}$   &$\mathbf{2.648\times 10^{-2}}$         \\
			Standard deviation $\sigma$                   &$0.261$    &$2.04\times 10^{-3}$   &$\mathbf{7.36\times 10^{-4}}$           \\
			\bottomrule
		\end{tabular}
	\end{center}
\end{table}

\subsubsection{PDE example 3: Inviscid Burgers' equation}
In this example, we consider the inviscid Burgers' equation with periodic boundary condition: 
\begin{equation}
	\begin{dcases}
			\partial_t u +  \partial_x \left(\frac{u^2}{2}\right)= 0, \quad & (x,t)\in (-\pi,\pi)\times \mathbb{R}^+,\\
			u(-\pi,t) = u(\pi,t), & t\in \mathbb{R}^+.
	\end{dcases}
	\label{equation:eg_ivbg}
\end{equation}
This is a challenging problem due to the hyperbolic nature of the system \eqref{equation:eg_ivbg}, as its solution can produce  shocks over time even if the initial condition is smooth. 
The training data is generated by solving the true PDE \eqref{equation:eg_ivbg} using 
a ninth-order finite difference weighted essentially non-oscillatory ( WENO) scheme with the fourth-order Runge--Kutta time discretization, starting from $800$ different initial conditions 
in the following form 
\begin{equation*}
	u(x,0)=a_0 + \sum_{n=1}^{10}\big(a_n\text{cos}(nx) + b_n\text{sin}(nx)\big),
\end{equation*}
where the Fourier coefficients $\{a_n,b_n\}$ are randomly sampled {from} uniform distributions on pre-defined intervals, with $a_0\sim U[-\frac12,\frac12]$ and $a_n,b_n \sim U[-\frac{1}{n}, \frac{1}{n}]$. Starting from each initial condition, we collect $10$ snapshot data over time with the stepsizes randomly drawn from the interval $[0.05,0.15]$. 
The data are sampled on a uniform mesh of $64$ nodal points.  
Note that our dataset only contains smooth solutions, and its size is much smaller than that used in \cite{chen2022deep}, where $20,000$ trajectory data were used with $10,000$ data containing discontinuous solutions. 

\begin{figure}[ht!]
	\def\sc{0.33}
	\def\shift{-7mm}
	\centering
	\begin{subfigure}[t]{\sc\textwidth}
		\centering
		\shifttext{\shift}{\includegraphics[width=\linewidth]{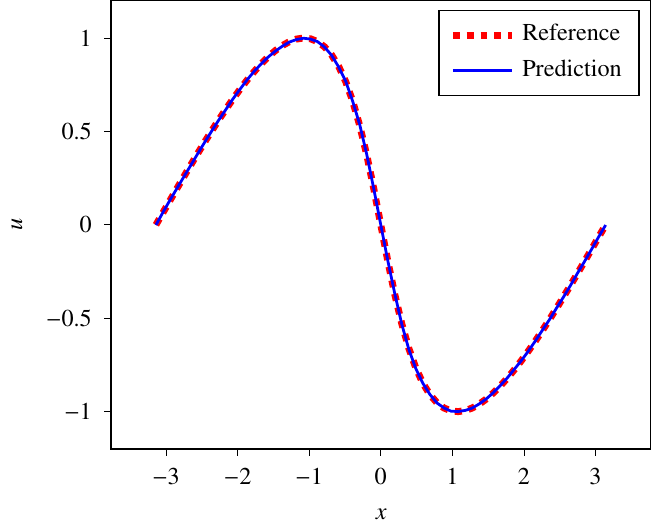}}
		\caption{$u(x,t=0.5)$}
	\end{subfigure}\quad
	\begin{subfigure}[t]{\sc\textwidth}
		\centering
		\shifttext{\shift}{\includegraphics[width=\linewidth]{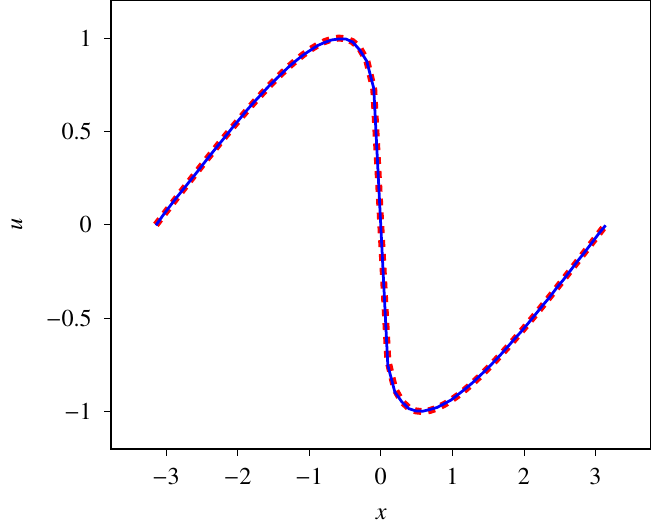}}
		\caption{$u(x,t=1.0)$}
	\end{subfigure}%
	
	\smallskip
	
	\begin{subfigure}[t]{\sc\textwidth}
		\centering
		\shifttext{\shift}{\includegraphics[width=\linewidth]{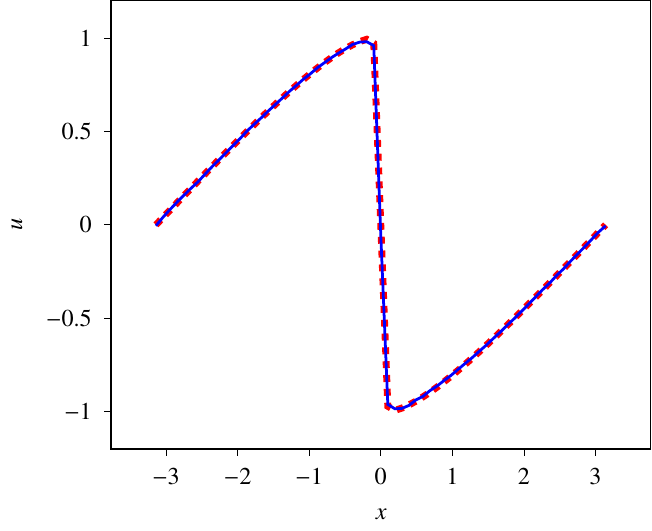}}
		\caption{$u(x,t=1.5)$}
	\end{subfigure}\quad
	\begin{subfigure}[t]{\sc\textwidth}
		\centering
		\shifttext{\shift}{\includegraphics[width=\linewidth]{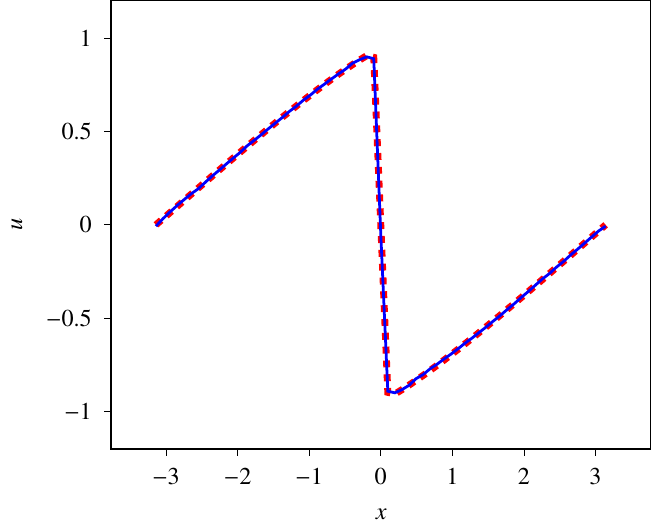}}
		\caption{$u(x,t=2.0)$}
	\end{subfigure}
	\caption{Inviscid Burgers' equation: Comparison of the reference solution and the learned model solution given by the GDSG method at $t = 0.5, 1.0, 1.5, 2.0$.}
	\label{fig:pred_inviscidBurger}
\end{figure}

In this example, we conduct the learning task of the evolution operator in nodal space, on the mesh of $64$ uniform {grids}.  
We employ the OSG-Net with the special deep neural network proposed in \cite{chen2022deep} as the basic block. The parameters in the loss function of the GDSG method are set as $\lambda=2$ and $Q=5$. The network is trained for up to $5,000$ epochs with a batch size equal to $90$. We then validate the trained models of the baseline, LISG, and GDSG methods on $100$ test trajectories with $M=20$. The average prediction errors $\overline{\mathcal{E}}$ and the standard deviation $\sigma$ are presented in 
Table \ref{tab:resume_inviscid}. We see that the LISG and GDSG methods produce more accurate and robust predictions than the baseline, while the GDSG method outperforms the LISG method. 
We also use an initial condition $u(x,0)=-\sin(x)$ and conduct simulations using the trained model of the GDSG method for time up to $t=2$. 
The predicted solutions at several different time instances are plotted in Figure \ref{fig:pred_inviscidBurger}, along with the reference solution. 
As we can see, the solution starts to develop shock at $x=0$ when $t\ge1$, 
and the prediction results agree well with the reference solution. 
It is worth noting that our model is trained to learn the evolution operator based on only smooth solution data and without any knowledge of the true equation. Interestingly, 
the learned evolution operator is able to produce shock structure developed over time.    

\begin{table}[ht!]
	\caption{Inviscid Burgers' equation: Average prediction error $\overline{\mathcal{E}}$ on the test set and the standard deviation $\sigma$ of prediction errors.}
	\label{tab:resume_inviscid}
	\begin{center}
		\begin{tabular}{llll}
			\toprule
			\multicolumn{1}{c}{}                            &Baseline             &LISG          &GDSG  \\
			\midrule
			Prediction error $\overline{\mathcal{E}}$                 &$3.024\times 10^{-2}$    &$2.957\times 10^{-2}$   &$\mathbf{2.433\times 10^{-2}}$         \\
			Standard deviation $\sigma$                   &$0.0118$    &$8.62\times 10^{-3}$   &$\mathbf{7.31\times 10^{-3}}$           \\
			\bottomrule
		\end{tabular}
	\end{center}
\end{table}

\subsubsection{PDE example 4: Two-dimensional convection-diffusion equation}

To demonstrate the applicability of our Deep-OSG approach for multidimensional problems, we consider a two-dimensional  convection-diffusion equation \cite{wu2020data} with periodic boundary conditions: 
\begin{equation}
	\begin{cases}
			\partial_t u + \alpha_1 \partial_{x}u + \alpha_2 \partial_{y}u= \sigma_1 \partial_{xx}u + \sigma_2 \partial_{yy}u, \quad & (x,y,t)\in (-\pi,\pi)\times (-\pi,\pi)\times \mathbb{R}^+,\\
			u(-\pi,y, t) = u(\pi,y, t), & (y,t)\in (-\pi,\pi)\times \mathbb{R}^+,\\
			\partial_x u(-\pi,y, t) = \partial_x u(\pi,y, t), & (y,t)\in (-\pi,\pi)\times \mathbb{R}^+,\\
			u(x,-\pi, t) = u(x,\pi, t), & (x,t)\in (-\pi,\pi)\times \mathbb{R}^+,\\
			\partial_y u(x,-\pi, t) = \partial_y u(x, \pi, t), & (x,t)\in (-\pi,\pi)\times \mathbb{R}^+,
	\end{cases}
	\label{equation:eg_conv_diff}
\end{equation}
where the parameters are taken as $(\alpha_1, \alpha_2)=(1,0.7)$ and $(\sigma_1, \sigma_2)=(0.1,0.16)$. 

\begin{figure}[ht!]
	\def\sc{0.25}
	\def\sC{0.05}
	\def\SC{0.8}
	\def\shift{-10pt}
	\captionsetup[subfigure]{labelformat=empty}
	\centering
	
	\begin{subfigure}[c]{\sc\textwidth}
		\centering
		\includegraphics[width=\SC\linewidth]{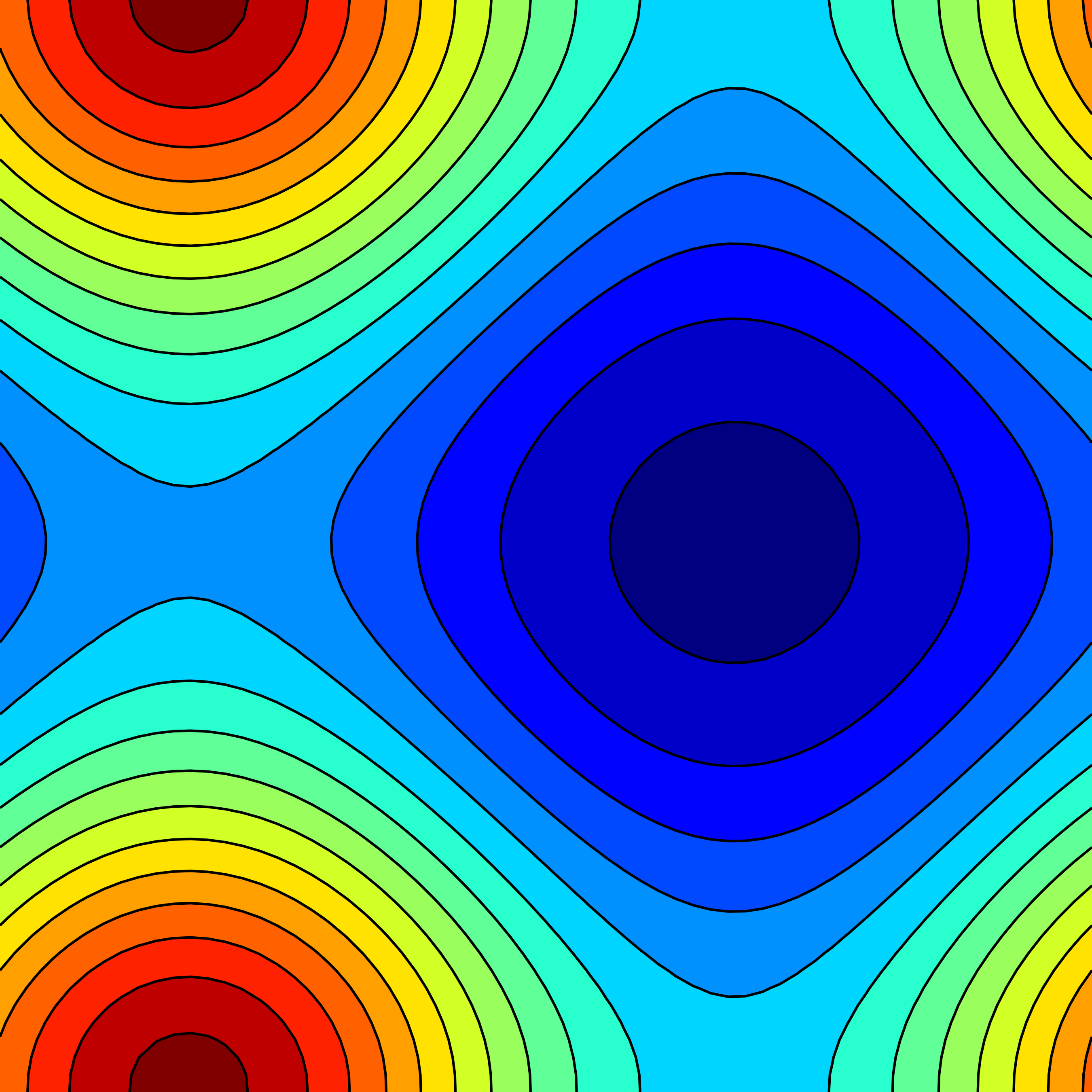}
		\caption{True, $u(x,y,t=1.5)$}
	\end{subfigure}\hspace*{\shift}%
	\begin{subfigure}[c]{\sC\textwidth}
		\centering
		\includegraphics{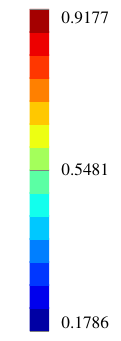}
		\caption{}
	\end{subfigure}\quad
	\begin{subfigure}[c]{\sc\textwidth}
		\centering
		\includegraphics[width=\SC\linewidth]{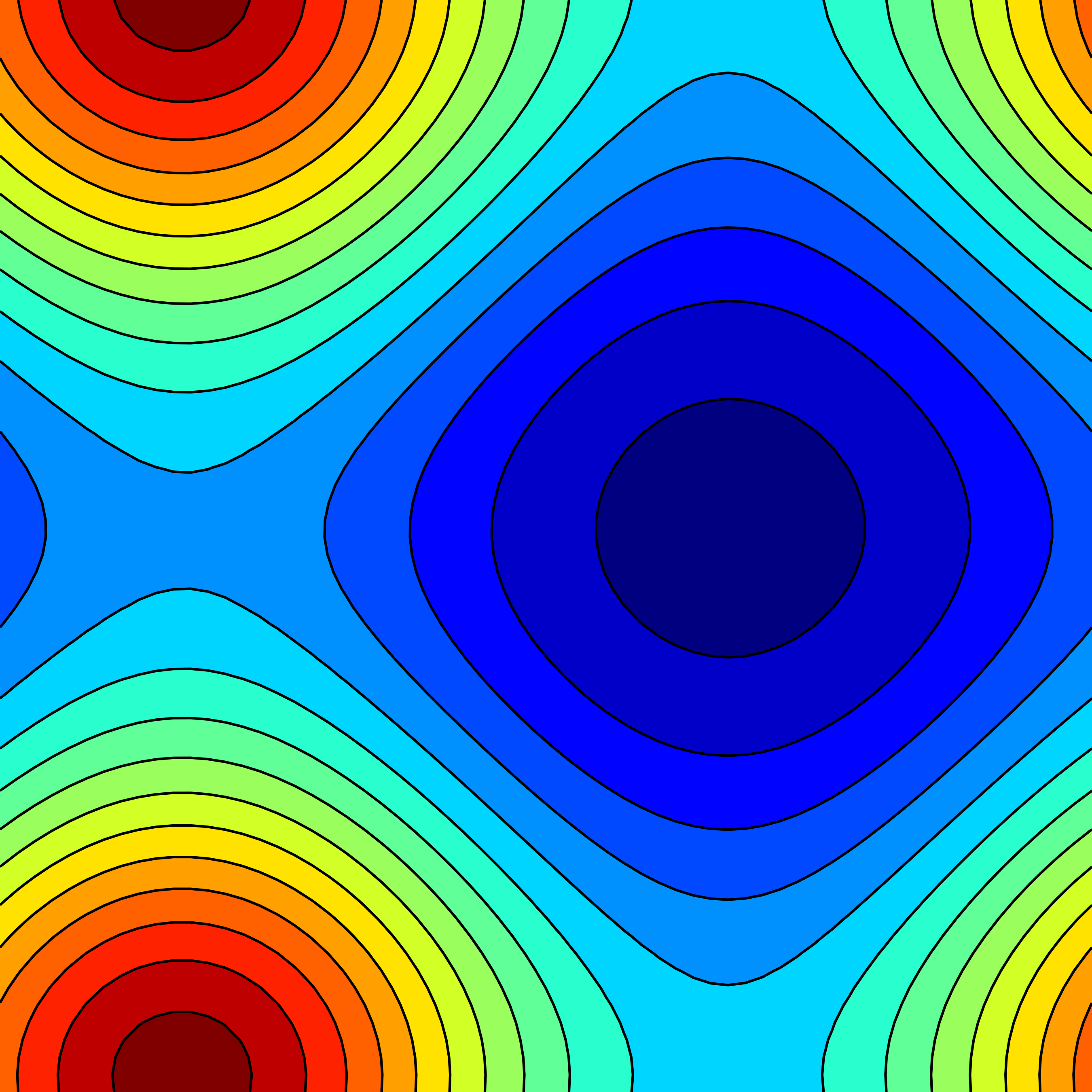}
		\caption{GDSG, $u^{pre}(x,y,t=1.5)$}
	\end{subfigure}\hspace*{\shift}%
	\begin{subfigure}[c]{\sC\textwidth}
		\centering
		\includegraphics{./Figures/convdiff_color_u1.5.pdf}
		\caption{}
	\end{subfigure}\quad
	\begin{subfigure}[c]{\sc\textwidth}
		\centering
		\includegraphics[width=\SC\linewidth]{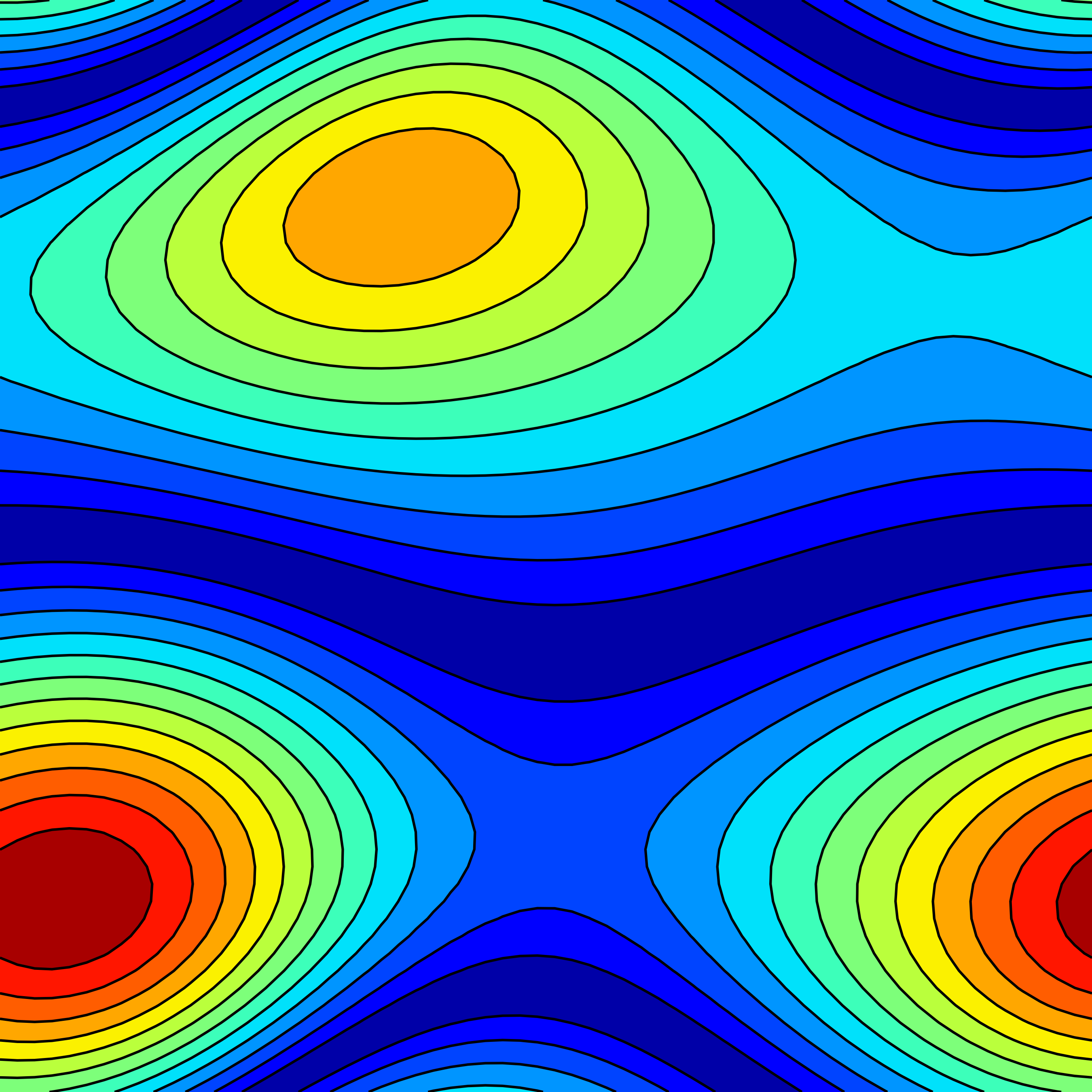}
		\caption{GDSG, absolute error $t=1.5$}
	\end{subfigure}\hspace*{\shift}%
	\begin{subfigure}[c]{\sC\textwidth}
		\centering
		\includegraphics{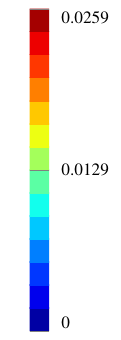}
		\caption{}
	\end{subfigure}%
	
	\smallskip
	
	\begin{subfigure}[c]{\sc\textwidth}
		\centering
		\includegraphics[width=\SC\linewidth]{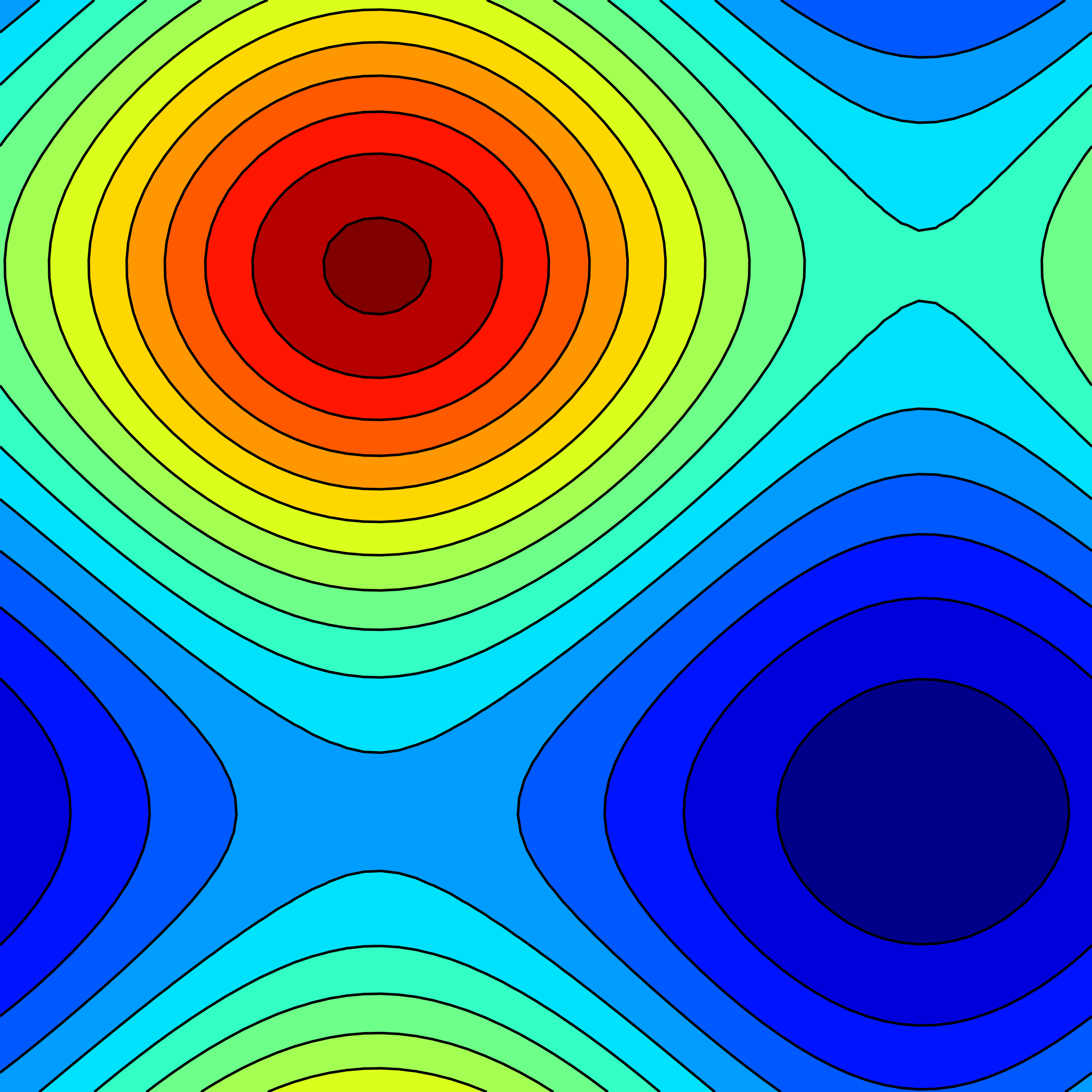}
		\caption{True, $u(x,y,t=3)$}
	\end{subfigure}\hspace*{\shift}%
	\begin{subfigure}[c]{\sC\textwidth}
		\centering
		\includegraphics{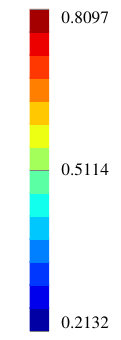}
		\caption{}
	\end{subfigure}\quad
	\begin{subfigure}[c]{\sc\textwidth}
		\centering
		\includegraphics[width=\SC\linewidth]{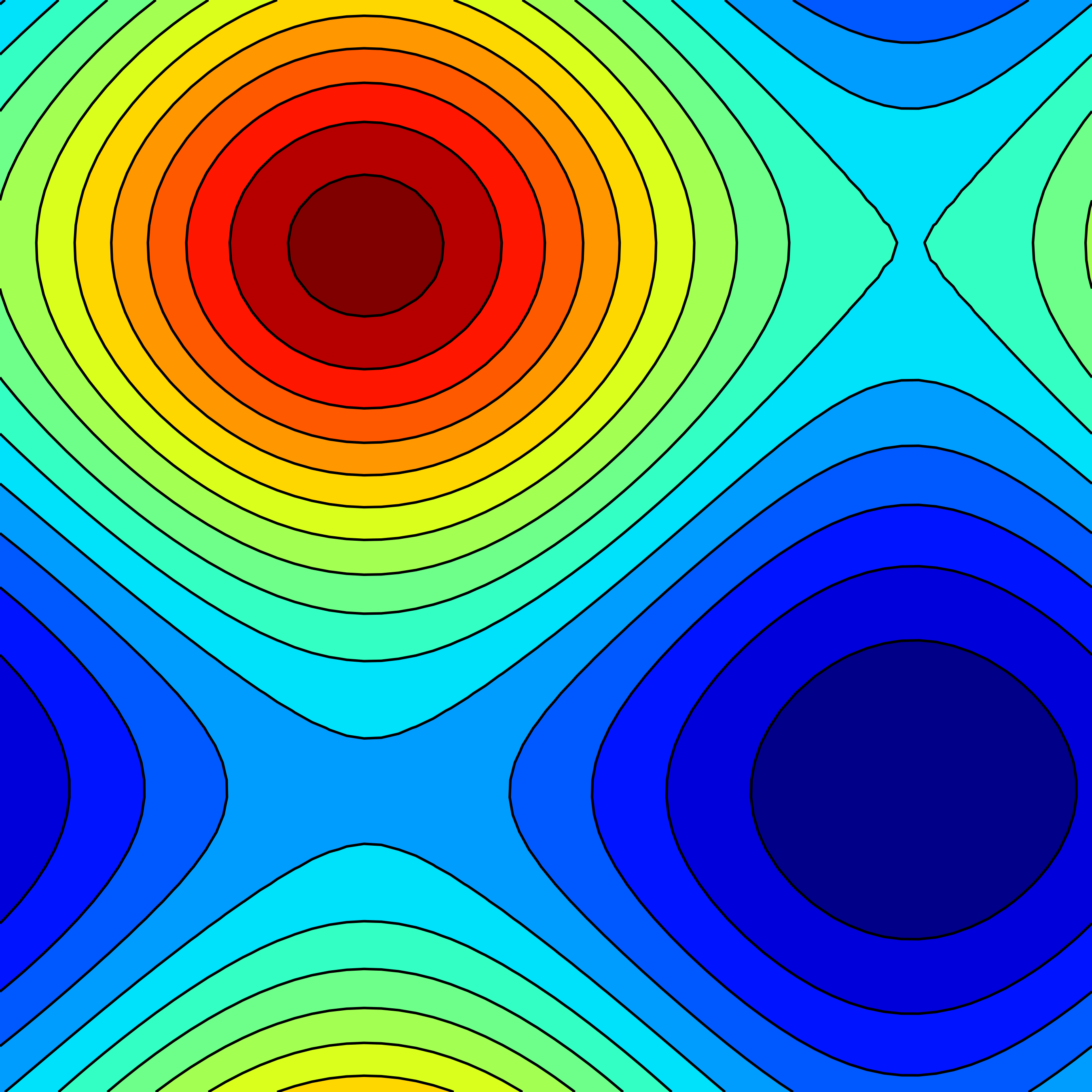}
		\caption{GDSG, $u^{pre}(x,y,t=3)$}
	\end{subfigure}\hspace*{\shift}%
	\begin{subfigure}[c]{\sC\textwidth}
		\centering
		\includegraphics{./Figures/convdiff_color_u3.pdf}
		\caption{}
	\end{subfigure}\quad
	\begin{subfigure}[c]{\sc\textwidth}
		\centering
		\includegraphics[width=\SC\linewidth]{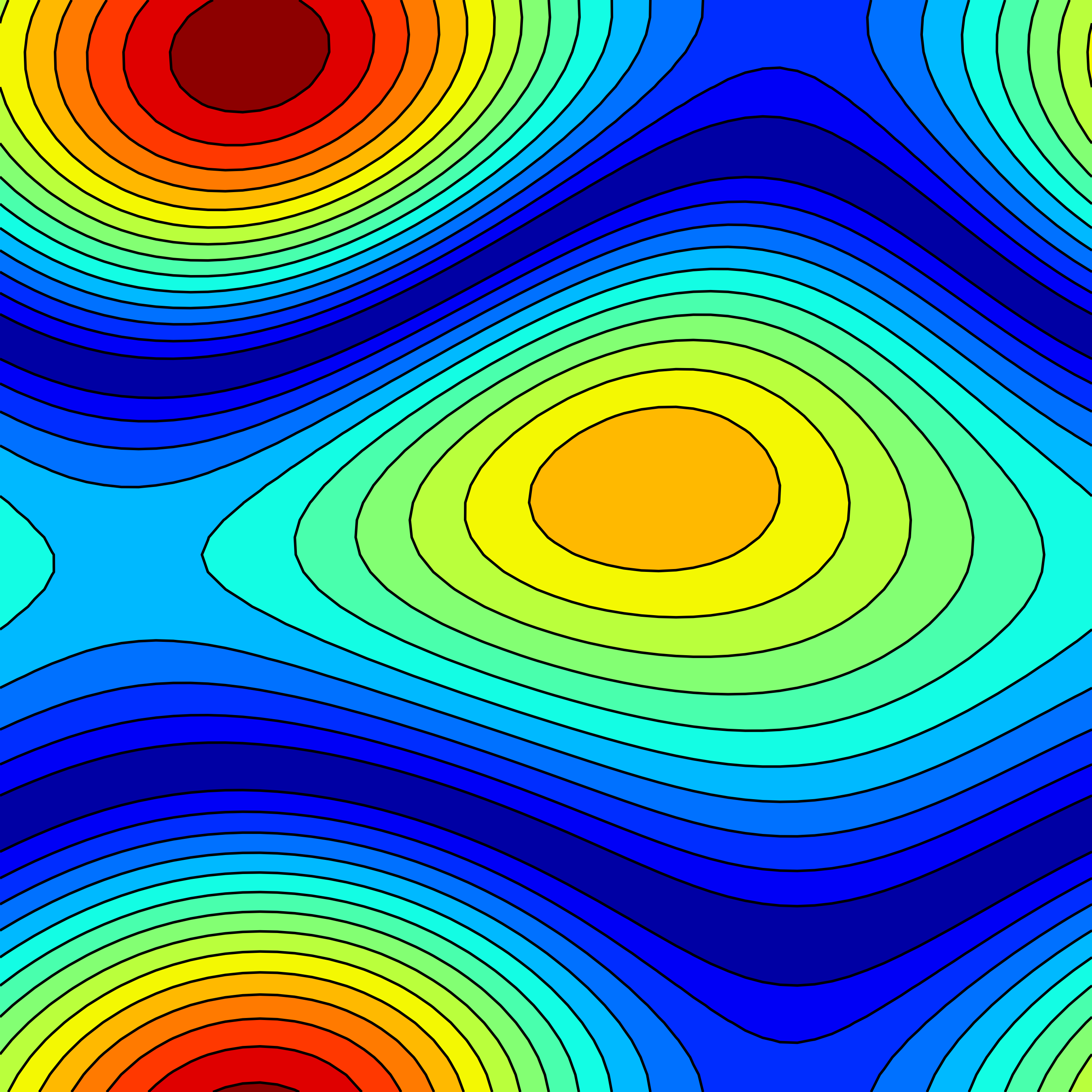}
		\caption{GDSG, absolute error $t=3$}
	\end{subfigure}\hspace*{\shift}%
	\begin{subfigure}[c]{\sC\textwidth}
		\centering
		\includegraphics{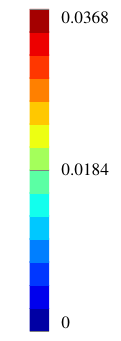}
		\caption{}
	\end{subfigure}%
	\caption{Two-dimensional convection-diffusion equation: Contour plots of the solutions and errors at $t=1.5$ and $t=3$. Left: true solution; middle: learned model solution given by the GDSG method; right: absolute error in the learned model solution.}
	\label{fig:pred_conv_diff}
\end{figure}

The learning task {for} this problem is conducted in modal space, with the finite-dimensional 
approximation space $\mathbb V_n$ taken as the span of $n=25$ basis functions as in \cite{wu2020data}.  
The training data consists of $J=2,000$ bursts of short ``trajectories'' with two forward time stepsizes $\{\Delta_{1,j},\Delta_{2,j}\}_{j=1}^J$ randomly sampled in the interval  $[0.05,0.15]$. 
Our dataset is much smaller than that employed in \cite{wu2020data}, where $1,000,000$ data pairs were used for training.  
For neural network modeling, we adopt a deep multi-step recursive OSG-Net with $3$ blocks, and each block contains $3$ hidden layers of equal width of $40$ neurons. The parameters in the loss function of the GDSG method are set as $\lambda=2$ and $Q=5$. 
After training the network for up to $500$ epochs with a batch size of $90$, 
we validate the trained models on $100$ test trajectories with $M=30$. 
The results of the baseline, LISG, and GDSG methods are listed and compared in Table \ref{tab:resume_conv_diff}. 
Again, we observe that the model obtained by the GDSG and LISG methods produce more accurate and robust predictions than the baseline method, while the performance of the GDSG method is the best. For further validation, we take an initial condition from the test set and present the contours of the prediction solution of the GDSG method at $t=1.5$ and $t=3$ in Figure \ref{fig:pred_conv_diff}, along with the true solution for comparison. One can see 
 good agreement between the predicted and true solutions.  

\begin{table}[ht!]
	\caption{Two-dimensional convection-diffusion equation: Average prediction error $\overline{\mathcal{E}}$ and the standard deviation $\sigma$ of prediction errors.}
	\label{tab:resume_conv_diff}
	\begin{center}
		\begin{tabular}{llll}
			\toprule
			\multicolumn{1}{c}{}                            &Baseline             &LISG          &GDSG  \\
			\midrule
			Prediction error $\overline{\mathcal{E}}$                 &$1.063\times 10^{-2}$    &$1.104\times 10^{-2}$   &$\mathbf{9.093\times 10^{-3}}$         \\
			Standard deviation $\sigma$                   &$1.42\times 10^{-3}$    &$7.28\times 10^{-4}$   &$\mathbf{2.68\times 10^{-4}}$           \\
			\bottomrule
		\end{tabular}
	\end{center}
\end{table}

\subsubsection{PDE example 5: Two-dimensional Navier--Stokes equations}
In the last example, we consider the two-dimensional incompressible Navier--Stokes equations:
\begin{equation}
	\begin{cases}
			\partial_t \omega (\bm{x},t) + {\bf u}(\bm{x},t) \cdot \nabla \omega (\bm{x},t) = \nu \Delta \omega(\bm{x},t) + f(\bm{x}), \quad  & \bm{x}\in (0,1)^2, t> 0,\\
			\nabla \cdot {\bf u}(\bm{x},t) = 0, & \bm{x}\in (0,1)^2, t>0,\\
			\omega (\bm{x},t=0) = \omega_0(\bm{x}),& \bm{x}\in (0,1)^2,
	\end{cases}
	\label{equation:eg_ns}
\end{equation}
where $\bm{x}=(x_1,x_2)$ represents the spatial coordinates, ${\bf u}(\bm{x},t)$ is the velocity field, $\omega = \nabla \times {\bf u}$ is the vorticity, $\nu=10^{-3}$ denotes the viscosity, and $f({\bm x})=0.1(\sin (2\pi(x_1+x_2)) + \cos (2\pi(x_1+x_2)))$ stands for a periodic external force. All the data are generated on a $64\times 64$ Cartesian grid. For the training data, we sample $100$ different initial conditions $\omega_0(\bm{x}) \sim \mu$ where $\mu = N(0,7^3(-\Delta + 49 I)^{-2.5})$ with periodic boundary conditions {following} \cite{li2021fourier}. For each initial condition, we collect the snapshot data at 
$M=110$ time instances with the time lags randomly drawn from the interval $[0.5,1.5]$. To reduce the impact of Gaussian initialization, we only use the last $51$ snapshots as training data. 

In this example, we learn the evolution operator of the vorticity $\omega$ in nodal space, on the  $64\times 64$ uniform grid. 
For neural network modeling, we employ the Fourier neural operator proposed in \cite{li2021fourier} as the basic block to construct our OSG-Net. The input of such an OSG-Net is a tensor with shape $2\times 64\times 64$. The first channel {corresponds to} the vorticity field $\omega_\text{in} \in \mathbb{R}^{64\times 64}$, and the second channel contains repeated time step size $\Delta$ at each grid point. The output tensor is a single-channel tensor, \textit{i.e.}~the predicted vorticity field $\omega_\text{out}\in \mathbb{R}^{64\times 64}$. 
For less memory usage and faster training, we set $Q=1$ in the loss function of the GDSG method, while the parameter $\lambda$ is set as $0.5$. 
The OSG-Net is trained for up to $1,000$ epochs with a batch size of $20$. 

\begin{figure}[ht!]
	\centering
	\includegraphics[width=0.675\linewidth]{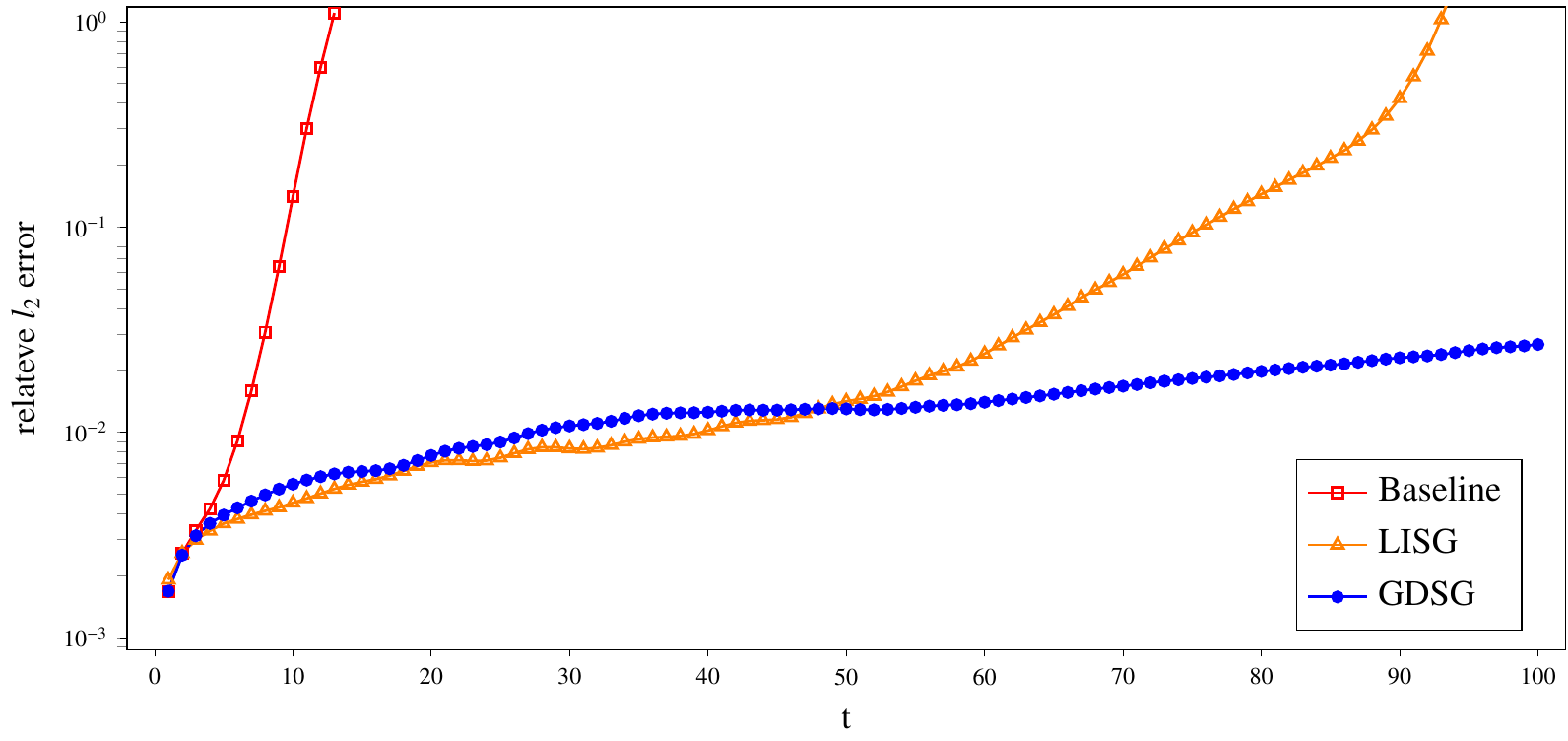}
	\caption{Navier--Stokes equations: Evolution of the prediction errors over time. The curves of baseline and LISG methods are not shown completely as the errors blow up.}
	\label{fig:err_ns}
\end{figure}

\begin{figure}[ht!]
	\captionsetup[subfigure]{labelformat=empty}
	\centering
	\def\sc{0.25}
	\def\sC{0.05}
	\def\SC{0.8}
	\def\shift{-10pt}
	\begin{subfigure}[c]{\sc\textwidth}
		\centering
		\includegraphics[width=\SC\linewidth]{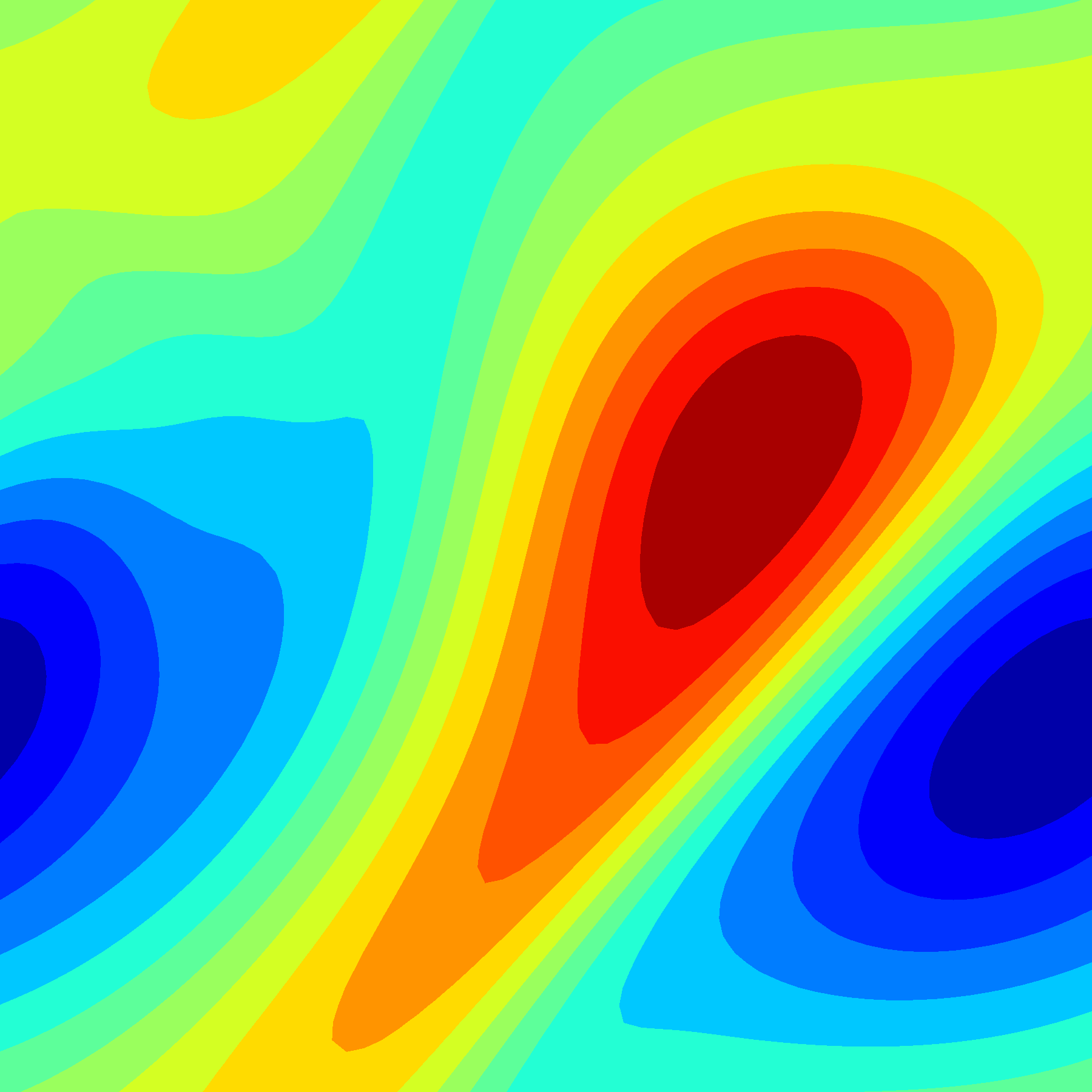}
		\caption{True, $t=60$}
	\end{subfigure}\hspace*{\shift}%
	\begin{subfigure}[c]{\sC\textwidth}
		\centering
		\includegraphics{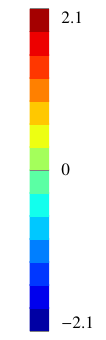}
		\caption{}
	\end{subfigure}\quad
	\begin{subfigure}[c]{\sc\textwidth}
		\centering
		\includegraphics[width=\SC\linewidth]{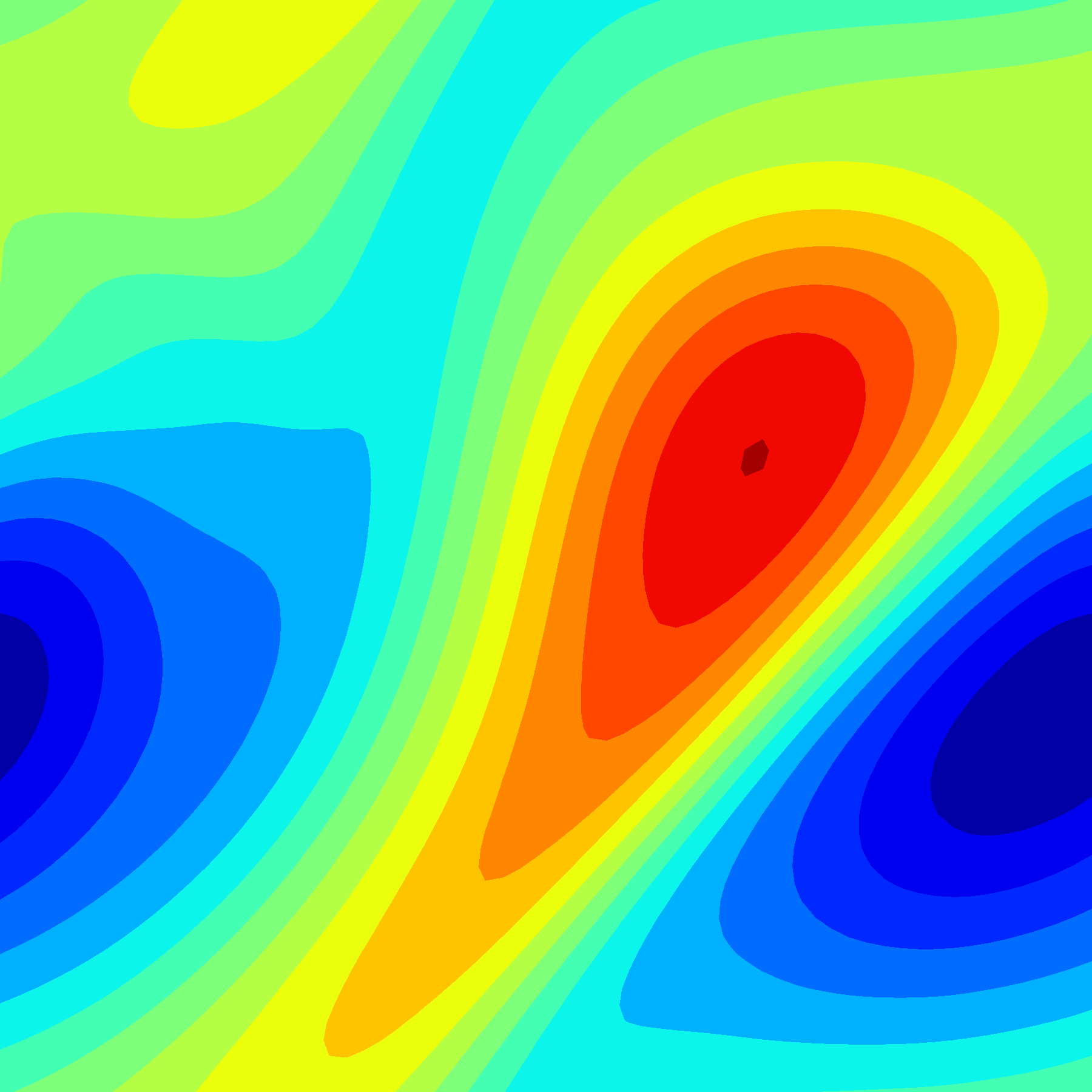}
		\caption{LISG, $t=60$}
	\end{subfigure}\hspace*{\shift}%
	\begin{subfigure}[c]{\sC\textwidth}
		\centering
		\includegraphics{./Figures/ns_color_w.pdf}
		\caption{}
	\end{subfigure}\quad
	\begin{subfigure}[c]{\sc\textwidth}
		\centering
		\includegraphics[width=\SC\linewidth]{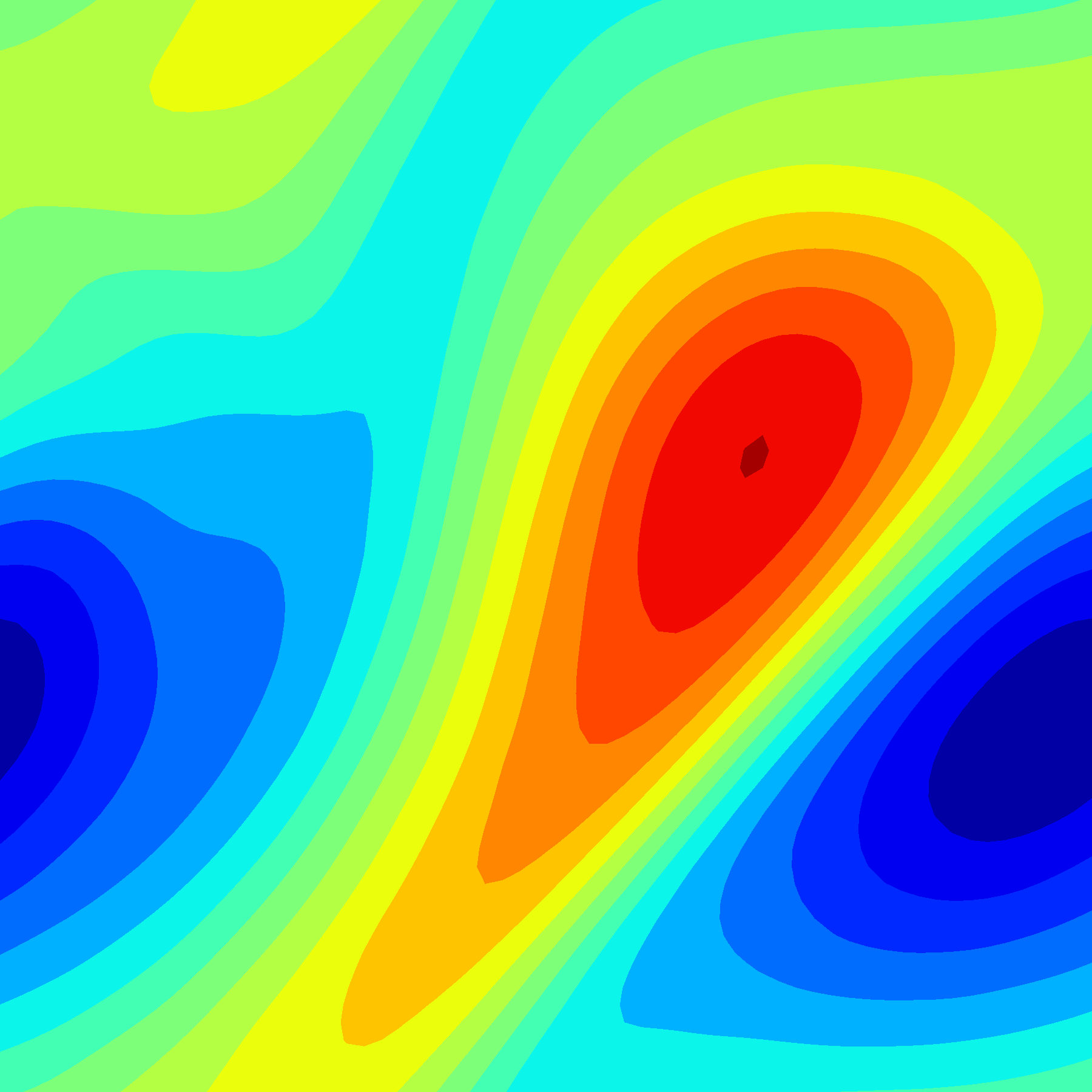}
		\caption{GDSG, $t=60$}
	\end{subfigure}\hspace*{\shift}%
	\begin{subfigure}[c]{\sC\textwidth}
		\centering
		\includegraphics{./Figures/ns_color_w.pdf}
		\caption{}
	\end{subfigure}%
	
	\smallskip
	
	\begin{subfigure}[c]{\sc\textwidth}
		\centering
		\includegraphics[width=\SC\linewidth]{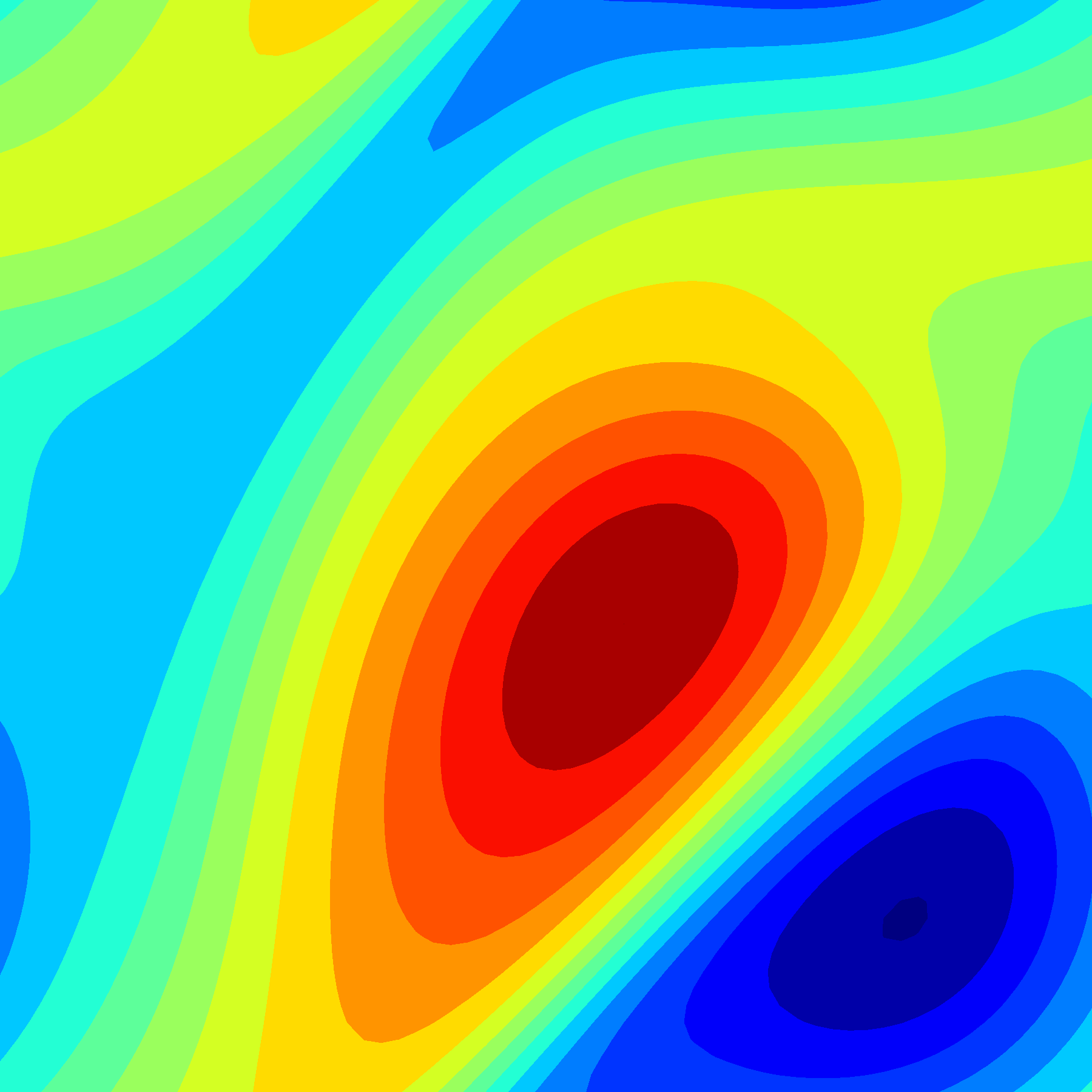}
		\caption{True, $t=80$}
	\end{subfigure}\hspace*{\shift}
	\begin{subfigure}[c]{\sC\textwidth}
		\centering
		\includegraphics{./Figures/ns_color_w.pdf}
		\caption{}
	\end{subfigure}\quad
	\begin{subfigure}[c]{\sc\textwidth}
		\centering
		\includegraphics[width=\SC\linewidth]{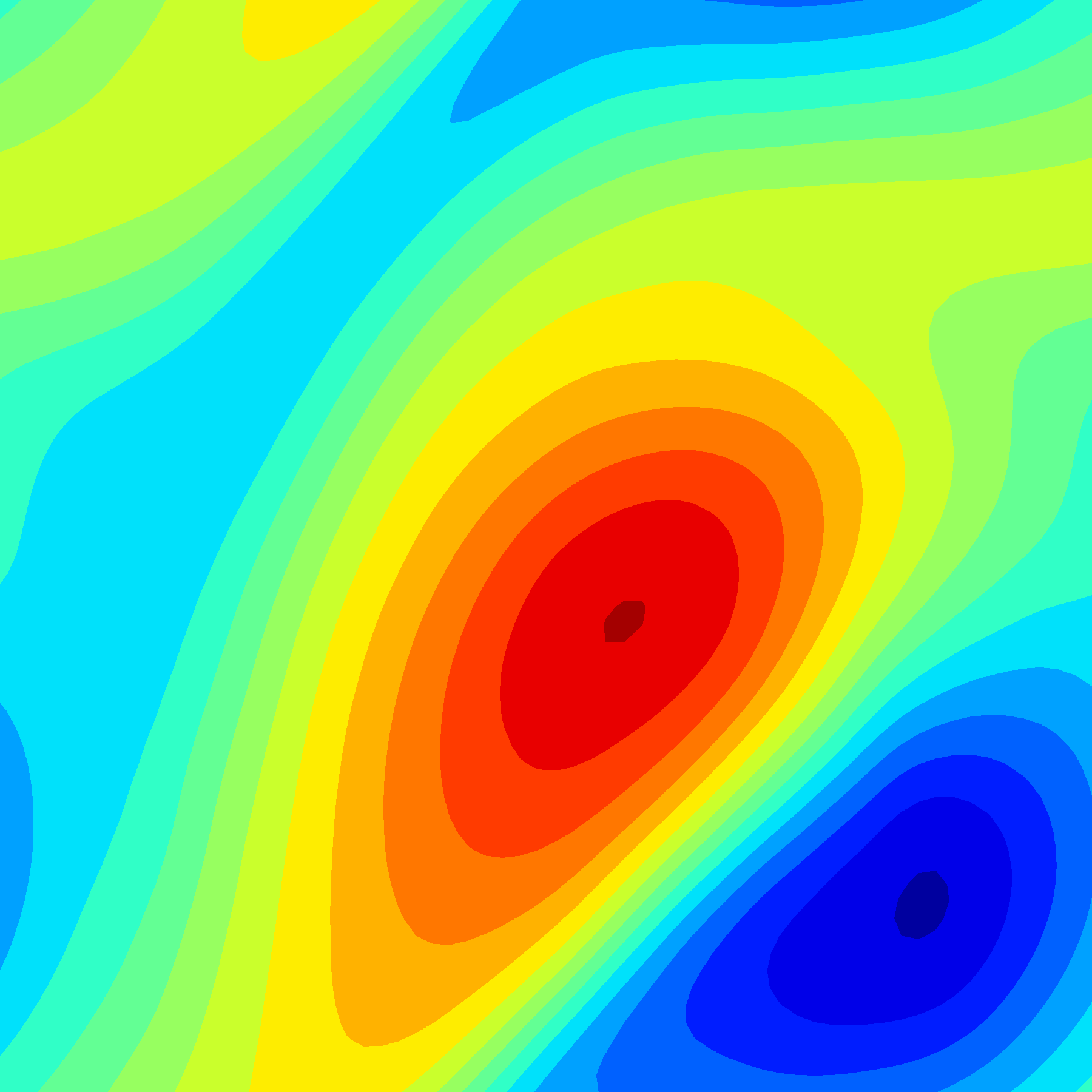}
		\caption{LISG, $t=80$}
	\end{subfigure}\hspace*{\shift}%
    \begin{subfigure}[c]{\sC\textwidth}
		\centering
		\includegraphics{./Figures/ns_color_w.pdf}
		\caption{}
	\end{subfigure}\quad
	\begin{subfigure}[c]{\sc\textwidth}
		\centering
		\includegraphics[width=\SC\linewidth]{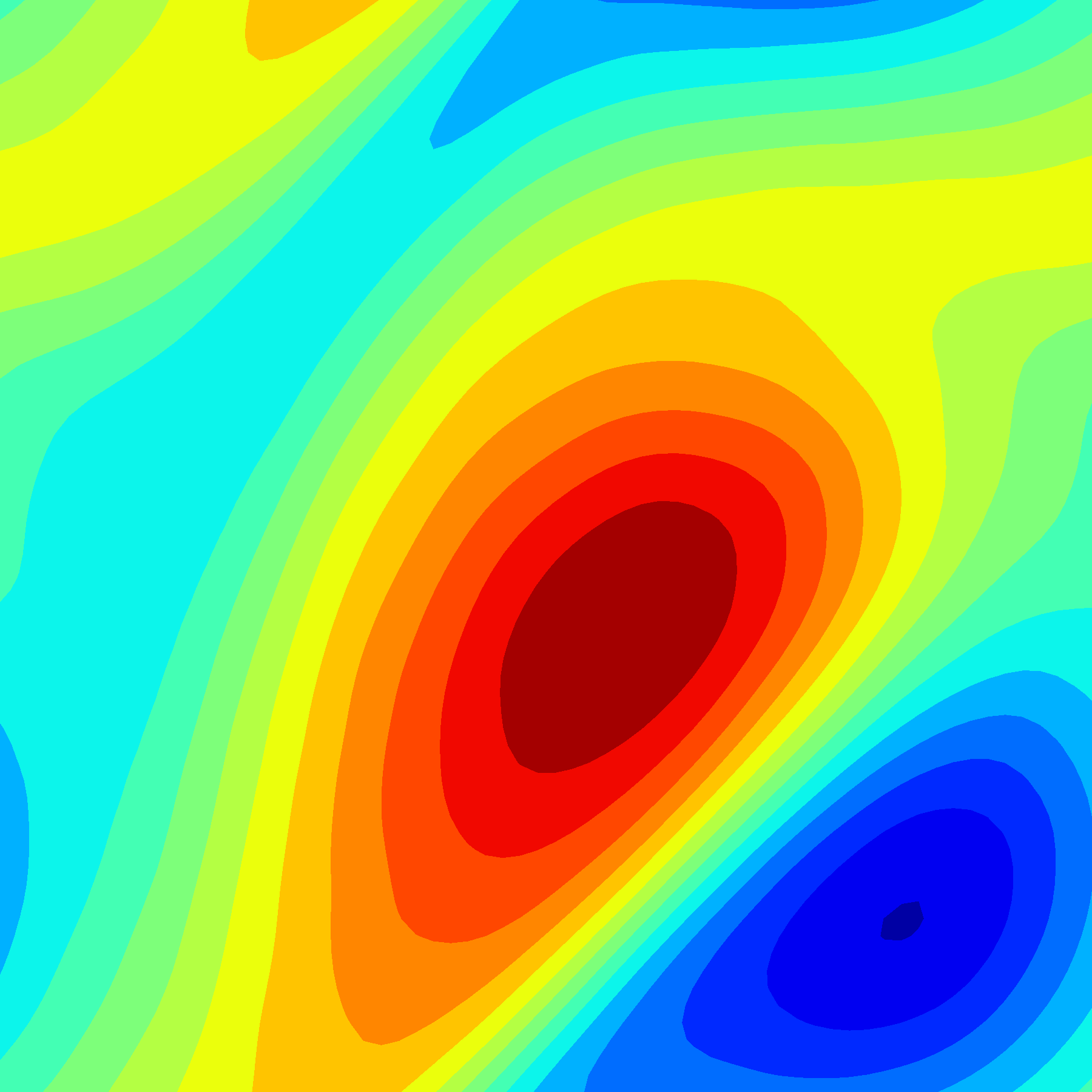}
		\caption{GDSG, $t=80$}
	\end{subfigure}\hspace*{\shift}%
	\begin{subfigure}[c]{\sC\textwidth}
		\centering
		\includegraphics{./Figures/ns_color_w.pdf}
		\caption{}
	\end{subfigure}%
	\smallskip
	
	\begin{subfigure}[c]{\sc\textwidth}
		\centering
		\includegraphics[width=\SC\linewidth]{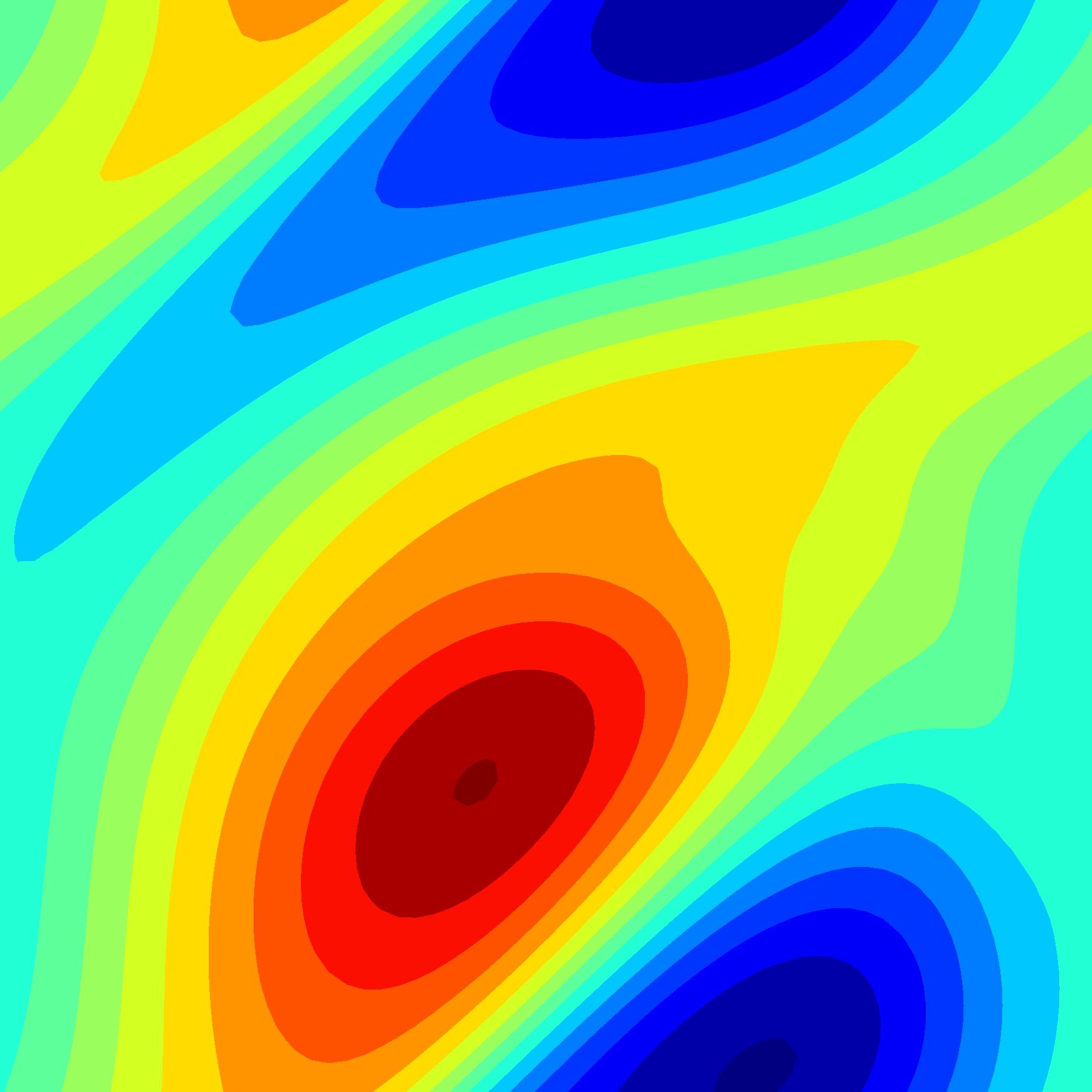}
		\caption{True, $t=100$}
	\end{subfigure}\hspace*{\shift}%
	\begin{subfigure}[c]{\sC\textwidth}
		\centering
		\includegraphics{./Figures/ns_color_w.pdf}
		\caption{}
	\end{subfigure}\quad
	\begin{subfigure}[c]{\sc\textwidth}
		\centering
		\includegraphics[width=\SC\linewidth]{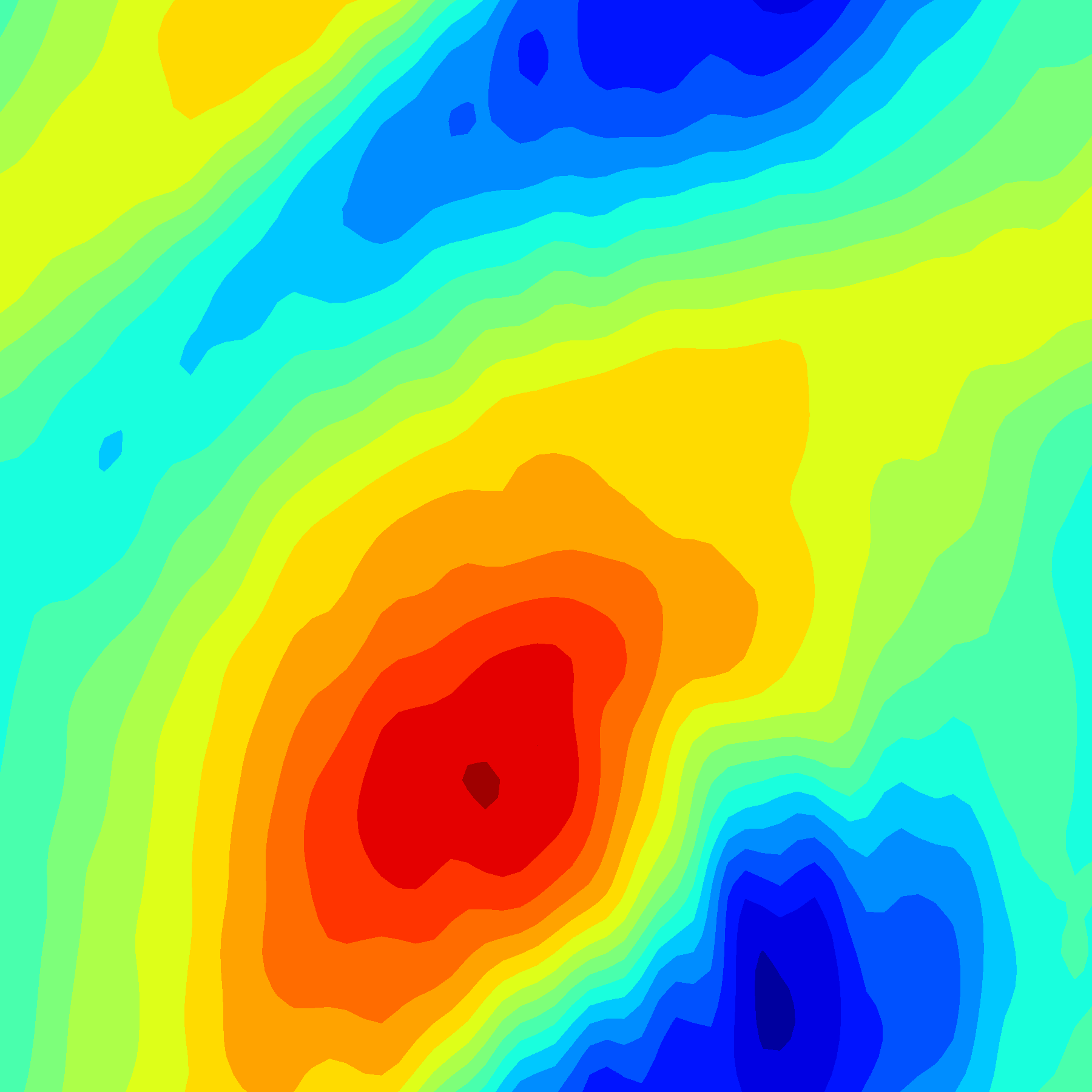}
		\caption{LISG, $t=100$}
	\end{subfigure}\hspace*{\shift}%
    \begin{subfigure}[c]{\sC\textwidth}
		\centering
		\includegraphics{./Figures/ns_color_w.pdf}
		\caption{}
	\end{subfigure}\quad
	\begin{subfigure}[c]{\sc\textwidth}
		\centering
		\includegraphics[width=\SC\linewidth]{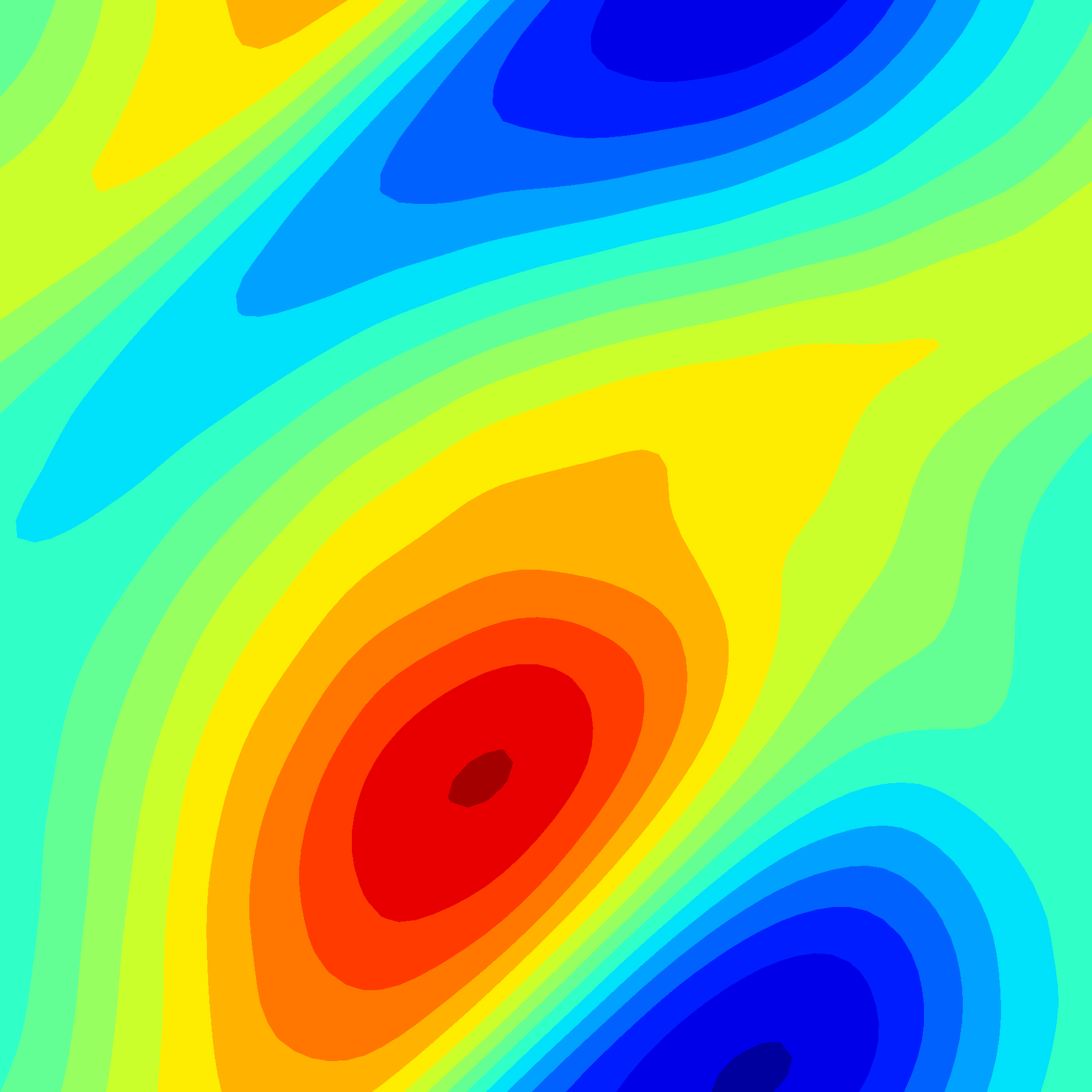}
		\caption{GDSG, $t=100$}
	\end{subfigure}\hspace*{\shift}%
	\begin{subfigure}[c]{\sC\textwidth}
		\centering
		\includegraphics{./Figures/ns_color_w.pdf}
		\caption{}
	\end{subfigure}%
	\caption{Navier--Stokes equations: 
		The true and predicted vorticity fields given by the LISG and GDSG methods at $t=60, 80, 100$.}
	\label{fig:pred_ns}
\end{figure}

\begin{figure}[ht!]
	\captionsetup[subfigure]{labelformat=empty}
	\centering
	\def\sc{0.25}
	\def\sC{0.05}
	\def\SC{0.8}
	\def\shift{-10pt}
	
	\begin{subfigure}[c]{\sc\textwidth}
		\centering
		\includegraphics[width=\SC\linewidth]{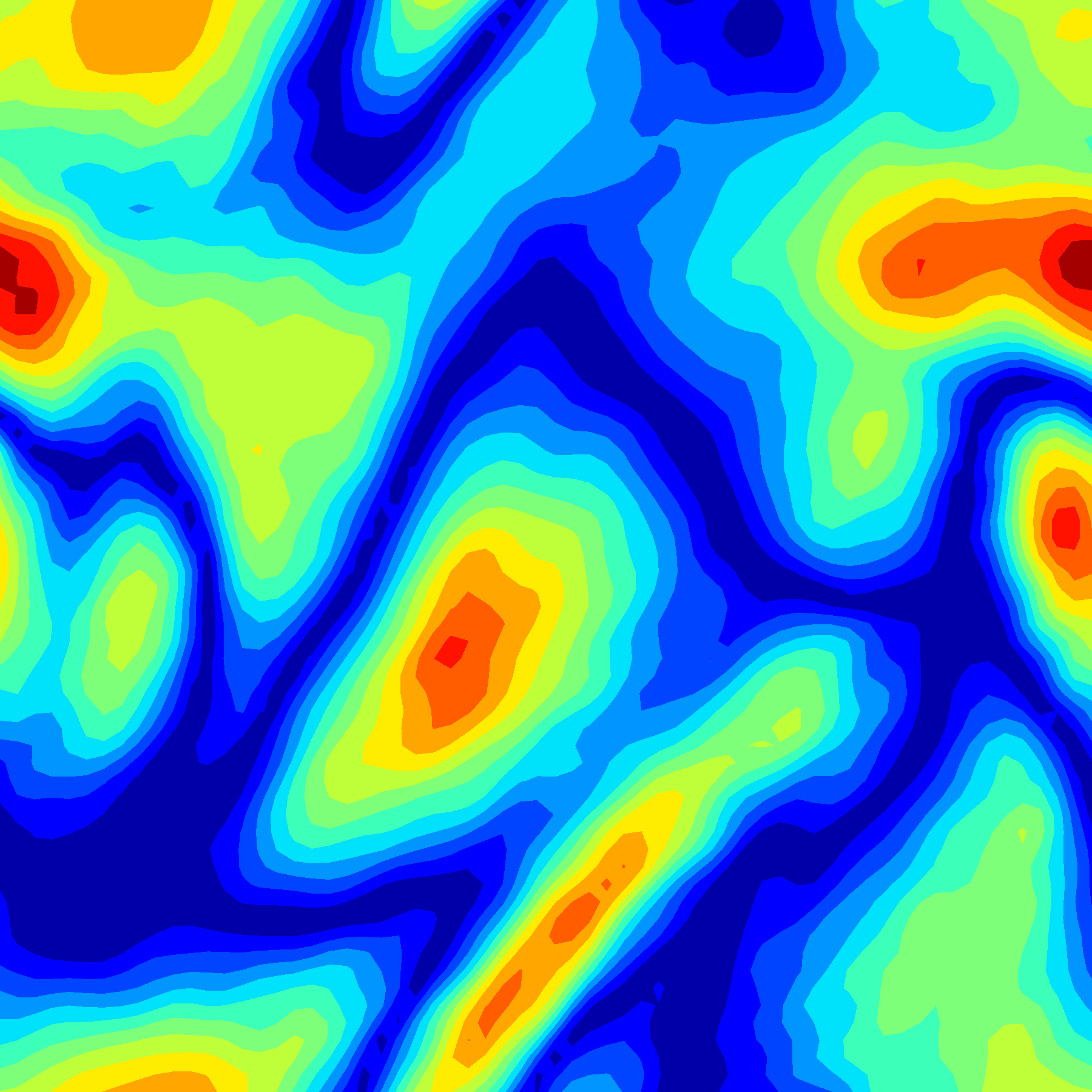}
		\caption{LISG, absolute error $t=60$}
	\end{subfigure}\hspace*{\shift}%
	\begin{subfigure}[c]{\sC\textwidth}
		\centering
		\includegraphics{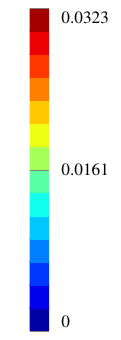}
		\caption{}
	\end{subfigure}\quad
	\begin{subfigure}[c]{\sc\textwidth}
		\centering
		\includegraphics[width=\SC\linewidth]{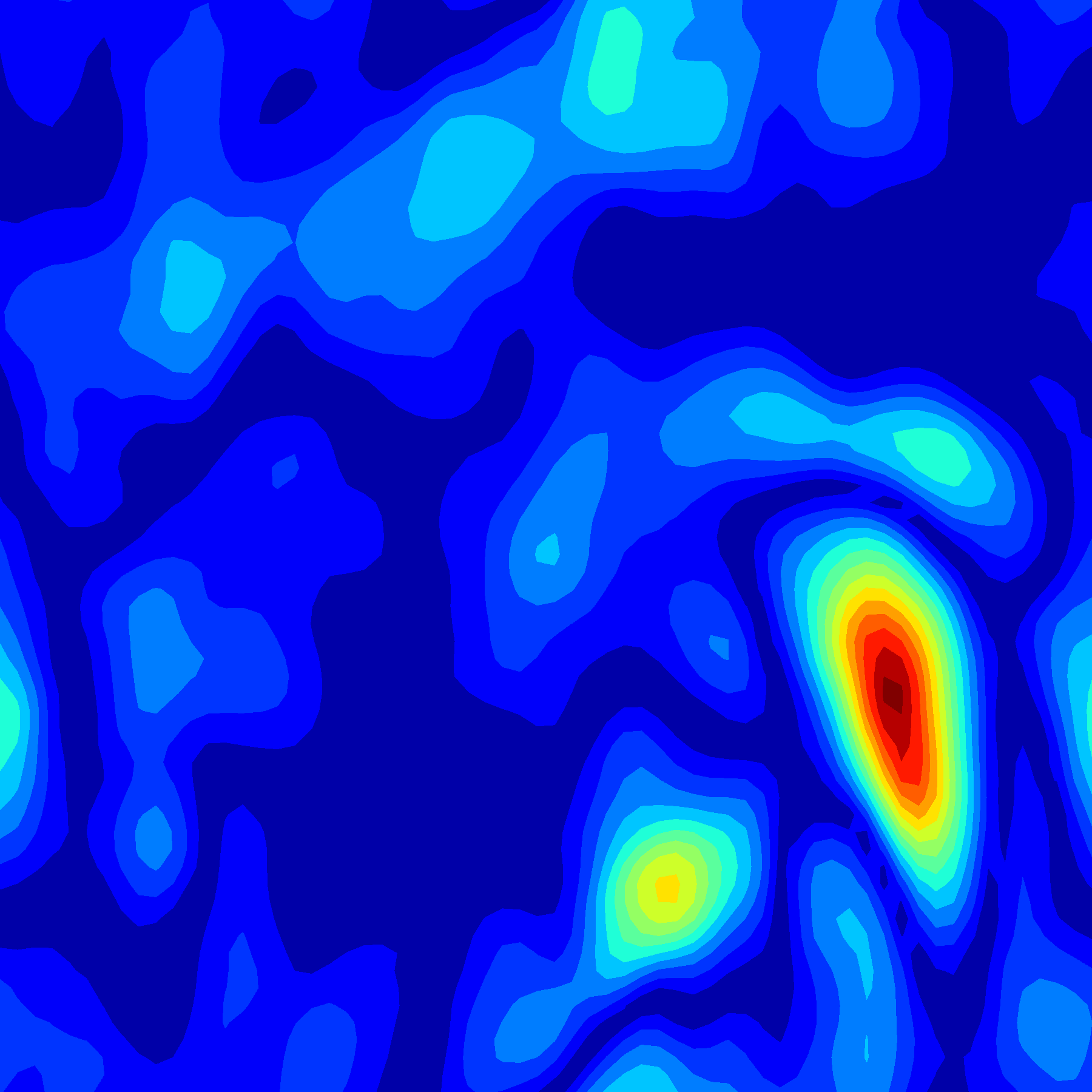}
		\caption{LISG, absolute error $t=80$}
	\end{subfigure}\hspace*{\shift}%
	\begin{subfigure}[c]{\sC\textwidth}
		\centering
		\includegraphics{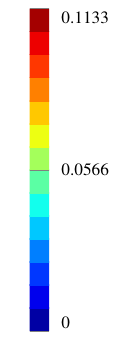}
		\caption{}
	\end{subfigure}\quad
	\begin{subfigure}[c]{\sc\textwidth}
		\centering
		\includegraphics[width=\SC\linewidth]{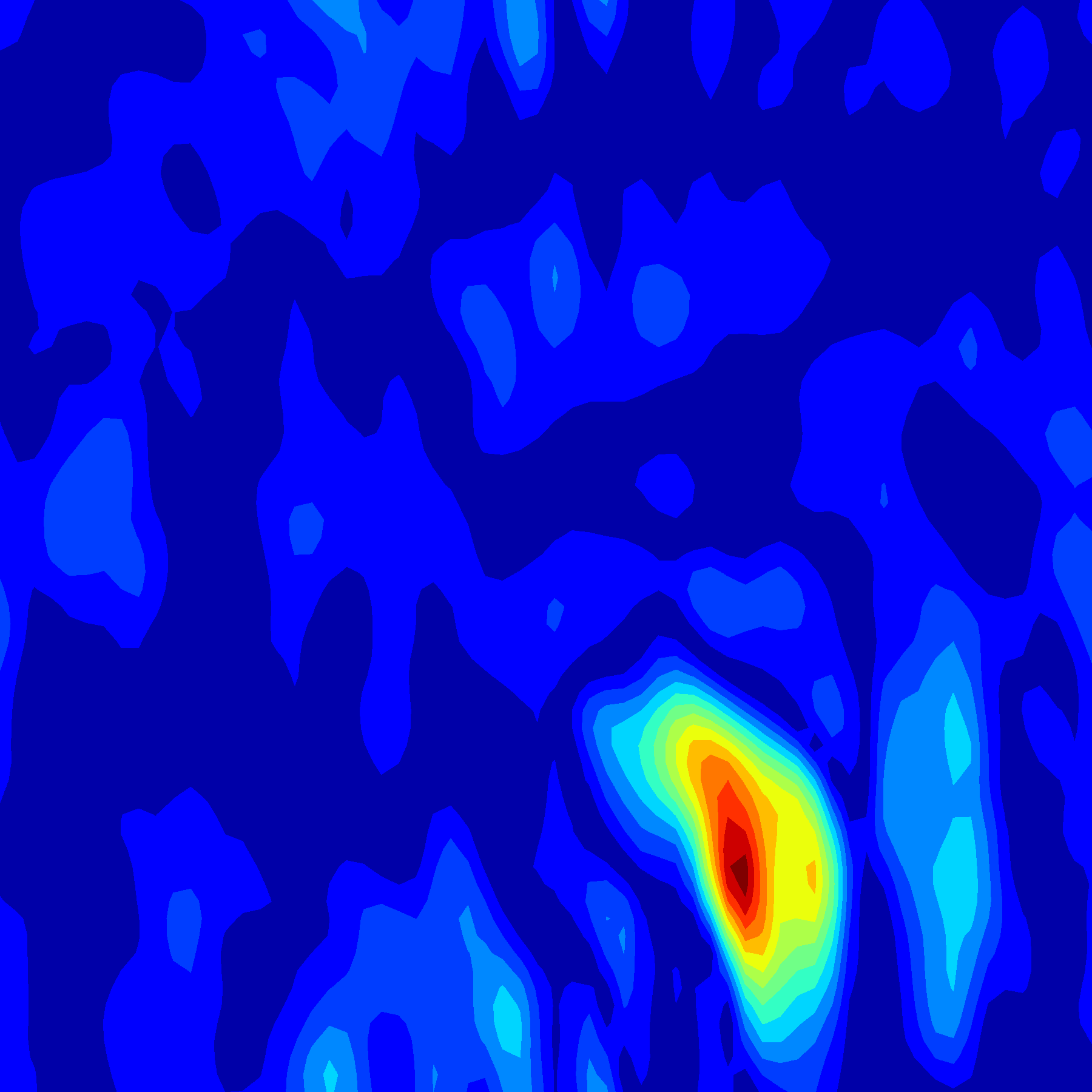}
		\caption{LISG, absolute error $t=100$}
	\end{subfigure}\hspace*{\shift}%
	\begin{subfigure}[c]{\sC\textwidth}
		\centering
		\includegraphics{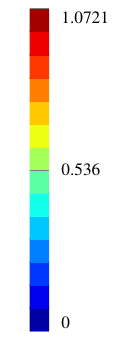}
		\caption{}
	\end{subfigure}%
	
	\caption{Navier--Stokes equations: 
		The absolute errors in the predicted solutions of the LISG method at $t=60, 80, 100$.}
	\label{fig:pred_ns_errB}
\end{figure}

\begin{figure}[ht!]
	\captionsetup[subfigure]{labelformat=empty}
	\centering
	\def\sc{0.25}
	\def\sC{0.05}
	\def\SC{0.8}
	\def\shift{-10pt}
	
	\begin{subfigure}[c]{\sc\textwidth}
		\centering
		\includegraphics[width=\SC\linewidth]{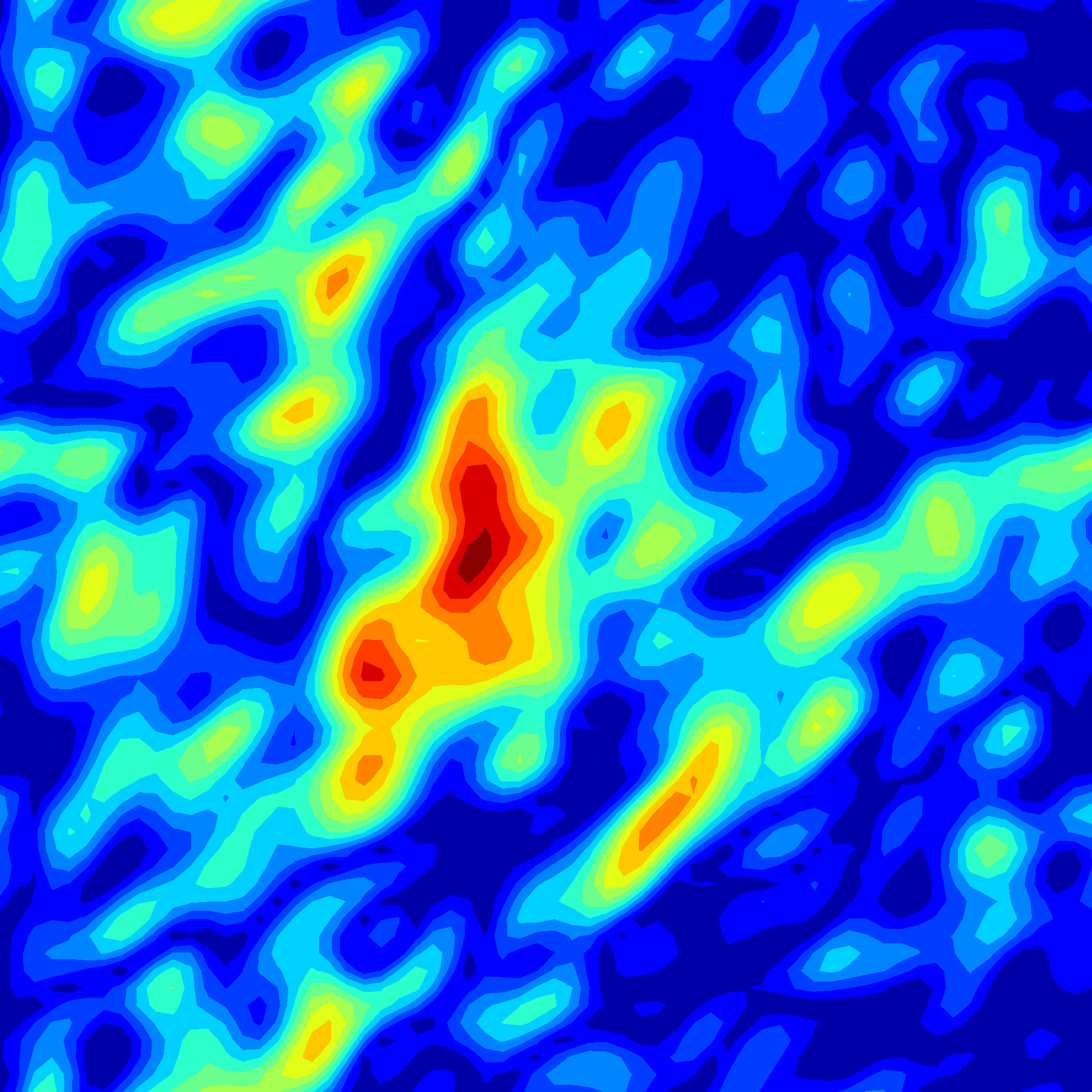}
		\caption{GDSG, absolute error $t=60$}
	\end{subfigure}\hspace*{\shift}%
	\begin{subfigure}[c]{\sC\textwidth}
		\centering
		\includegraphics{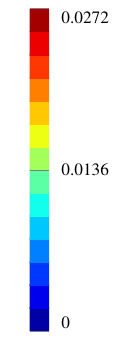}
		\caption{}
	\end{subfigure}\quad
	\begin{subfigure}[c]{\sc\textwidth}
		\centering
		\includegraphics[width=\SC\linewidth]{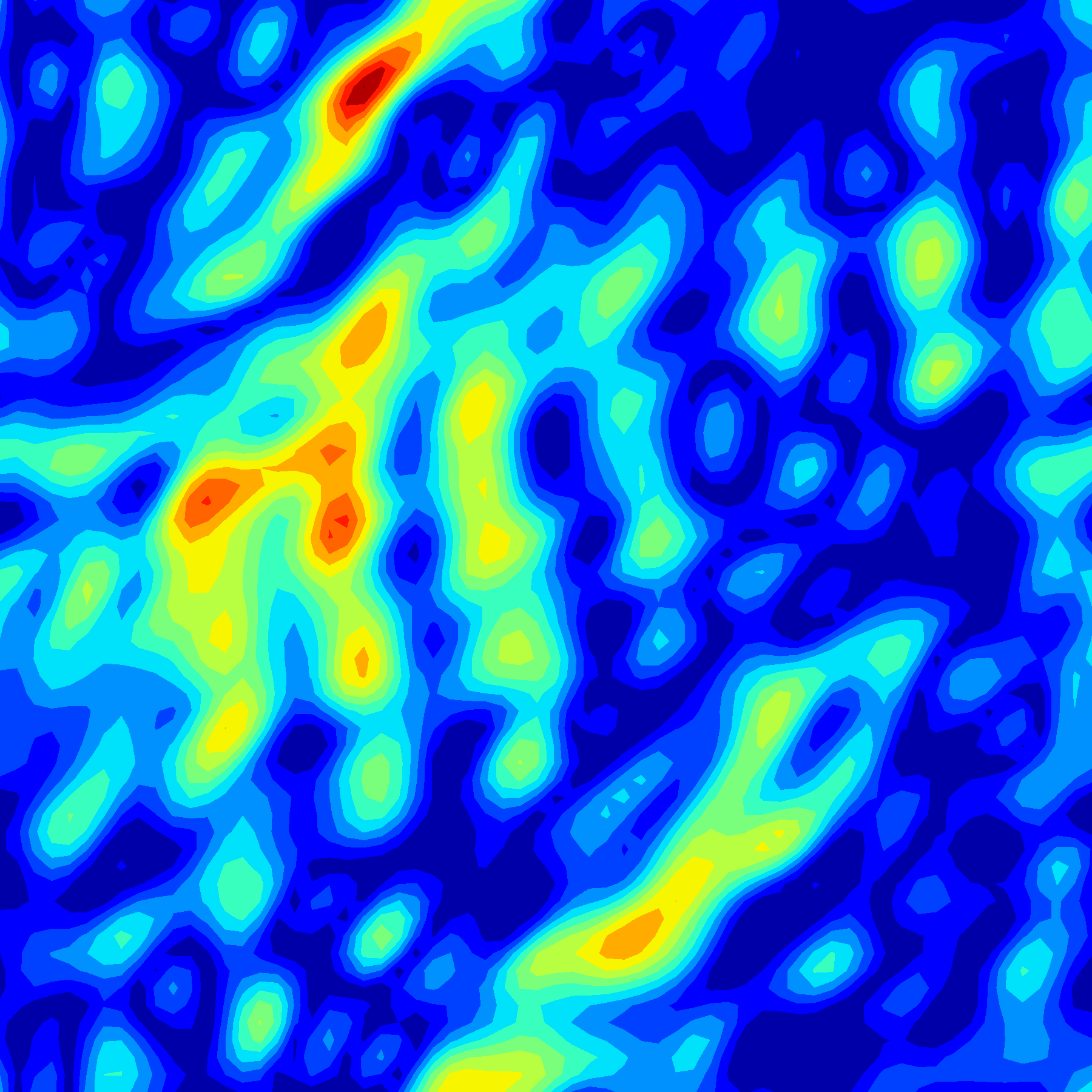}
		\caption{GDSG, absolute error $t=80$}
	\end{subfigure}\hspace*{\shift}%
	\begin{subfigure}[c]{\sC\textwidth}
		\centering
		\includegraphics{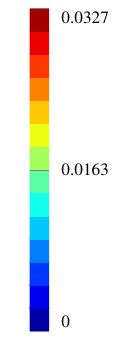}
		\caption{}
	\end{subfigure}\quad
	\begin{subfigure}[c]{\sc\textwidth}
		\centering
		\includegraphics[width=\SC\linewidth]{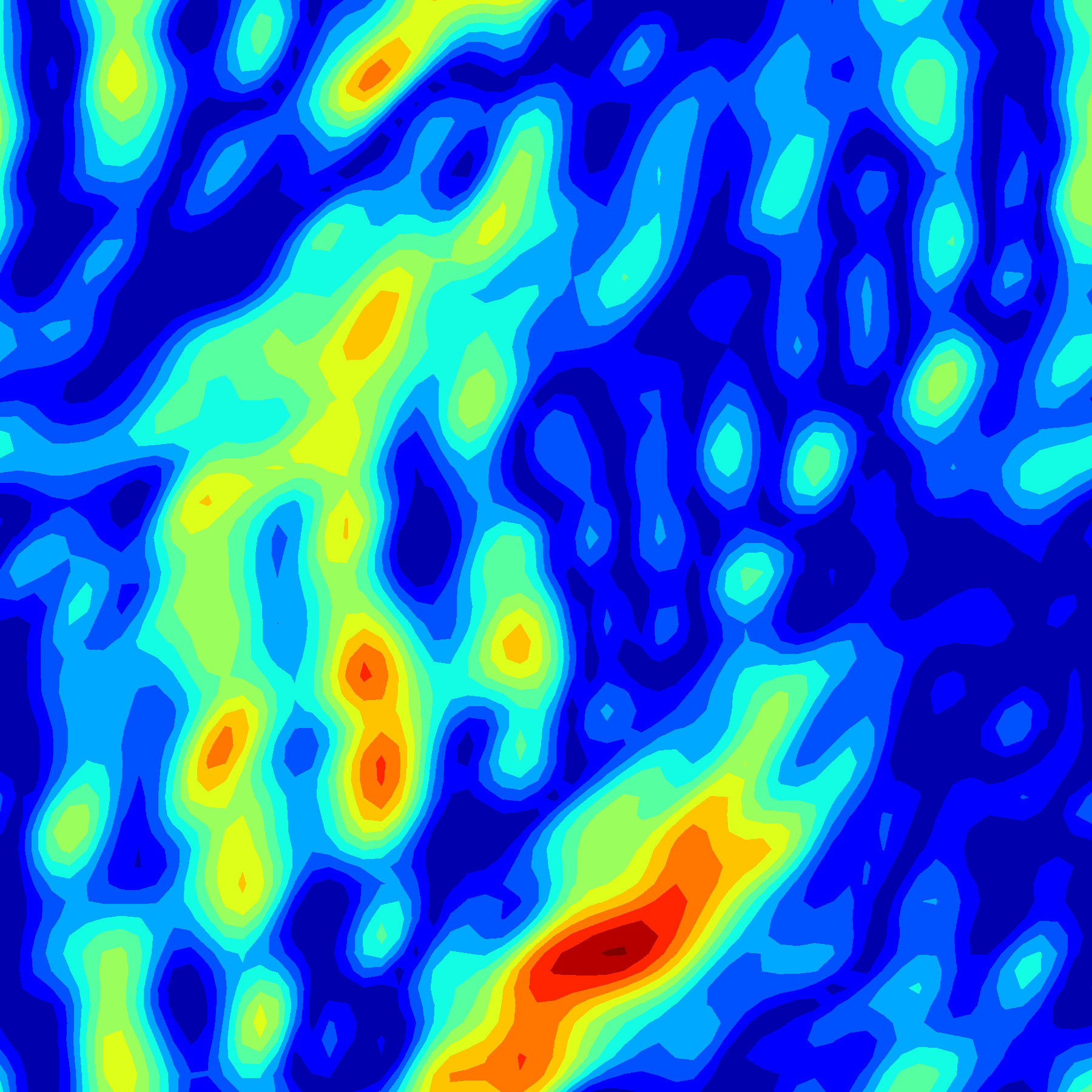}
		\caption{GDSG, absolute error $t=100$}
	\end{subfigure}\hspace*{\shift}%
	\begin{subfigure}[c]{\sC\textwidth}
		\centering
		\includegraphics{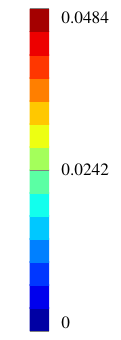}
		\caption{}
	\end{subfigure}%
	
	\caption{Navier--Stokes equations: 
		The absolute errors in the predicted solutions of the GDSG method at $t=60, 80, 100$.}
	\label{fig:pred_ns_errC}
\end{figure}

The trained models by the baseline, LISG, and GDSG methods are validated on a test dataset of $100$ long trajectories with $M=100$. The prediction errors $\{ {\mathcal E}(t_m) \}_{m=1}^M$ defined by \eqref{equation:rel_err} are computed, and their evolution is shown in Figure \ref{fig:err_ns}. 
We see that the prediction errors of the baseline method grow very fast and finally blows up, indicating the trained model is not stable. 
For both the LISG and GDSG methods, the trained models perform fairly well when $t<50$. However, the error curve for the LISG method lifts rapidly after $t>50$. Only the GDSG method produces a reliable model for $50< t \le 100$, and the relative error of its prediction at $t=100$ remains smaller than $3\%$. The average prediction errors and standard deviation 
are further compared in Table \ref{tab:resume_ns} for 
 the three methods. We clearly see the remarkable advantages 
 of the GDSG method in both accuracy and robustness. 
 For further validation, we take an initial condition from the test set and 
 compare the predicted vorticity obtained by the LISG and GDSG methods in Figure \ref{fig:pred_ns}, along with the reference solution. The corresponding absolute errors are shown in Figures  \ref{fig:pred_ns_errB} and \ref{fig:pred_ns_errC}.
 We observe that the prediction results given by the LISG method become visibly unacceptable at $t=100$, while the GDSG method still produces accurate predictions that agree well with the reference solution. All these observations further confirm the importance and remarkable benefits of embedding the semigroup property for evolution operator learning. 

\begin{table}[ht!]
	\caption{Navier--Stokes equations: Average prediction error $\overline{\mathcal{E}}$ on the test set and the standard deviation $\sigma$ of prediction errors.}
	\label{tab:resume_ns}
	\begin{center}
		\begin{tabular}{llll}
			\toprule
			\multicolumn{1}{c}{}                            &Baseline             &LISG          &GDSG  \\
			\midrule
			Prediction error $\overline{\mathcal{E}}$                  &$1.143\times 10^{31}$    &$6.034$   &$\mathbf{0.0158}$         \\
			Standard deviation $\sigma$                   &$1.51\times 10^{31}$    &$4.32\times 10^{5}$   &$\mathbf{0.0155}$           \\
			\bottomrule
		\end{tabular}
	\end{center}
\end{table}


\newpage 

\section{Conclusions}
\label{section:conclude}

In this paper, we have presented a framework for learning operators in semigroup and modeling unknown autonomous ODE and PDE systems using time series data collected at varied time lags. 
Unlike the previous {FML} works  \cite{qin2019data,wu2020data,chen2022deep} which focused on learning a single evolution operator with a fixed time step, our new framework is designed to learn a family of evolution operators with variable time steps, which constitute a semigroup for an autonomous system. 
The semigroup property is very crucial, as it links the system's evolutionary behaviors across varying time scales. 
For the first time, we have proposed an approach of embedding the semigroup property into 
the data-driven learning process, based on a novel neural network architecture and new loss functions. 
Rigorous error estimates and variance analysis have been carried out to understand 
the prediction accuracy and robustness of our approach. 
We have shown that the semigroup awareness our model 
significantly improves the prediction accuracy and greatly enhances the robustness and stability. 
Moreover, our model allows one to arbitrarily choose the time steps for prediction and ensures that the predicted results are well self-matched and consistent for different partitions of the time interval. 
The proposed framework is very feasible, can be combined with any suitable neural networks, and is applicable to data-driven learning of general autonomous ODEs and PDEs.
We have presented extensive numerical examples, including a stiff ODE system and the complicated Navier--Stokes equations, which have shown that our approach with the embedded semigroup property produces more accurate, robust, and reliable models compared to the purely data-driven approach. 
Moreover, embedding the semigroup property also notably reduces the required data, slows down the error growth over time, and greatly enhances the stability for long-time prediction. 
  
\appendix
\section{}

\subsection{Data normalization}
\label{section:dn}
In this work, every component(channel) of the vectors(the tensors) is normalized to $[-1,1]$ before network training, by using their maximum and minimum values in the dataset. 
We choose the ODE example 1 and compare the performance of the trained models with and without data normalization; see Table \ref{tab:impact_normalization}. The results clearly justify the benefits of using data normalization. 
\begin{table}[ht]
\caption{Linear ODEs: Average prediction error $\overline{\mathcal{E}}$ and standard deviation $\sigma$ on the test set, with and without data normaliaztion}
\label{tab:impact_normalization}
    \begin{center}
        \begin{tabular}{lllll}
            \toprule
            \multicolumn{1}{c}{}                                &                       &Baseline                               &LISG                                  &GDSG                            \\
            \midrule
            \multirow{2}{*}{Prediction error $\overline{\mathcal{E}}$} &With normalization     &$\mathbf{5.558\times 10^{-3}}$         &$\mathbf{1.708\times 10^{-3}}$        &$\mathbf{7.652\times 10^{-4}}$  \\
                                                                &No normalization       &$2.179\times 10^{-2}$                  &$3.736\times 10^{-3}$                 &$4.640\times 10^{-3}$           \\
            \midrule
            \multirow{2}{*}{Standard deviation $\sigma$}        &With normalization     &$1.951\times 10^{-3}$                  &$\mathbf{3.440\times 10^{-4}}$        &$\mathbf{4.932\times 10^{-5}}$  \\
                                                                &No normalization       &$\mathbf{1.479\times 10^{-3}}$         &$1.151\times 10^{-3}$                 &$4.200\times 10^{-4}$           \\
            \bottomrule
        \end{tabular}
    \end{center}
\end{table}

\subsection{Dynamic validation}
\label{section:dv}

We now present a few results to justify the advantages of using dynamic validation technique, which was mentioned in Remark \ref{rem:dyval}. 
Table \ref{tab:impact_dv} displays the results {for} the periodic attractor example (see section \ref{section:periodic_attractor}), which has merely $50$ bursts of short trajectories in the dataset. It is seen that the dynamic validation technique enhances the performance for all the three methods. 

\begin{table}[ht]
\caption{Periodic attractor: Average prediction error $\overline{\mathcal{E}}$ and standard deviation $\sigma$ on the test set, with and without dynamic validation.}
\label{tab:impact_dv}
    \begin{center}
        \begin{tabular}{lllll}
            \toprule
            \multicolumn{1}{c}{}                                &                               &Baseline                            &LISG                                  &GDSG                               \\
            \midrule
            \multirow{2}{*}{Prediction error $\overline{\mathcal{E}}$} &Dynamic validation set         &$\mathbf{6.705\times 10^{-3}}$      &$\mathbf{5.465\times 10^{-3}}$        &$\mathbf{5.175\times 10^{-3}}$     \\
                                                                &Fixed validation set           &$0.01315$                           &$0.01306$                             &$8.309\times 10^{-3}$              \\
            \midrule
            \multirow{2}{*}{Standard deviation $\sigma$}        &Dynamic validation set         &$\mathbf{6.58\times 10^{-3}}$       &$\mathbf{4.34\times 10^{-3}}$         &$\mathbf{9.38\times 10^{-4}}$      \\
                                                                &Fixed validation set           &$6.93\times 10^{-3}$                &$5.27\times 10^{-3}$                  &$1.26\times 10^{-3}$               \\
            \bottomrule
        \end{tabular}
    \end{center}
\end{table}

	\subsection{Multi-step recurrent OSG-Net}
\label{section:share_parameters}

In section \ref{sec:OSG-Net}, we have proposed a multi-step 
recursive OSG-Net architecture with multiple OSG-Net blocks, as a variant of the recursive ResNet given in \cite{qin2019data}. 
If all the blocks in recursive ResNet share the same parameters, 
then it reduces to the so-called recurrent ResNet \cite{qin2019data}. 
Following the recurrent ResNet \cite{qin2019data}, one may similarly design the \textit{multi-step recurrent OSG-Net} as follows: 
\begin{equation*}
	{\bf N}_{\theta} =  \left[  {\bf I}_n + \frac{\Delta}K {\mathcal N}_{{\bm \theta}} \bigg(\cdot, \frac{\Delta}K \bigg) \right] \circ \cdots 
	\circ \left[  {\bf I}_n + \frac{\Delta}K {\mathcal N}_{{\bm \theta}} \bigg( \cdot, \frac{\Delta}K \bigg) \right],
\end{equation*}
where the parameters $\bm{\theta}$ are shared among all the $K$ blocks. To compare the performance of recurrent and recursive OSG-Nets, we conduct experiments on the  \textit{glycolytic oscillator system} (see section \ref{section:gly}) and the \textit{advection equation} (see section \ref{section:advection_eq}). In the experiments, we adopt the recurrent and recursive OSG-Nets with $4$ blocks, and the other training configurations are the same as in sections  \ref{section:gly} and \ref{section:advection_eq}. The experimental results are shown in Table \ref{tab:multistep},  which indicates the superior performance of the recursive OSG-Net over the recurrent OSG-Net.

\begin{table}[ht]
\caption{Average prediction error $\overline{\mathcal{E}}$ on the test set.}
\label{tab:multistep}
    \begin{center}
        \begin{tabular}{lllll}
            \toprule
            \multicolumn{1}{c}{}                                &                               & Baseline                            &  LISG                                  & GDSG                               \\
            \midrule
            \multirow{2}{*}{Glycolytic oscillator} & Recursive OSG-Net         & $\mathbf{0.04439}$    & $\mathbf{0.0216}$   & $\mathbf{8.138\times 10^{-3}}$     \\
                                                                       & Recurrent OSG-Net         & $12.25$      & $0.06515$  & $0.01434$              \\
            \midrule
            \multirow{2}{*}{ Advection equation}        & Recursive OSG-Net         & $\mathbf{3.519\times 10^{-3}}$    & $\mathbf{4.062\times 10^{-3}}$   & $\mathbf{2.034\times 10^{-3}}$      \\
                                                                & Recurrent OSG-Net              & $0.01881$               & $0.01934$   & $0.01509$             \\
            \bottomrule
        \end{tabular}
    \end{center}
\end{table}


\bibliographystyle{model1-num-names}
\bibliography{bib}

\begin{thebibliography}{52}
\expandafter\ifx\csname natexlab\endcsname\relax\def\natexlab#1{#1}\fi
\providecommand{\url}[1]{\texttt{#1}}
\providecommand{\href}[2]{#2}
\providecommand{\path}[1]{#1}
\providecommand{\DOIprefix}{doi:}
\providecommand{\ArXivprefix}{arXiv:}
\providecommand{\URLprefix}{URL: }
\providecommand{\Pubmedprefix}{pmid:}
\providecommand{\doi}[1]{\href{http://dx.doi.org/#1}{\path{#1}}}
\providecommand{\Pubmed}[1]{\href{pmid:#1}{\path{#1}}}
\providecommand{\bibinfo}[2]{#2}
\ifx\xfnm\relax \def\xfnm[#1]{\unskip,\space#1}\fi
\bibitem[{Bongard and Lipson(2007)}]{bongard2007automated}
\bibinfo{author}{J.~Bongard}, \bibinfo{author}{H.~Lipson},
\newblock \bibinfo{title}{Automated reverse engineering of nonlinear dynamical
  systems},
\newblock \bibinfo{journal}{Proceedings of the National Academy of Sciences}
  \bibinfo{volume}{104} (\bibinfo{year}{2007}) \bibinfo{pages}{9943--9948}.
\bibitem[{Schmidt and Lipson(2009)}]{schmidt2009distilling}
\bibinfo{author}{M.~Schmidt}, \bibinfo{author}{H.~Lipson},
\newblock \bibinfo{title}{Distilling free-form natural laws from experimental
  data},
\newblock \bibinfo{journal}{Science} \bibinfo{volume}{324}
  (\bibinfo{year}{2009}) \bibinfo{pages}{81--85}.
\bibitem[{Brunton et~al.(2016)Brunton, Proctor, and
  Kutz}]{brunton2016discovering}
\bibinfo{author}{S.~L. Brunton}, \bibinfo{author}{J.~L. Proctor},
  \bibinfo{author}{J.~N. Kutz},
\newblock \bibinfo{title}{Discovering governing equations from data by sparse
  identification of nonlinear dynamical systems},
\newblock \bibinfo{journal}{Proceedings of the National Academy of Sciences}
  \bibinfo{volume}{113} (\bibinfo{year}{2016}) \bibinfo{pages}{3932--3937}.
\bibitem[{Rudy et~al.(2017)Rudy, Brunton, Proctor, and Kutz}]{rudy2017data}
\bibinfo{author}{S.~H. Rudy}, \bibinfo{author}{S.~L. Brunton},
  \bibinfo{author}{J.~L. Proctor}, \bibinfo{author}{J.~N. Kutz},
\newblock \bibinfo{title}{Data-driven discovery of partial differential
  equations},
\newblock \bibinfo{journal}{Science Advances} \bibinfo{volume}{3}
  (\bibinfo{year}{2017}) \bibinfo{pages}{e1602614}.
\bibitem[{Tibshirani(1996)}]{tibshirani1996regression}
\bibinfo{author}{R.~Tibshirani},
\newblock \bibinfo{title}{Regression shrinkage and selection via the lasso},
\newblock \bibinfo{journal}{Journal of the Royal Statistical Society: Series B
  (Methodological)} \bibinfo{volume}{58} (\bibinfo{year}{1996})
  \bibinfo{pages}{267--288}.
\bibitem[{Donoho(2006)}]{donoho2006compressed}
\bibinfo{author}{D.~L. Donoho},
\newblock \bibinfo{title}{Compressed sensing},
\newblock \bibinfo{journal}{IEEE Transactions on information theory}
  \bibinfo{volume}{52} (\bibinfo{year}{2006}) \bibinfo{pages}{1289--1306}.
\bibitem[{Candes et~al.(2006)Candes, Romberg, and Tao}]{candes2006stable}
\bibinfo{author}{E.~J. Candes}, \bibinfo{author}{J.~K. Romberg},
  \bibinfo{author}{T.~Tao},
\newblock \bibinfo{title}{Stable signal recovery from incomplete and inaccurate
  measurements},
\newblock \bibinfo{journal}{Communications on Pure and Applied Mathematics}
  \bibinfo{volume}{59} (\bibinfo{year}{2006}) \bibinfo{pages}{1207--1223}.
\bibitem[{Schaeffer and McCalla(2017)}]{schaeffer2017sparse}
\bibinfo{author}{H.~Schaeffer}, \bibinfo{author}{S.~G. McCalla},
\newblock \bibinfo{title}{Sparse model selection via integral terms},
\newblock \bibinfo{journal}{Physical Review E} \bibinfo{volume}{96}
  (\bibinfo{year}{2017}) \bibinfo{pages}{023302}.
\bibitem[{Schaeffer(2017)}]{schaeffer2017learning}
\bibinfo{author}{H.~Schaeffer},
\newblock \bibinfo{title}{Learning partial differential equations via data
  discovery and sparse optimization},
\newblock \bibinfo{journal}{Proceedings of the Royal Society A: Mathematical,
  Physical and Engineering Sciences} \bibinfo{volume}{473}
  (\bibinfo{year}{2017}) \bibinfo{pages}{20160446}.
\bibitem[{Tran and Ward(2017)}]{tran2017exact}
\bibinfo{author}{G.~Tran}, \bibinfo{author}{R.~Ward},
\newblock \bibinfo{title}{Exact recovery of chaotic systems from highly
  corrupted data},
\newblock \bibinfo{journal}{Multiscale Modeling \& Simulation}
  \bibinfo{volume}{15} (\bibinfo{year}{2017}) \bibinfo{pages}{1108--1129}.
\bibitem[{Long et~al.(2018)Long, Lu, Ma, and Dong}]{long2018pde}
\bibinfo{author}{Z.~Long}, \bibinfo{author}{Y.~Lu}, \bibinfo{author}{X.~Ma},
  \bibinfo{author}{B.~Dong},
\newblock \bibinfo{title}{P{DE}-{N}et: {L}earning {PDE}s from data},
\newblock in: \bibinfo{booktitle}{International Conference on Machine
  Learning}, \bibinfo{organization}{PMLR}, \bibinfo{year}{2018}, pp.
  \bibinfo{pages}{3208--3216}.
\bibitem[{Raissi(2018)}]{raissi2018deep}
\bibinfo{author}{M.~Raissi},
\newblock \bibinfo{title}{Deep hidden physics models: Deep learning of
  nonlinear partial differential equations},
\newblock \bibinfo{journal}{The Journal of Machine Learning Research}
  \bibinfo{volume}{19} (\bibinfo{year}{2018}) \bibinfo{pages}{932--955}.
\bibitem[{Raissi et~al.(2018)Raissi, Perdikaris, and
  Karniadakis}]{raissi2018multistep}
\bibinfo{author}{M.~Raissi}, \bibinfo{author}{P.~Perdikaris},
  \bibinfo{author}{G.~E. Karniadakis},
\newblock \bibinfo{title}{Multistep neural networks for data-driven discovery
  of nonlinear dynamical systems},
\newblock \bibinfo{journal}{arXiv preprint arXiv:1801.01236}
  (\bibinfo{year}{2018}).
\bibitem[{Long et~al.(2019)Long, Lu, and Dong}]{long2019pde}
\bibinfo{author}{Z.~Long}, \bibinfo{author}{Y.~Lu}, \bibinfo{author}{B.~Dong},
\newblock \bibinfo{title}{P{DE}-{N}et 2.0: {L}earning {PDE}s from data with a
  numeric-symbolic hybrid deep network},
\newblock \bibinfo{journal}{Journal of Computational Physics}
  \bibinfo{volume}{399} (\bibinfo{year}{2019}) \bibinfo{pages}{108925}.
\bibitem[{Raissi et~al.(2019)Raissi, Perdikaris, and
  Karniadakis}]{raissi2019physics}
\bibinfo{author}{M.~Raissi}, \bibinfo{author}{P.~Perdikaris},
  \bibinfo{author}{G.~E. Karniadakis},
\newblock \bibinfo{title}{Physics-informed neural networks: A deep learning
  framework for solving forward and inverse problems involving nonlinear
  partial differential equations},
\newblock \bibinfo{journal}{Journal of Computational physics}
  \bibinfo{volume}{378} (\bibinfo{year}{2019}) \bibinfo{pages}{686--707}.
\bibitem[{Sun et~al.(2020)Sun, Zhang, and Schaeffer}]{sun2020neupde}
\bibinfo{author}{Y.~Sun}, \bibinfo{author}{L.~Zhang},
  \bibinfo{author}{H.~Schaeffer},
\newblock \bibinfo{title}{Neu{PDE}: {N}eural network based ordinary and partial
  differential equations for modeling time-dependent data},
\newblock in: \bibinfo{booktitle}{Mathematical and Scientific Machine
  Learning}, \bibinfo{organization}{PMLR}, \bibinfo{year}{2020}, pp.
  \bibinfo{pages}{352--372}.
\bibitem[{Chen et~al.(2021)Chen, Matsubara, and Yaguchi}]{chen2021neural}
\bibinfo{author}{Y.~Chen}, \bibinfo{author}{T.~Matsubara},
  \bibinfo{author}{T.~Yaguchi},
\newblock \bibinfo{title}{Neural symplectic form: {L}earning {H}amiltonian
  equations on general coordinate systems},
\newblock \bibinfo{journal}{Advances in Neural Information Processing Systems}
  \bibinfo{volume}{34} (\bibinfo{year}{2021}) \bibinfo{pages}{16659--16670}.
\bibitem[{Atkinson et~al.(2019)Atkinson, Subber, Wang, Khan, Hawi, and
  Ghanem}]{atkinson2019data}
\bibinfo{author}{S.~Atkinson}, \bibinfo{author}{W.~Subber},
  \bibinfo{author}{L.~Wang}, \bibinfo{author}{G.~Khan},
  \bibinfo{author}{P.~Hawi}, \bibinfo{author}{R.~Ghanem},
\newblock \bibinfo{title}{Data-driven discovery of free-form governing
  differential equations},
\newblock \bibinfo{journal}{arXiv preprint arXiv:1910.05117}
  (\bibinfo{year}{2019}).
\bibitem[{Chen et~al.(2022)Chen, Luo, Liu, Xu, and Zhang}]{chen2022symbolic}
\bibinfo{author}{Y.~Chen}, \bibinfo{author}{Y.~Luo}, \bibinfo{author}{Q.~Liu},
  \bibinfo{author}{H.~Xu}, \bibinfo{author}{D.~Zhang},
\newblock \bibinfo{title}{Symbolic genetic algorithm for discovering open-form
  partial differential equations ({SGA}-{PDE})},
\newblock \bibinfo{journal}{Physical Review Research} \bibinfo{volume}{4}
  (\bibinfo{year}{2022}) \bibinfo{pages}{023174}.
\bibitem[{Raissi et~al.(2017)Raissi, Perdikaris, and
  Karniadakis}]{raissi2017machine}
\bibinfo{author}{M.~Raissi}, \bibinfo{author}{P.~Perdikaris},
  \bibinfo{author}{G.~E. Karniadakis},
\newblock \bibinfo{title}{Machine learning of linear differential equations
  using {G}aussian processes},
\newblock \bibinfo{journal}{Journal of Computational Physics}
  \bibinfo{volume}{348} (\bibinfo{year}{2017}) \bibinfo{pages}{683--693}.
\bibitem[{Mangan et~al.(2017)Mangan, Kutz, Brunton, and
  Proctor}]{mangan2017model}
\bibinfo{author}{N.~M. Mangan}, \bibinfo{author}{J.~N. Kutz},
  \bibinfo{author}{S.~L. Brunton}, \bibinfo{author}{J.~L. Proctor},
\newblock \bibinfo{title}{Model selection for dynamical systems via sparse
  regression and information criteria},
\newblock \bibinfo{journal}{Proceedings of the Royal Society A: Mathematical,
  Physical and Engineering Sciences} \bibinfo{volume}{473}
  (\bibinfo{year}{2017}) \bibinfo{pages}{20170009}.
\bibitem[{Brunton et~al.(2017)Brunton, Brunton, Proctor, Kaiser, and
  Kutz}]{brunton2017chaos}
\bibinfo{author}{S.~L. Brunton}, \bibinfo{author}{B.~W. Brunton},
  \bibinfo{author}{J.~L. Proctor}, \bibinfo{author}{E.~Kaiser},
  \bibinfo{author}{J.~N. Kutz},
\newblock \bibinfo{title}{Chaos as an intermittently forced linear system},
\newblock \bibinfo{journal}{Nature Communications} \bibinfo{volume}{8}
  (\bibinfo{year}{2017}) \bibinfo{pages}{19}.
\bibitem[{Wu and Xiu(2019)}]{wu2019numerical}
\bibinfo{author}{K.~Wu}, \bibinfo{author}{D.~Xiu},
\newblock \bibinfo{title}{Numerical aspects for approximating governing
  equations using data},
\newblock \bibinfo{journal}{Journal of Computational Physics}
  \bibinfo{volume}{384} (\bibinfo{year}{2019}) \bibinfo{pages}{200--221}.
\bibitem[{Wu et~al.(2020)Wu, Qin, and Xiu}]{wu2020structure}
\bibinfo{author}{K.~Wu}, \bibinfo{author}{T.~Qin}, \bibinfo{author}{D.~Xiu},
\newblock \bibinfo{title}{Structure-preserving method for reconstructing
  unknown {H}amiltonian systems from trajectory data},
\newblock \bibinfo{journal}{SIAM Journal on Scientific Computing}
  \bibinfo{volume}{42} (\bibinfo{year}{2020}) \bibinfo{pages}{A3704--A3729}.
\bibitem[{Keller and Du(2021)}]{keller2021discovery}
\bibinfo{author}{R.~T. Keller}, \bibinfo{author}{Q.~Du},
\newblock \bibinfo{title}{Discovery of dynamics using linear multistep
  methods},
\newblock \bibinfo{journal}{SIAM Journal on Numerical Analysis}
  \bibinfo{volume}{59} (\bibinfo{year}{2021}) \bibinfo{pages}{429--455}.
\bibitem[{Xu et~al.(2020)Xu, Chang, and Zhang}]{xu2020dlga}
\bibinfo{author}{H.~Xu}, \bibinfo{author}{H.~Chang},
  \bibinfo{author}{D.~Zhang},
\newblock \bibinfo{title}{D{LGA}-{PDE}: Discovery of {PDE}s with incomplete
  candidate library via combination of deep learning and genetic algorithm},
\newblock \bibinfo{journal}{Journal of Computational Physics}
  \bibinfo{volume}{418} (\bibinfo{year}{2020}) \bibinfo{pages}{109584}.
\bibitem[{Xu et~al.(2021)Xu, Zhang, and Zeng}]{xu2021deep}
\bibinfo{author}{H.~Xu}, \bibinfo{author}{D.~Zhang}, \bibinfo{author}{J.~Zeng},
\newblock \bibinfo{title}{Deep-learning of parametric partial differential
  equations from sparse and noisy data},
\newblock \bibinfo{journal}{Physics of Fluids} \bibinfo{volume}{33}
  (\bibinfo{year}{2021}) \bibinfo{pages}{037132}.
\bibitem[{Xu and Zhang(2021)}]{xu2021robust}
\bibinfo{author}{H.~Xu}, \bibinfo{author}{D.~Zhang},
\newblock \bibinfo{title}{Robust discovery of partial differential equations in
  complex situations},
\newblock \bibinfo{journal}{Physical Review Research} \bibinfo{volume}{3}
  (\bibinfo{year}{2021}) \bibinfo{pages}{033270}.
\bibitem[{Qin et~al.(2019)Qin, Wu, and Xiu}]{qin2019data}
\bibinfo{author}{T.~Qin}, \bibinfo{author}{K.~Wu}, \bibinfo{author}{D.~Xiu},
\newblock \bibinfo{title}{Data driven governing equations approximation using
  deep neural networks},
\newblock \bibinfo{journal}{Journal of Computational Physics}
  \bibinfo{volume}{395} (\bibinfo{year}{2019}) \bibinfo{pages}{620--635}.
\bibitem[{Wu and Xiu(2020)}]{wu2020data}
\bibinfo{author}{K.~Wu}, \bibinfo{author}{D.~Xiu},
\newblock \bibinfo{title}{Data-driven deep learning of partial differential
  equations in modal space},
\newblock \bibinfo{journal}{Journal of Computational Physics}
  \bibinfo{volume}{408} (\bibinfo{year}{2020}) \bibinfo{pages}{109307}.
\bibitem[{Chen et~al.(2022)Chen, Churchill, Wu, and Xiu}]{chen2022deep}
\bibinfo{author}{Z.~Chen}, \bibinfo{author}{V.~Churchill},
  \bibinfo{author}{K.~Wu}, \bibinfo{author}{D.~Xiu},
\newblock \bibinfo{title}{Deep neural network modeling of unknown partial
  differential equations in nodal space},
\newblock \bibinfo{journal}{Journal of Computational Physics}
  \bibinfo{volume}{449} (\bibinfo{year}{2022}) \bibinfo{pages}{110782}.
\bibitem[{Churchill and Xiu(2007)}]{churchill2023flow}
\bibinfo{author}{V.~Churchill}, \bibinfo{author}{D.~Xiu},
\newblock \bibinfo{title}{Flow map learning for unknown dynamical systems:
  Overview, implementation, and benchmarks},
\newblock \bibinfo{journal}{Journal of Machine Learning for Modeling and
  Computing}  (\bibinfo{year}{2007}).
\bibitem[{Qin et~al.(2021)Qin, Chen, Jakeman, and Xiu}]{qin2021deep}
\bibinfo{author}{T.~Qin}, \bibinfo{author}{Z.~Chen}, \bibinfo{author}{J.~D.
  Jakeman}, \bibinfo{author}{D.~Xiu},
\newblock \bibinfo{title}{Deep learning of parameterized equations with
  applications to uncertainty quantification},
\newblock \bibinfo{journal}{International Journal for Uncertainty
  Quantification} \bibinfo{volume}{11} (\bibinfo{year}{2021}).
\bibitem[{Fu et~al.(2020)Fu, Chang, and Xiu}]{fu2020learning}
\bibinfo{author}{X.~Fu}, \bibinfo{author}{L.-B. Chang},
  \bibinfo{author}{D.~Xiu},
\newblock \bibinfo{title}{Learning reduced systems via deep neural networks
  with memory},
\newblock \bibinfo{journal}{Journal of Machine Learning for Modeling and
  Computing} \bibinfo{volume}{1} (\bibinfo{year}{2020}).
\bibitem[{Qin et~al.(2021)Qin, Chen, Jakeman, and Xiu}]{qin2021data}
\bibinfo{author}{T.~Qin}, \bibinfo{author}{Z.~Chen}, \bibinfo{author}{J.~D.
  Jakeman}, \bibinfo{author}{D.~Xiu},
\newblock \bibinfo{title}{Data-driven learning of nonautonomous systems},
\newblock \bibinfo{journal}{SIAM Journal on Scientific Computing}
  \bibinfo{volume}{43} (\bibinfo{year}{2021}) \bibinfo{pages}{A1607--A1624}.
\bibitem[{Chen and Xiu(2021)}]{chen2021generalized}
\bibinfo{author}{Z.~Chen}, \bibinfo{author}{D.~Xiu},
\newblock \bibinfo{title}{On generalized residual network for deep learning of
  unknown dynamical systems},
\newblock \bibinfo{journal}{Journal of Computational Physics}
  \bibinfo{volume}{438} (\bibinfo{year}{2021}) \bibinfo{pages}{110362}.
\bibitem[{Su et~al.(2021)Su, Chou, and Xiu}]{su2021deep}
\bibinfo{author}{W.-H. Su}, \bibinfo{author}{C.-S. Chou},
  \bibinfo{author}{D.~Xiu},
\newblock \bibinfo{title}{Deep learning of biological models from data:
  {A}pplications to {ODE} models},
\newblock \bibinfo{journal}{Bulletin of Mathematical Biology}
  \bibinfo{volume}{83} (\bibinfo{year}{2021}) \bibinfo{pages}{1--19}.
\bibitem[{Churchill and Xiu(2022)}]{churchill2022deep}
\bibinfo{author}{V.~Churchill}, \bibinfo{author}{D.~Xiu},
\newblock \bibinfo{title}{Deep learning of chaotic systems from
  partially-observed data},
\newblock \bibinfo{journal}{Journal of Machine Learning for Modeling and
  Computing} \bibinfo{volume}{3} (\bibinfo{year}{2022}).
\bibitem[{Chen and Xiu(2023)}]{chen2023learning}
\bibinfo{author}{Y.~Chen}, \bibinfo{author}{D.~Xiu},
\newblock \bibinfo{title}{Learning stochastic dynamical system via flow map
  operator},
\newblock \bibinfo{journal}{arXiv preprint arXiv:2305.03874}
  (\bibinfo{year}{2023}).
\bibitem[{Lu et~al.(2021)Lu, Jin, Pang, Zhang, and
  Karniadakis}]{lu2021learning}
\bibinfo{author}{L.~Lu}, \bibinfo{author}{P.~Jin}, \bibinfo{author}{G.~Pang},
  \bibinfo{author}{Z.~Zhang}, \bibinfo{author}{G.~E. Karniadakis},
\newblock \bibinfo{title}{Learning nonlinear operators via {D}eep{ON}et based
  on the universal approximation theorem of operators},
\newblock \bibinfo{journal}{Nature Machine Intelligence} \bibinfo{volume}{3}
  (\bibinfo{year}{2021}) \bibinfo{pages}{218--229}.
\bibitem[{Cai et~al.(2021)Cai, Wang, Lu, Zaki, and Karniadakis}]{cai2021deepm}
\bibinfo{author}{S.~Cai}, \bibinfo{author}{Z.~Wang}, \bibinfo{author}{L.~Lu},
  \bibinfo{author}{T.~A. Zaki}, \bibinfo{author}{G.~E. Karniadakis},
\newblock \bibinfo{title}{Deep{M}\&{M}net: Inferring the electroconvection
  multiphysics fields based on operator approximation by neural networks},
\newblock \bibinfo{journal}{Journal of Computational Physics}
  \bibinfo{volume}{436} (\bibinfo{year}{2021}) \bibinfo{pages}{110296}.
\bibitem[{Mao et~al.(2021)Mao, Lu, Marxen, Zaki, and
  Karniadakis}]{mao2021deepm}
\bibinfo{author}{Z.~Mao}, \bibinfo{author}{L.~Lu}, \bibinfo{author}{O.~Marxen},
  \bibinfo{author}{T.~A. Zaki}, \bibinfo{author}{G.~E. Karniadakis},
\newblock \bibinfo{title}{Deep{M}\&{M}net for hypersonics: Predicting the
  coupled flow and finite-rate chemistry behind a normal shock using
  neural-network approximation of operators},
\newblock \bibinfo{journal}{Journal of computational physics}
  \bibinfo{volume}{447} (\bibinfo{year}{2021}) \bibinfo{pages}{110698}.
\bibitem[{Kovachki et~al.(2021)Kovachki, Li, Liu, Azizzadenesheli,
  Bhattacharya, Stuart, and Anandkumar}]{kovachki2021neural}
\bibinfo{author}{N.~Kovachki}, \bibinfo{author}{Z.~Li},
  \bibinfo{author}{B.~Liu}, \bibinfo{author}{K.~Azizzadenesheli},
  \bibinfo{author}{K.~Bhattacharya}, \bibinfo{author}{A.~Stuart},
  \bibinfo{author}{A.~Anandkumar},
\newblock \bibinfo{title}{Neural operator: Learning maps between function
  spaces},
\newblock \bibinfo{journal}{arXiv preprint arXiv:2108.08481}
  (\bibinfo{year}{2021}).
\bibitem[{Li et~al.(2020)Li, Kovachki, Azizzadenesheli, Liu, Stuart,
  Bhattacharya, and Anandkumar}]{li2020multipole}
\bibinfo{author}{Z.~Li}, \bibinfo{author}{N.~Kovachki},
  \bibinfo{author}{K.~Azizzadenesheli}, \bibinfo{author}{B.~Liu},
  \bibinfo{author}{A.~Stuart}, \bibinfo{author}{K.~Bhattacharya},
  \bibinfo{author}{A.~Anandkumar},
\newblock \bibinfo{title}{Multipole graph neural operator for parametric
  partial differential equations},
\newblock \bibinfo{journal}{Advances in Neural Information Processing Systems}
  \bibinfo{volume}{33} (\bibinfo{year}{2020}) \bibinfo{pages}{6755--6766}.
\bibitem[{Li et~al.(2021{\natexlab{a}})Li, Kovachki, Azizzadenesheli, liu,
  Bhattacharya, Stuart, and Anandkumar}]{li2021fourier}
\bibinfo{author}{Z.~Li}, \bibinfo{author}{N.~B. Kovachki},
  \bibinfo{author}{K.~Azizzadenesheli}, \bibinfo{author}{B.~liu},
  \bibinfo{author}{K.~Bhattacharya}, \bibinfo{author}{A.~Stuart},
  \bibinfo{author}{A.~Anandkumar},
\newblock \bibinfo{title}{Fourier neural operator for parametric partial
  differential equations},
\newblock in: \bibinfo{booktitle}{International Conference on Learning
  Representations}, \bibinfo{year}{2021}{\natexlab{a}}.
\bibitem[{Li et~al.(2021{\natexlab{b}})Li, Kovachki, Azizzadenesheli, Liu,
  Bhattacharya, Stuart, and Anandkumar}]{li2021markov}
\bibinfo{author}{Z.~Li}, \bibinfo{author}{N.~Kovachki},
  \bibinfo{author}{K.~Azizzadenesheli}, \bibinfo{author}{B.~Liu},
  \bibinfo{author}{K.~Bhattacharya}, \bibinfo{author}{A.~Stuart},
  \bibinfo{author}{A.~Anandkumar},
\newblock \bibinfo{title}{Markov neural operators for learning chaotic
  systems},
\newblock \bibinfo{journal}{arXiv preprint arXiv:2106.06898}
  (\bibinfo{year}{2021}{\natexlab{b}}).
\bibitem[{You et~al.(2022)You, Yu, D'Elia, Gao, and Silling}]{you2022nonlocal}
\bibinfo{author}{H.~You}, \bibinfo{author}{Y.~Yu}, \bibinfo{author}{M.~D'Elia},
  \bibinfo{author}{T.~Gao}, \bibinfo{author}{S.~Silling},
\newblock \bibinfo{title}{Nonlocal kernel network ({NKN}): a stable and
  resolution-independent deep neural network},
\newblock \bibinfo{journal}{Journal of Computational Physics}
  \bibinfo{volume}{469} (\bibinfo{year}{2022}) \bibinfo{pages}{111536}.
\bibitem[{He et~al.(2016)He, Zhang, Ren, and Sun}]{he2016deep}
\bibinfo{author}{K.~He}, \bibinfo{author}{X.~Zhang}, \bibinfo{author}{S.~Ren},
  \bibinfo{author}{J.~Sun},
\newblock \bibinfo{title}{Deep residual learning for image recognition},
\newblock in: \bibinfo{booktitle}{Proceedings of the IEEE conference on
  computer vision and pattern recognition}, \bibinfo{year}{2016}, pp.
  \bibinfo{pages}{770--778}.
\bibitem[{Smith(2017)}]{smith2017cyclical}
\bibinfo{author}{L.~N. Smith},
\newblock \bibinfo{title}{Cyclical learning rates for training neural
  networks},
\newblock in: \bibinfo{booktitle}{2017 IEEE {W}inter {C}onference on
  {A}pplications of {C}omputer {V}ision (WACV)}, \bibinfo{organization}{IEEE},
  \bibinfo{year}{2017}, pp. \bibinfo{pages}{464--472}.
\bibitem[{Abadi et~al.(2015)Abadi, Agarwal, Barham, Brevdo, Chen, Citro,
  Corrado, Davis, Dean, Devin, Ghemawat, Goodfellow, Harp, Irving, Isard, Jia,
  Jozefowicz, Kaiser, Kudlur, Levenberg, Man\'{e}, Monga, Moore, Murray, Olah,
  Schuster, Shlens, Steiner, Sutskever, Talwar, Tucker, Vanhoucke, Vasudevan,
  Vi\'{e}gas, Vinyals, Warden, Wattenberg, Wicke, Yu, and
  Zheng}]{tensorflow2015-whitepaper}
\bibinfo{author}{M.~Abadi}, \bibinfo{author}{A.~Agarwal},
  \bibinfo{author}{P.~Barham}, \bibinfo{author}{E.~Brevdo},
  \bibinfo{author}{Z.~Chen}, \bibinfo{author}{C.~Citro}, \bibinfo{author}{G.~S.
  Corrado}, \bibinfo{author}{A.~Davis}, \bibinfo{author}{J.~Dean},
  \bibinfo{author}{M.~Devin}, \bibinfo{author}{S.~Ghemawat},
  \bibinfo{author}{I.~Goodfellow}, \bibinfo{author}{A.~Harp},
  \bibinfo{author}{G.~Irving}, \bibinfo{author}{M.~Isard},
  \bibinfo{author}{Y.~Jia}, \bibinfo{author}{R.~Jozefowicz},
  \bibinfo{author}{L.~Kaiser}, \bibinfo{author}{M.~Kudlur},
  \bibinfo{author}{J.~Levenberg}, \bibinfo{author}{D.~Man\'{e}},
  \bibinfo{author}{R.~Monga}, \bibinfo{author}{S.~Moore},
  \bibinfo{author}{D.~Murray}, \bibinfo{author}{C.~Olah},
  \bibinfo{author}{M.~Schuster}, \bibinfo{author}{J.~Shlens},
  \bibinfo{author}{B.~Steiner}, \bibinfo{author}{I.~Sutskever},
  \bibinfo{author}{K.~Talwar}, \bibinfo{author}{P.~Tucker},
  \bibinfo{author}{V.~Vanhoucke}, \bibinfo{author}{V.~Vasudevan},
  \bibinfo{author}{F.~Vi\'{e}gas}, \bibinfo{author}{O.~Vinyals},
  \bibinfo{author}{P.~Warden}, \bibinfo{author}{M.~Wattenberg},
  \bibinfo{author}{M.~Wicke}, \bibinfo{author}{Y.~Yu},
  \bibinfo{author}{X.~Zheng}, \bibinfo{title}{{TensorFlow}: Large-scale machine
  learning on heterogeneous systems}, \bibinfo{year}{2015}. \URLprefix
  \url{https://www.tensorflow.org/}, \bibinfo{note}{software available from
  tensorflow.org}.
\bibitem[{Robertson(1966)}]{robertson1966solution}
\bibinfo{author}{H.~Robertson},
\newblock \bibinfo{title}{The solution of a set of reaction rate equations},
\newblock \bibinfo{journal}{Numerical analysis: An introduction}
  \bibinfo{volume}{178182} (\bibinfo{year}{1966}).
\bibitem[{Daniels and Nemenman(2015)}]{daniels2015efficient}
\bibinfo{author}{B.~C. Daniels}, \bibinfo{author}{I.~Nemenman},
\newblock \bibinfo{title}{Efficient inference of parsimonious phenomenological
  models of cellular dynamics using {S}-systems and alternating regression},
\newblock \bibinfo{journal}{Plo{S} {O}ne} \bibinfo{volume}{10}
  (\bibinfo{year}{2015}) \bibinfo{pages}{e0119821}.

\end{thebibliography}
\end{document}